%% file: 0_neurips_2025.tex
\documentclass{article}

    \PassOptionsToPackage{numbers, compress}{natbib}

\usepackage[dandb, final]{neurips_2025}

\usepackage[utf8]{inputenc} 
\usepackage[T1]{fontenc}    
\usepackage{hyperref}       
\usepackage{url}            
\usepackage{booktabs}       
\usepackage{amsfonts}       
\usepackage{nicefrac}       
\usepackage{microtype}      
\usepackage[dvipsnames]{xcolor}         
\usepackage{caption}
\usepackage{float}
\usepackage{titletoc}
\usepackage{soul}
\usepackage{hhline}
\usepackage{multirow}
\usepackage{graphicx}
\usepackage{pifont}
\usepackage{amsmath}
\usepackage{newtxtext}
\usepackage{subcaption} 
\usepackage{makecell}
\usepackage{multirow}
\usepackage{colortbl}
\usepackage{array}
\usepackage{rotating}
\usepackage{wrapfig}
\usepackage{enumitem}
\hypersetup{
    colorlinks = true,
    linkcolor = red,
    citecolor = Magenta,
    urlcolor = blue
}
\newcommand{\da}[1]{\(\color{red}{\downarrow}\)}
\newcommand{\cmark}{\text{\ding{51}}}
\newcommand{\xmark}{\text{\ding{55}}}

\newcommand{\DatasetName}{DeepMoiréFake}

\title{Through the Lens: Benchmarking Deepfake Detectors Against Moiré-Induced Distortions}

\author{%
  Razaib Tariq\thanks{Equal contribution.} \qquad Minji Heo\footnotemark[1] \qquad Simon S. Woo\thanks{Corresponding author.}\\
  Sungkyunkwan University, South Korea\\
  \texttt{\{razaibtariq,minji.h0224,swoo\}@g.skku.edu} \\
  \And
  Shahroz Tariq\footnotemark[2] \\
CSIRO's Data61, Australia\\
\texttt{shahroz.tariq@data61.csiro.au}}

\begin{document}

\maketitle

\begin{abstract}
Deepfake detection remains a pressing challenge, particularly in real-world settings where smartphone-captured media from digital screens often introduces Moiré artifacts that can distort detection outcomes. This study systematically evaluates state-of-the-art (SOTA) deepfake detectors on Moiré-affected videos, an issue that has received little attention. We collected a dataset of 12,832 videos, spanning 35.64 hours, from the Celeb-DF, DFD, DFDC, UADFV, and FF++ datasets, capturing footage under diverse real-world conditions, including varying screens, smartphones, lighting setups, and camera angles. To further examine the influence of Moiré patterns on deepfake detection, we conducted additional experiments using our DeepMoiréFake, referred to as (DMF) dataset and two synthetic Moiré generation techniques. Across 15 top-performing detectors, our results show that Moiré artifacts degrade performance by as much as 25.4\%, while synthetically generated Moiré patterns lead to a 21.4\% drop in accuracy. Surprisingly, demoiréing methods, intended as a mitigation approach, instead worsened the problem, reducing accuracy by up to 17.2\%. These findings underscore the urgent need for detection models that can robustly handle Moiré distortions alongside other real-world challenges, such as compression, sharpening, and blurring. By introducing the DMF dataset, we aim to drive future research toward closing the gap between controlled experiments and practical deepfake detection. 
\end{abstract}

\input{tex_files/1_introduction}

\input{tex_files/2_related}
\input{tex_files/3_methodology}

\input{tex_files/4_experiment}

\input{tex_files/5_results}

\input{tex_files/6_discussion}

\input{tex_files/7_conclusion}

\bibliographystyle{unsrtnat}
\bibliography{references,FakeAVCeleb_ref}
\newpage

\newpage
 
\setcounter{page}{1}

\input{tex_files/appendix-new}

\end{document}

%% file: tex_files/1_introduction.tex

\section{Introduction}
\label{sec:intro}
\begin{wrapfigure}{r}{0.5\textwidth}  
\vspace{-10pt}
  \centering
  \includegraphics[trim={15pt 20pt 15pt 20pt},clip,width=1\linewidth]{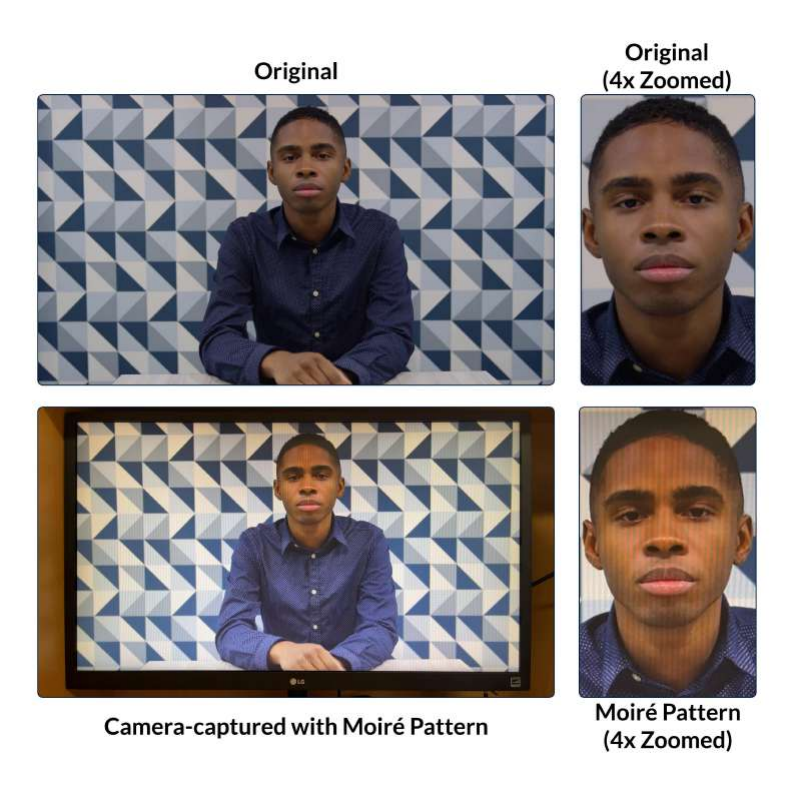}
    \caption{\textbf{Original vs. Moiré pattern}} 
    \label{fig:originalvsmoiré}
  \vspace{-10pt}
\end{wrapfigure}

The rise of deepfake technology has transformed how digital media can be manipulated, presenting a growing threat across the internet and social networking platforms. Deepfakes, which are artificially generated or altered videos that convincingly imitate real individuals, pose significant risks to privacy, security, and the spread of misinformation. The increasing ease with which deepfakes can be created exacerbates this issue~\cite{identityfraud}, as their realism often deceives the general public and sophisticated detection algorithms. 
Advances in deepfake generation techniques, such as those using Generative Adversarial Networks (GANs) \cite{GANs} and other deep learning models~\cite {faceswap, deepfakes}, including diffusion models~\cite{DiffusionSurvey}, have made detection an extremely challenging task~\cite{seow2022comprehensive, 41, 36, shahroz-acm} in real-world scenarios on the Internet. While efforts have been made to develop robust detection systems~\cite{wang2023gan, kumar2023gan, javed2022faceswap, jain2022dataless, t-gd, tariq2021one, 35, 28, 37, binh, Shahrozpaper, HasamADGD}, such algorithms are predominantly evaluated in controlled environments using benchmark datasets. However, real-world scenarios introduce various challenges, including environmental factors and media-sharing distortions, which can significantly impact detection accuracy~\cite{jeon2025seeing}. One of the most prominent challenges arises when deepfake content is viewed on screens and recorded using smartphone cameras. Although naive screen capture is available, Digital Rights Management (DRM) on many platforms often disables it. Accordingly, we focus on the prevalent smartphone screen-recapture scenario that users adopt for casual sharing or unauthorized reproduction. In practice, the same deepfake can exhibit drastically different visual characteristics when viewed directly on a screen versus when captured by a camera, adding an extra layer of complexity for detection systems (See~\autoref{fig:originalvsmoiré}). This common real-world scenario introduces visual artifacts known as Moiré patterns, which occur due to the interference between the pixel grid of the display and the camera sensor~\cite{He_2019_ICCV}. These Moiré patterns, often undetected by the human eye, severely disrupt deepfake detection algorithms, highlighting a critical gap between controlled environment performance and practical, real-world conditions.

In this paper, we investigate the impact of Moiré patterns and compression on deepfake detection systems across three scenarios: (i) Authentic Moiré patterns, (ii) Synthetic Moiré patterns, and (iii) Compression Attacks. Authentic Moiré patterns are introduced when users record content displayed on the screen with smartphones, degrading detection accuracy by distorting key visual features~\cite{yang2023doing}. For instance, a deepfake video of President Putin was shown nationwide on television declaring martial law, which was captured with a smartphone, clearly showing signs of a Moiré pattern~\cite{putintv1,putintv2} shared on X (formerly Twitter). Another example of a smartphone-captured deepfake on social media is creating false narratives by a broadcaster announcing President Macron rescheduling a visit due to an assassination attempt~\cite{fakenews}. Synthetic Moiré patterns, on the other hand, are deliberately generated either through pixel-level manipulation or by capturing screen-displayed content with controlled distortions, obscuring the artifacts that deepfake detectors rely on. Finally, Compression Attacks simulate real-world video uploads to social networking sites (SNS), where compression artifacts combine with Moiré patterns to impair detection systems further. 

\begin{wraptable}{r}{0.75\textwidth}
\centering
\vspace{-10pt}
\caption{\textbf{\DatasetName\ Details:} We selected a subset of videos from five famous deepfake datasets and manually captured them under various conditions, which resulted in a total playback time of 35.64 hours containing Moiré patterns across 12,832 videos (802$\times$4 (screens)$\times$2 (phones)$\times$2 (lightning conditions).}
\label{table:videos_count}
\vspace{-8pt}
\resizebox{1\linewidth}{!}{%
\begin{tabular}{lcccccc} 
\hline
\textsc{\textbf{Name}} & \begin{tabular}[c]{@{}c@{}}\textsc{\textbf{Real}}\\ \textsc{\textbf{Videos}}\\\textit{(People)}\end{tabular} & \begin{tabular}[c]{@{}c@{}}\textsc{\textbf{Fake}}\\\textsc{\textbf{Videos}}\\\textit{(People)}\end{tabular} & \begin{tabular}[c]{@{}c@{}}\textsc{\textbf{Videos}}\\\textsc{\textbf{from}}\\\textsc{\textbf{Dataset}}\end{tabular} & \begin{tabular}[c]{@{}c@{}}\textsc{\textbf{Duration}}\\\textsc{\textbf{per Video}}\\\textit{(secs)}\end{tabular} & \begin{tabular}[c]{@{}c@{}}\textsc{\textbf{Total}}\\\textsc{ \textbf{Videos}}\\\textsc{\textbf{Captured}}\end{tabular}&\begin{tabular}[c]{@{}c@{}}\textsc{\textbf{Captured}}\\\textsc{ \textbf{Duration}}\\\textit{(hours)}\end{tabular} \\ 
\hline
FF++~\cite{faceforensics++} & 200 & 200 & 400 & 10 &6400& 17.78 \\
DFD~\cite{DFD} & 28 & 28 & 56 & 10 & 896&2.49 \\
DFDC~\cite{DFDC} & 66 & 66 & 132 & 10 &2112& 5.87 \\
CelebDF~\cite{CelebDF} & 58 & 58 & 116 & 10 & 1856& 5.16 \\
UADFV~\cite{UADFV} & 49 & 49 & 98 & 10 & 1568& 4.36 \\ 
\hline
\textbf{Total} & \textbf{401} & \textbf{401} & \textbf{802} & - &12832& \textbf{35.64} \\
\hline
\end{tabular}
}
\vspace{-10pt}
\end{wraptable}
We address these challenges by introducing the DMF dataset, the first deepfake dataset to incorporate Moiré patterns into public deepfake datasets. It includes diverse videos captured from four screens under two lighting conditions with two smartphones, providing a realistic benchmark for evaluating the resilience of state-of-the-art deepfake detectors.~\autoref{table:videos_count} and~\autoref{table:video_specs} detail the dataset’s variations, and video-capturing specifications, reflecting practical challenges in deepfake detection. 
Unlike our previous studies~\cite{10647902} and~\cite{tariq2024beyond}, this work adds details not covered before. Specifically, we present a broader analysis of deepfake detectors, examine distortions beyond Moiré patterns, and evaluate compression effects and mitigation strategies in depth. These differences are discussed in Section~\ref{sec:related}, and the various angles used in our analysis are outlined in Appendix~\ref{sec:new_angles}. To assess the effectiveness of our dataset, we performed an extensive evaluation using 15 different deepfake generation methods. Additionally, we examine the impact of defense methods such as demoiréing techniques on the performance of these detection algorithms as a potential mitigation strategy. We summarize our main contributions as follows:
\begin{enumerate}[leftmargin=10pt]
    \item\textbf{Moiré Pattern Attacks, Scenarios, and Datasets:} We propose Authentic Moiré patterns and Synthetic Moiré patterns. Constructively, we developed the first Moiré Pattern-impacted deepfake datasets to evaluate both real-world cases. They are captured with four different computer screens using two different smartphone cameras under two different lighting conditions on videos from FaceForensics++ (FF++)~\cite{faceforensics++}, CelebDF~\cite{CelebDF}, the DeepFake Detection (DFD)~\cite{DFD}, the DeepFake Detection Challenge (DFDC)~\cite{DFDC} and UADFV~\cite{UADFV} dataset. DMF is released publicly under DOI-based restricted terms and conditions to support further research on Moiré-induced challenges in deepfake detection\footnote{\url{https://doi.org/10.7910/DVN/XYOSYW}}, and our evaluation codes are publicly available here\footnotemark \footnotetext{\url{https://github.com/Razaib-Tariq/DeepMoireFake}}.
    \item\textbf{Extensive Moiré Pattern Evaluation and Benchmarking:} We conducted an extensive empirical study using our DMF dataset and 15 detectors to determine how Moiré patterns from camera-captured deepfake videos on digital screens affect detector performance. This helps in understanding real-world application challenges and vulnerabilities with current detection methods.
    \item\textbf{Mitigation and Defense Approach with Demoiréing.} To remove the Moiré pattern from DMF videos, we propose the state-of-the-art defense methods and apply them using five image and two video demoiréing methods, evaluate these demoiréd videos using the identical 15 deepfake detectors, and present the effectiveness and implications of defense methods.
\end{enumerate}

\begin{wraptable}{r}{0.65\textwidth}
\centering
\vspace{-10pt}
\caption{Specifications and variations in the video-capturing setup.}
\label{table:video_specs}
\vspace{-8pt}
\resizebox{\linewidth}{!}{%
\begin{tabular}{l|l} 
\toprule
\textsc{\textbf{Name} (variations)}  & \textsc{\textbf{Details}} \\
 \hline
\textsc{Camera Angles} (4)  & Center, 45\textdegree\ left, 45\textdegree\ right and Handheld\\
\textsc{Lightning Conditions} (2)  & On and Off \\
\hline
\begin{tabular}[c]{@{}l@{}}\textsc{Screens} (4) \\ (60 (Hz) \end{tabular}  & \begin{tabular}[c]{@{}l@{}} LG (LED), BenQ (LED), Samsung (QHD-\\IPS), and Lenovo (UHD-IPS) \end{tabular} \\
\hline
\textsc{Phones} (2)  & \begin{tabular}[c]{@{}l@{}}iPhone 13 \\ Samsung S22 Plus \end{tabular}\\\hline
\textsc{Screen Resolution} (2)  & 1980x1080, 3840x2160 \\
\textsc{Capture Resolution} (1)  & 1980x1080 \\
\textsc{Frame Rate} (1) & 30 fps \\\hline
\textsc{Video Capture Apps} (2)  & \begin{tabular}[c]{@{}l@{}} OBS Studio, DroidCam (iOS) \\ IP Webcam (Android) \end{tabular}\\
\bottomrule
\end{tabular}
}
\vspace{-10pt}
\end{wraptable}
\noindent
Our real-world evaluation revealed that the presence of Moiré patterns caused an average performance decline of 10.7\% in deepfake detectors, with reductions reaching up to 25.4\% in extreme cases. Additionally, implementing demoiréing as a defense further decreased detection accuracy, with an average decline of 6.1\% and up to 17.2\% in severe cases. These findings highlight the need for further research to understand better the interaction between demoiréing techniques and deepfake detection algorithms.


%% file: tex_files/2_related.tex
\section{Related Work}
\textbf{\textsc{Deepfake Generation}.} Deepfake video generation leverages advanced deep learning techniques such as variational autoencoders (VAEs)~\cite{VAE}, generative adversarial networks (GANs)~\cite{GANs}, and diffusion models~\cite{diffusionpaper} to produce highly realistic manipulated videos. Common deepfake manipulations include face swapping, face reenactment, face attribute editing, and face synthesis~\cite{DeepfakeSurvey1, sok}.~\textit{Face swapping} replaces a target face with a source face while preserving attributes such as skin color, expressions, and the surrounding environment~\cite{faceswap}. \textit{Face reenactment} transfers expressions and movements from a source face to a target, retaining the target’s appearance and identity. This technique uses facial motion capture and deep learning to modify the target's movements based on a driving image, video, or pose~\cite{NeuralTexture,Face2Face,FaceRenactment}. ~\textit{Face attribute} editing alters specific facial features, such as age, expressions, or skin tone, using generative models among GANs and VAEs. It can focus on single attributes or edit multiple attributes simultaneously~\cite{FaceAttributeEditing, heo2025fakechain}. Finally,~\textit{face synthesis} employs GANs to create hyper-realistic human faces that do not exist. While it has applications in gaming and fashion, it also poses risks, such as enabling fake identities on social networks to spread misinformation~\cite{FaceSynthesis,ProGAN}.
    
\begin{wraptable}{r}{0.75\textwidth}
\vspace{-15pt}
\centering
\caption{A comparison of publicly available Deepfake datasets.}
\label{tab:quantitativecomparison}
\vspace{-8pt}
\resizebox{1\linewidth}{!}{%
\begin{tabular}{lrrrll} 
\toprule
\textsc{\textbf{Dataset}} & \multicolumn{1}{c}{\begin{tabular}[c]{@{}c@{}}\textsc{\textbf{Real}} \\\textsc{\textbf{Videos}}\end{tabular}} & \multicolumn{1}{c}{\begin{tabular}[c]{@{}c@{}}\textsc{\textbf{Fake}}\\\textsc{\textbf{Videos}}\end{tabular}} & \multicolumn{1}{c}{\begin{tabular}[c]{@{}c@{}}\textsc{\textbf{Total}}\\\textsc{\textbf{Videos}}\end{tabular}}     & \multicolumn{1}{l}{\begin{tabular}[l]{@{}l@{}}\textsc{\textbf{Encoding}}\\\textsc{\textbf{Artifacts}}\end{tabular}} &  \multicolumn{1}{l}{\begin{tabular}[l]{@{}l@{}}\textsc{\textbf{Acquisition}}\\\textsc{\textbf{Artifacts}}\end{tabular}} \\ 
\hline
UADFV~\cite{UADFV} & 49 & 49 & 98  & \textcolor{red}{\xmark} & \textcolor{red}{\xmark} \\
DeepfakeTIMIT~\cite{df-timit} & 640 & 320 & 960  & \textcolor{green}{\cmark} (Compress.) & \textcolor{red}{\xmark}\\
FF++~\cite{faceforensics++} & 1,000 & 4,000 & 5,000 &  \textcolor{green}{\cmark} (Compress.) & \textcolor{red}{\xmark}\\
CelebDF~\cite{CelebDF} & 590 & 5,639 & 6,229 &  \textcolor{red}{\xmark} & \textcolor{red}{\xmark}\\
DFD~\cite{DFD} & 363 & 3,000 & 3,363 &  \textcolor{red}{\xmark} & \textcolor{red}{\xmark}\\
DeeperForensics~\cite{jiang2020deeperforensics} & 50,000 & 10,000 & 60,000  & \textcolor{red}{\xmark}& \textcolor{red}{\xmark} \\
DFDC~\cite{dolhansky2020deepfake} & 23,654 & 104,500 & 128,154  & \textcolor{green}{\cmark} (Compress.) & \textcolor{green}{\cmark} (Lighting)\\
KoDF~\cite{kwon2021kodf} & 62,166 & 175,776 & 237,942  & \textcolor{red}{\xmark} & \textcolor{red}{\xmark}\\ 
FakeAVCeleb~\cite{FakeAVCeleb} & 500 & 19,500 & 20,000  & \textcolor{red}{\xmark} & \textcolor{red}{\xmark} \\
\hline
\textbf{Ours} & \textbf{401} & \textbf{401} &  \textbf{802} &  \textcolor{red}{\xmark} &\textcolor{green}{\cmark} \textbf{(Moiré)}\\
\bottomrule
\end{tabular}
}
\vspace{-10pt}
\end{wraptable}
\noindent\textbf{\textsc{Deepfake Detection}.}~As deepfake generation technology advances, effective detection methods become increasingly critical to prevent misuse. Deepfake detection relies on deep learning models that identify subtle artifacts often imperceptible to the human eye. Techniques include convolutional neural networks~\cite{Shahroz1, Shahroz2, Minha_FReTAL, Minha_CoReD, SAMGAN, SAMTAR}, temporal analysis~\cite{CLRNet}, frequency domain analysis~\cite{binh}, and attention mechanisms using transformers~\cite{coccomini2022combining,app12062953,M2TR}. Detection methods are typically developed and evaluated using datasets such as FaceForensics++ (FF++)~\cite{faceforensics++}, CelebDF~\cite{CelebDF}, UADFV~\cite{UADFV}, and FakeAVCeleb~\cite{FakeAVCeleb}. A comparative analysis of existing datasets is presented in~\autoref{tab:quantitativecomparison}. 

Our work evaluates deepfake detection systems under three distinct real-world scenarios. Captured Moiré Pattern Attack (CMPA) simulates authentic Moiré patterns generated when users record deepfake videos from screens, causing distortions that obscure critical visual features and degrade detection accuracy. Synthetic Moiré Pattern Attacks (SMPA) investigate the effects of artificial Moiré patterns using methods such as SMPA-MA~\cite{MoiréAttack} and SMPA-SPS~\cite{MoireAttack2}. Finally, Compression Attack (CA) explores how video compression artifacts from SNS uploads interact with Moiré patterns, further degrading deepfake detection performance.

\label{sec:related}

\noindent \textbf{\textsc{Moiré Patterns in Deepfake Detection. }}
While deepfake detection has seen significant advancements, the impact of Moiré patterns on detection performance remains an underexplored challenge. Prior studies, including our previous works~\cite{10647902} and~\cite{tariq2024beyond}, have investigated deepfake detection under various conditions; however, these efforts were limited in scope. The former employed a restricted set of detection methods, offering only preliminary insights on a constrained dataset, whereas the latter expanded the evaluation to more detectors but lacked real-world scenarios where Moiré patterns could be actively exploited. Furthermore, these studies did not systematically assess critical factors such as image distortion, compression effects, and mitigation strategies, essential for improving robustness in practical applications.

In contrast, our study extensively explores these missing aspects and introduces the novel DMF dataset to fill these gaps. The dataset comprises videos captured from four screens under two lighting conditions using two smartphones, enabling robust and practical evaluations. By capturing real-world interference patterns under controlled variations, the DMF dataset enables robust and practical evaluations, providing a more comprehensive benchmark for deepfake detection. Moreover, we analyze both authentic and synthetic Moiré patterns, extending prior research~\cite{10647902} and~\cite{tariq2024beyond}. Beyond this, we systematically investigate mitigation strategies, including demoiréing, denoising, and deblurring, revealing significant trade-offs where removing Moiré patterns may inadvertently reduce detection accuracy. Our empirical analysis spans 15 deepfake detectors and rigorously evaluates the interplay between Moiré patterns and compression artifacts, providing a robust real-world assessment.

%% file: tex_files/3_methodology.tex
\section{Dataset Collection and Generation}
\label{sec:methodology}

\begin{figure}[htbp]
    \centering
    \includegraphics[trim={20pt 55pt 18pt 5pt},clip,width=\linewidth]{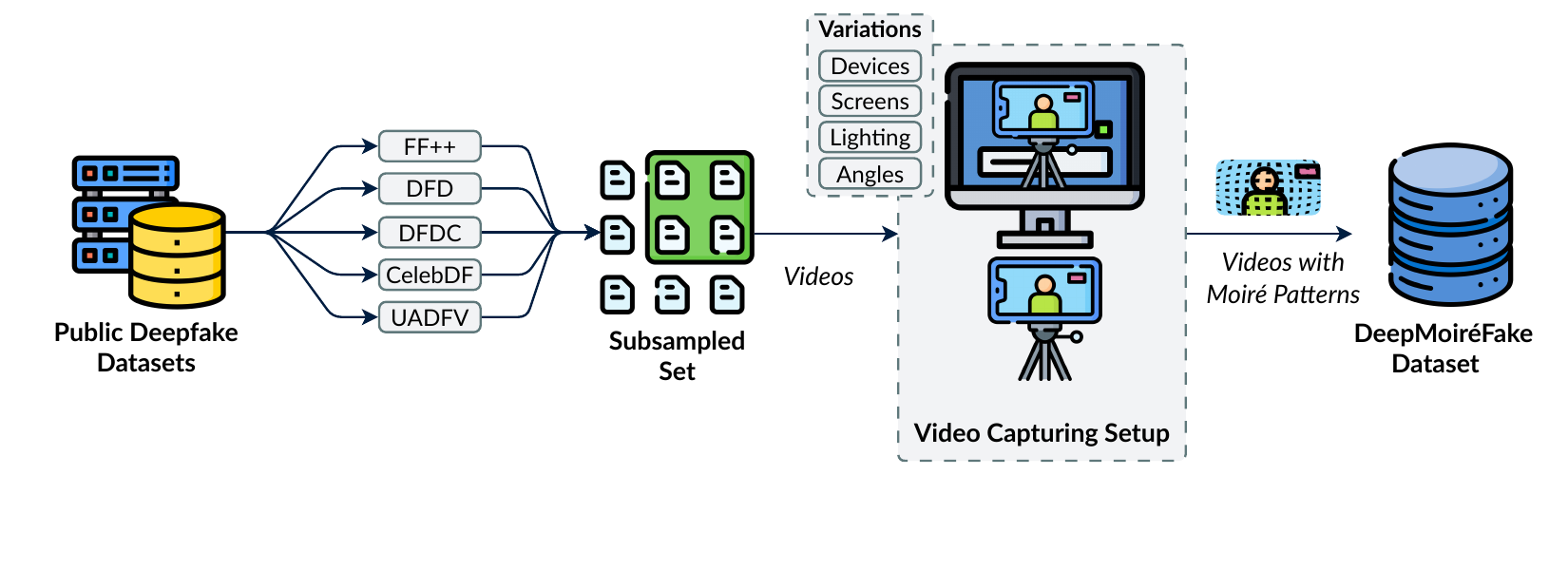}
    \caption{Our manual Moiré pattern collection pipeline.}
    \label{fig:Pipeline}
\end{figure}

\noindent\textbf{\textsc{Selection and Collection.}}
We selected five public deepfake datasets, UADFV~\cite{UADFV}, FaceForensics++~\cite{faceforensics++}, DFD~\cite{DFD}, DFDC~\cite{DFDC}, and CelebDF~\cite{CelebDF} as a representative set, covering a variety of settings detailed in~\autoref{tab:quantitativecomparison}. The overall process of generating our DeepMoiréFake dataset
from these source datasets is illustrated in~\autoref{fig:Pipeline}, which shows the pipeline of subsampling videos and capturing them under various screen and device conditions to induce Moiré patterns. UADFV~\cite{UADFV} dataset was created using the FakeApp tool and contains 49 real YouTube videos, each paired with a corresponding deepfake video. Videos are approximately 11 seconds long, with a resolution of 294×500 pixels. FaceForensics++ (FF++)~\cite{faceforensics++} dataset includes 1,000 real YouTube videos and 1,000 deepfake videos generated using four techniques: Deepfake~\cite{deepfakes}, Face2Face~\cite{Face2Face}, Faceswap~\cite{faceswap}, and NeuralTexture~\cite{NeuralTexture}. It provides 4,000 manipulated videos in three quality levels uncompressed (raw), lightly compressed (C23), and highly compressed (C40), enabling evaluations across varying compression levels. DFD~\cite{DFD} dataset by Google/Jigsaw consists of 3,068 deepfake videos generated from 363 original videos of 28 consented individuals representing diverse genders, ages, and ethnicities. DFDC~\cite{DFDC} dataset is the largest public faceswap dataset, featuring over 100,000 videos from 3,426 paid actors. Most videos are in 1080p resolution and include a mix of deepfake, GAN-based, and non-learned techniques, with an average of 14.4 videos per individual. CelebDF~\cite{CelebDF} dataset contains 590 real and 5,639 deepfake videos, sourced from over two million frames of YouTube interviews with 59 celebrities of diverse genders, ages, and ethnicities.

\textbf{\textsc{Dataset Subset Selection}.}~ We initially selected a subset of videos from five public deepfake datasets for dataset generation, ensuring a balanced representation across gender and ethnicity. Firstly, from the FaceForensics++ (FF++) dataset, we randomly chose 50 real and 50 fake videos from each of the four sub-datasets, totaling 400 videos. For the DFD, DFDC\footnote{For DFDC, we selected videos from the preview version, containing 5000 videos of 66 unique individuals.}, CelebDF, and UADFV datasets, we selected one real and one fake video for each unique individual, resulting in 56, 132, 116, and 98 videos, respectively (see \autoref{table:videos_count}), resulting in a carefully selected 802 videos from these datasets. Our main objective was to ensure a diverse representation from these deepfake datasets while maintaining a manageable number of manual videos for manual handling when creating Moiré pattern videos for our DMF dataset.

\noindent \textbf{\textsc{Generation of \DatasetName\ (DMF) Dataset}.}~The DMF dataset addresses the limitations of existing deepfake datasets by replicating real-world conditions, where videos are captured on monitor screens using mobile devices. This approach introduces distortions, such as Moiré patterns and screen-specific characteristics, which are captured through mobile device cameras, thereby facilitating the evaluation of deepfake detection methods in practical scenarios.
The dataset includes real and deepfake content recordings displayed on four monitors using two smartphone cameras (iPhone 13 and Samsung S22 Plus). Each smartphone was positioned on a stand 35 cm from the screen at various angles. Videos were captured under varied lighting conditions, with specifications detailed in \autoref{table:video_specs}. 
This carefully designed setup ensures a diverse range of screen types, lighting conditions, and device configurations, making the dataset valuable for advancing deepfake detection in real-world applications. During this stage, we ensured label accuracy via an automated process followed by manual verification and enforced consistency across screen camera setups. We also categorized each sample by device type, display screen, viewing angle, and lighting environment. These attributes are recorded in the dataset metadata to support reproducibility and downstream analysis.

\noindent \textbf{\textsc{Deepfake Detection}.} To evaluate deepfake detection performance across different datasets, we utilized 15 deepfake detectors. Specifically, we employed 10 image-based detectors, including SelfBlended~\cite{selfblended},
Rossler~\cite{faceforensics++}, ForgeryNet~\cite{he2021forgerynet}, Capsule-Forensics (Capsule)~\cite{nguyen2019capsule}, MAT~\cite{DBLP:journals/corr/abs-2103-02406}, CADDM~\cite{dong2023implicit}, CCViT~\cite{coccomini2022combining}, and ADD~\cite{binh} to assess detection results on the original dataset, Moiré pattern dataset, and demoiréd dataset. For the Rossler~\cite{faceforensics++}, we used pre-trained weights from three variations of the FaceForensics++ dataset: raw, C23, and C40, referred to as Rossler, Rossler C23, and Rossler C40, respectively. In our deepfake video detection experiments, we employed 5 detectors: AltFreezing~\cite{wang2023altfreezing}, FTCN~\cite{zheng2021exploring}, LRNet~\cite{sun2021improving}, and LipForensics~\cite{haliassos2021lips}. For LRNet, we used the BlazeFace (LRNet BF) and RetinaFace (LRNet RF) variants.

%% file: tex_files/4_experiment.tex
\section{Experimental Scenarios and Settings}
\label{experiment}
\noindent \textbf{Authentic Moiré Patterns or Captured Moiré Pattern Attack (CMPA).}~To simulate and evaluate a user-generated distortion, we propose a \textit{\textbf{Captured Moiré Pattern Attack (CMPA)}}. Moiré patterns, caused by interference between the pixel grids of the camera and monitor display~\cite{Schallek2009Stimulus-evoked}, are more intense with more significant resolution mismatches and are particularly pronounced on older technologies such as LCD, LED, and IPS displays compared to OLEDs. This effect diminishes as the distance between the camera and monitor increases~\cite{Saveljev:12}.  These distortions degrade deepfake detection accuracy by introducing artifacts that interfere with identifying critical visual features, especially over multiple frames.  Deepfakes were generated using a range of monitors with varying resolutions, such as 1080p LED, 1440p QHD IPS, and 4K UHD IPS displays, to simulate real-world conditions. Future datasets should include OLED displays and 8K monitors to improve the robustness of detection algorithms against these evolving challenges~\cite{DeepfakeBench}.

\begin{wrapfigure}{r}{0.5\textwidth}
    \centering
    \vspace{-10pt}
    \frame{\includegraphics[trim={2pt 2pt 440pt 2pt},clip,width=1\linewidth]{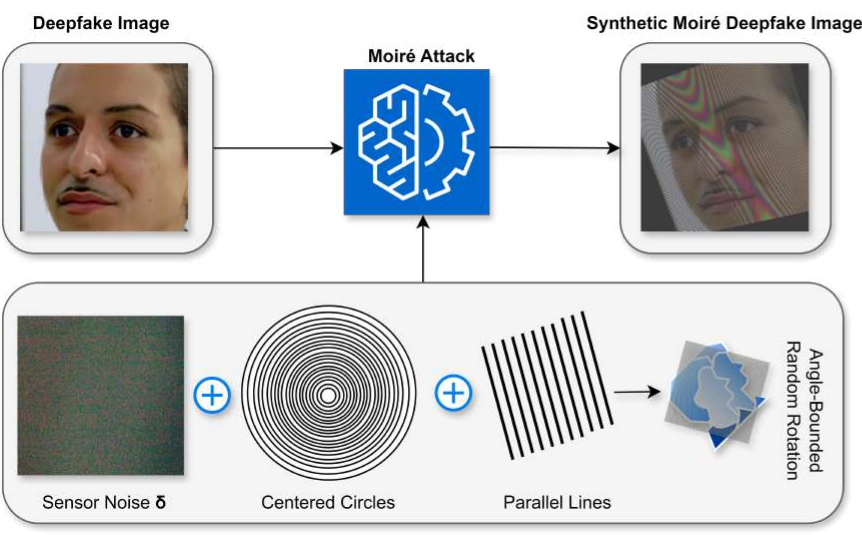}}
    \frame{\includegraphics[trim={438pt 2pt 8pt 2pt},clip,width=1\linewidth]{figures/synthetic-sides.pdf}}
    \caption{Synthetic Moiré Generation}
    \label{fig:synthetic}
    \vspace{-10pt}
\end{wrapfigure}

\noindent \textsc{\textbf{Synthetic Moiré Pattern Attacks (SMPA).}}~To evaluate the impact of interference patterns on deep learning models, we propose \textit{\textbf{Synthetic Moiré Pattern Attacks (SMPA)}}, which replicate noise artifacts commonly observed in screen recordings (see~\autoref{fig:synthetic}). These patterns degrade model performance by introducing complex distortions that are difficult to detect and eliminate.
The SMPA approach we used incorporates two methods: (1) \textbf{SMPA-MA}, which simulates real-world capture conditions by applying scaling, resampling, and random rotations to input images, as proposed by~\cite{MoiréAttack}, and (2) \textbf{SMPA-SPS}, which modulates parameters such as skew, contrast, and deviation while introducing non-linear distortions such as sine waves to replicate complex Moiré patterns~\cite{MoireAttack2}. By mimicking real-world Moiré artifacts, the SMPA demonstrates an efficient and effective adversarial approach, emphasizing the importance of designing robust detection algorithms to counter such attacks.

\noindent \textsc{\textbf{Compression Attack with CMPA and SMPA. }}
Uploading videos to Social Networking Sites (SNS) often introduces compression and quality degradation, adding new artifacts to the content. To replicate real-world scenarios, we propose the \textit{\textbf{Compression Attack (CA)}} on Moiré Patterns, which combines Moiré distortions with compression artifacts to simulate the impact of social media uploads. This approach leverages the widely used H.264 compression algorithm, adopted by platforms such as TikTok and YouTube~\cite{video_compression_social_media}, and standard techniques from FaceForensics++~\cite{faceforensics++}. By mirroring these real-world compression methods, we provide a realistic evaluation of how compression affects deepfake detection performance. We generated two compressed versions of our dataset, C23 and C40, to simulate the quality degradation caused by SNS uploads~\cite{moireonline1, moireonline2}. Compression reduces high-frequency information, introducing noise that interacts with existing Moiré patterns to create more complex distortions. These combined artifacts significantly degrade deepfake detection systems' performance, with increasing compression noise leading to further reductions in detection accuracy. This attack reveals a critical vulnerability in current detection systems, which often overlook the combined effects of Moiré patterns and compression artifacts. By exploiting compression, attackers can obscure signs of manipulation, enabling altered videos to evade detection on SNS platforms.

\textsc{\textbf{Preprocessing and Pretrained Weights.}} Our dataset comprised videos from various sources, including CelebDF~\cite{CelebDF}, DFD~\cite{DFD}, DFDC~\cite{DFDC}, and FaceForensics++~\cite{faceforensics++}. Each dataset was preprocessed according to the specific requirements of the respective deepfake detectors. For the demoiré experiments, an additional preprocessing step was applied to remove the Moiré pattern using state-of-the-art demoiréing methods~\cite{DMCNN, MBCNN, ESDNet, DDA}. We used the pretrained weights for each detector during the evaluation. We selected the top five performers on CMPA from the image detectors Rossler C23, MAT, CADDM, SelfBlended, and CCViT for SMPA, CA, demoiréing, fine-tuning, and retraining experiments. However, SelfBlended and CCViT did not exhibit any performance improvement during training, remaining at a static accuracy of 50\%. As a result, we excluded them from further analysis and focused on Rossler C23, MAT, and CADDM.

\noindent \textbf{\textsc{Moiré Removal methods.}} We employed state-of-the-art demoiréing methods such as DMCNN~\cite{DMCNN}, MBCNN~\cite{MBCNN}, and ESDNet~\cite{ESDNet} (under two settings) alongside DDA~\cite{DDA}, which is tailored for mobile devices, to remove the Moiré pattern from DMF videos: Firstly,
(i) \textbf{DMCNN} utilizes groups of convolutional layers for downsampling and deconvolutional layers to restore resolution. The final output image is produced by summing feature maps from all branches; Secondly, (ii) \textbf{MBCNN} features a learnable bandpass filter (LBF) for effective Moiré texture removal and employs a two-step tone mapping strategy for color restoration. This includes global tone mapping to correct color shifts and local fine-tuning for per-pixel accuracy. Thirdly, (iii) \textbf{ESDNet} integrates a Semantic-Aligned Scale-Aware Module (SAM) to handle scale variations of Moiré patterns. It enhances model effectiveness by extracting and dynamically fusing multi-scale features within the same semantic level, maintaining a lightweight network structure. and Lastly, (iv) \textbf{DDA} is optimized for real-time deployment on mobile devices. This method employs a parameter-shared supernet paradigm, ensuring resource efficiency without adding an extra parameter burden. It was selected because our data collection was performed using mobile devices.
The demoiréing experiments were conducted on image and video-based techniques, which are provided in the Appendix~\autoref{table:Demoiring} and~\autoref{table:videoDemoiring}.


\noindent
\textsc{\textbf{Metrics.}}~We evaluated the performance of deepfake detectors in our experiments using accuracy, AUC score, precision, recall, and F1-score. The main text reports the results based on the AUC score. For the CA, fine-tuning, and retraining settings, we report the best Accuracy. The results for the other metrics are available in the Appendix \ref{sec:Roc-curve}.


%% file: tex_files/5_results.tex
\section{Results} 
\label{sec:results}

\begin{wraptable}{r}{0.65\textwidth}
\centering
\vspace{-20pt}
 \caption{Performance on different playback screens.} 
 \label{table:Screens}
 \vspace{-8pt}
\resizebox{1\linewidth}{!}{%
\begin{tabular}{ll|c|cccc} 
\toprule
\multicolumn{2}{c|}{\multirow{2}{*}{\begin{tabular}[c]{@{}c@{}}\textsc{\textbf{Detectors}}\\(\textit{Type and Name})\end{tabular}}} & \multicolumn{1}{c|}{\multirow{2}{*}{\begin{tabular}[c]{@{}c@{}}\textsc{\textbf{Original}}\\\textsc{\textbf{Performance}}\end{tabular}}} & \multicolumn{4}{c}{\textbf{Videos captured from screens}} \\ 
\cline{4-7}
\multicolumn{2}{c|}{} &  & \textit{LG} & \textit{BenQ} & \textit{Lenovo} & \textit{Samsung} \\ 
\hline
 \multirow{5}{*}{\rotatebox[origin=c]{90}{\textsc{\textbf{Video}}}} & LRNet BF & \textit{61.7} & 54.9 & 55.3 & 55.9 & 53.2 \\
 & LRNet RF & \textit{62.2} & 58.8 & 60.5 & 58.7 & 58.8 \\
 & FTCN & \textit{90.2} & 65.9 & 65.3 & 70.6 & 68.9 \\
 & LipForensics & \textit{90.6} & 80.3 & 80.8 & \textbf{84.4} & 79.8 \\
 & AltFreezing & \textbf{\textit{92.5}} & \textbf{80.4} & \textbf{81.3} & 83.7 & \textbf{82.9} \\ 
\hline
 \multirow{10}{*}{\rotatebox[origin=c]{90}{\textsc{\textbf{Image}}}} & Rossler & \textit{67.7} & 56.2 & 54.5 & 59.4 & 56.9 \\
 & ADD & \textit{69.7} & 65.4 & 64.3 & 66.3 & 63.4 \\
 & Capsule & \textit{71.3} & 71.2 & 69.6 & 69.0 & 66.6 \\
 & ForgeryNet & \textit{76.9} & 61.5 & 61.8 & 66.5 & 63.6 \\
 & Rossler C40 & \textit{77.0} & 67.7 & 66.9 & 67.3 & 67.8 \\
 & Rossler C23 & \textit{86.5} & 68.6 & 67.4 & 74.5 & 70.9 \\
 & MAT & \textit{87.0} & 72.4 & 74.9 & 80.1 & 76.6 \\
 & CADDM & \textit{87.1} & 71.3 & 71.8 & 80.9 & 79.5 \\
 & SelfBlended & \textit{88.8} & 73.7 & 75.5 & 80.9 & 76.4 \\
 & CCViT & \textbf{\textit{95.0}} & \textbf{81.9} & \textbf{83.7} & \textbf{86.4} & \textbf{86.0} \\ 
\hline
\multicolumn{3}{c|}{\begin{tabular}[c]{@{}c@{}}\textbf{Avg. Performance loss }\\\textbf{(Moiré vs. Original)}\end{tabular}} & -11.6 & -11.4 & -8.0 & -10.2 \\
\bottomrule
\end{tabular}
}
\vspace{-12pt}
\end{wraptable}
\textsc{\textbf{CMPA -- Performance Under Various Playback Screen Settings. }}

In~\autoref{table:Screens}, the most significant visual Moiré artifacts were observed when videos were captured from the LG and BenQ screens, both of which use backlit LED technology with low pixel density and traditional RGB stripe subpixel layouts. These structural characteristics tend to amplify aliasing effects, particularly when captured through camera sensors, resulting in severe Moiré distortions. Correspondingly, the most substantial performance degradation in detection was also recorded for these two screens. This indicates that certain display technologies might amplify Moiré artifacts more than others. The variations in pixel arrangements, refresh rates, and anti-aliasing techniques across different screens likely contribute to the severity of these distortions. CCViT~\cite{coccomini2022combining} demonstrated the best detection performance across all screen environments, with an average AUC of 84.5\%. Meanwhile, Capsule~\cite{nguyen2019capsule} and LRNet showed robustness in this with the different capturing devices scenarios, and performance dropping by only 2-3 percentage points, for instance, on average, Rossler C23~\cite{faceforensics++} performance dropped to 16.1\%, whereas Capsule experienced only a 2.2\% drop. The performance from Capsule and LRNet is significantly low (around the mid-60s), making them impractical in the real world. Overall, we observed a similar trend in performance results across different screen configurations. In addition, we include performance results on videos captured at ±45° viewing angles in the Appendix~\autoref{table:45degrees}, further examining how angled perspectives affect detection robustness under Moiré interference.

\begin{figure*}[t]
    \centering
    \includegraphics[scale=1, clip,width=1\linewidth]{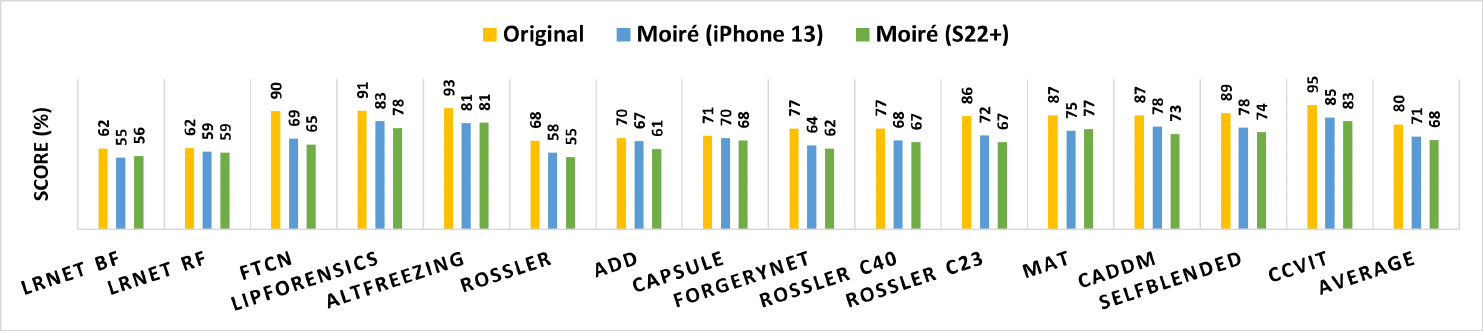}
     \caption{\textsc{\textbf{Different Capturing Devices:}} AUC performance of detectors dropped by 9.5 and 12.0 percentage points on average for videos with Moiré patterns captured by iPhone 13 and Samsung S22 Plus, with a maximum drop of 25.4 percentage points in the worst case.}
    \label{fig:CapturingDevices}
    \vspace{-15pt}
\end{figure*}

\textsc{\textbf{CMPA -- Performance with Different Capturing Devices. }} In \autoref{fig:CapturingDevices}, we illustrate the performance of detectors with original and Moiré pattern captured videos using iPhone and Samsung devices, showing a significant performance drop, highlighting the impact of Moiré artifacts on deepfake detection. The detection performance on videos captured using the Samsung S22 Plus was slightly worse on average than that captured with the iPhone. CCViT~\cite{coccomini2022combining} achieved the best performance across all scenarios, with 95\% on the original, 85\% on iPhone-captured, and 83\% on Samsung-captured images. The worst performance was observed with the LRNet models and Rossler model~\cite{faceforensics++}, where Rossler scored 68\% on the original, 58\% on iPhone-captured, and 55\% on Samsung-captured images, suggesting that the severity of Moiré interference may vary across different smartphone camera sensors and image processing pipelines. Overall, all detectors have a significant drop in performance, regardless of the capturing device used. This consistent degradation raises concerns about the generalizability of deepfake detectors in real-world scenarios where Moiré artifacts are commonly introduced during video playback or screen recording.

\begin{wraptable}{r}{0.30\textwidth}  
\centering
\vspace{-10pt}
\caption{Lighting Conditions}
\label{tab:lighting_conditions}
\resizebox{1\linewidth}{!}{%
\begin{tabular}{lcc}
\hline
\textsc{\textbf{Detectors}} & \textbf{On} & \textbf{Off} \\
\hline
LipForensics  & 81.3 & 78.7 \\
AltFreezing   & 82.1 & 80.6 \\
CCViT         & 84.5 & 83.5 \\
\hline
\end{tabular}%
}
\vspace{-5pt}
\end{wraptable}
\noindent
\textsc{\textbf{CMPA -- Performance Under Different Lighting Conditions. }} \autoref{tab:lighting_conditions} shows the performance of the (top-3) detectors when videos are captured by the camera on different screens that are exposed to different lighting conditions. 
In this scenario, we observed a very minimal performance change (around 1\%), which shows that the impact of the Moiré pattern remained the same irrespective of the lighting conditions (see Appendix \autoref{fig:lights} for results of all detectors).


\begin{wraptable}{r}{0.60\textwidth} 
\centering
\vspace{-10pt}
\caption{Comparison of deepfake detector performance in the presence and absence of Moiré Attacks.} 
 \label{table:Moire_Attack}
\resizebox{1\linewidth}{!}{
\begin{tabular}{l|c|ccc} 
\toprule
\multirow{2}{*}{\textsc{\textbf{Detectors}}}& \multirow{2}{*}{\begin{tabular}[c]{@{}c@{}}\textsc{\textbf{Without}}\\\textsc{\textbf{Attack}}\end{tabular}} & \multicolumn{3}{c}{\textsc{\textbf{Moiré Attack}}} \\ 
\cline{3-5} 
 &  & \textit{CMPA} & \textit{SMPA-MA} & \textit{SMPA-SPS} \\ 
\hline

Rossler C23 & \textbf{78.1} & \textbf{81.9} & 83.1 & 75.4 \\ 
\hline
MAT & 76.8 & 68.8 & 55.4 & 61.8 \\ 
\hline
CADDM & 73.0 & 73.1 & \textbf{86.8} & \textbf{80.7} \\ 
\bottomrule
\end{tabular}%
} 
\vspace{-10pt}
\end{wraptable}
\textsc{\textbf{SMPA -- Synthetic Moiré Pattern Attacks Results. }}
We examined two types of Synthetic Moiré Attacks. One is SMPA-MA, and the other is SMPA-SPS. Each Synthetic Moiré Attack framework is shown in (\autoref{fig:synthetic}). By using a subset of one variation of the camera-captured videos.
MAT shows the most severe performance degradation by synthetic Moiré attack among three detector models, with a performance drop of 21.4 percentage points (see \autoref{table:Moire_Attack}). Unlike MAT, which shows performance degradation after the Synthetic Moiré Attack, Rossler and CADDM show improved performance after SMPA-MA. 



\begin{table}[H]
\centering
\caption{CA baseline results under C23 and C40, evaluated with each detector’s pretrained weights.}
\label{table:baseline_C23_C40}
\resizebox{0.9\linewidth}{!}{
\begin{tabular}{l|cccc|cccc}

\toprule
\multirow{2}{*}{\textsc{\textbf{Detectors}}} &
  \multicolumn{4}{c|}{\textsc{\textbf{C23}}} &
  \multicolumn{4}{c}{\textsc{\textbf{C40}}} \\ \cline{2-9}
 & \textit{OG} & \textit{CMPA} & \textit{SMPA-MA} & \textit{SMPA-SPS} &
   \textit{OG} & \textit{CMPA} & \textit{SMPA-MA} & \textit{SMPA-SPS} \\ \hline
Rosseler C23 & \textbf{98.4} & \textbf{96.5} & \textbf{87.7} & 80.5 & \textbf{87.5} & \textbf{99.3} & 83.2 & \textbf{98.7} \\
MAT          & 86.7 & 66.1 & 55.4 & 56.5 & 75.3 & 66.5 & 52.2 & 60.8 \\
CADDM        & 97.7 & 96.4 & 86.7 & \textbf{90.3} & 80.1 & 99.0 & \textbf{84.8} & 96.8 \\ 
\bottomrule
\end{tabular}
}
\end{table}

\vspace{-15pt}

\begin{table}[H]
\centering
\caption{Performance of fine-tuned and retrained models on C23 and C40 compression attacks.} 
\label{table:C23_C40}
\resizebox{0.9\linewidth}{!}{
\begin{tabular}{c|l|cccc|cccc}
\toprule
 &
  \multirow{2}{*}{\textsc{\textbf{Detectors}}} &
  \multicolumn{4}{c|}{\textsc{\textbf{Fine-tune}}} &
  \multicolumn{4}{c}{\textsc{\textbf{Retrain}}} \\ \cline{3-10} 
\multicolumn{1}{l|}{} &
   &
  \textit{OG} &
  \textit{CMPA} &
  \textit{SMPA-MA} &
  \textit{SMPA-SPS} &
  \textit{OG} &
  \textit{CMPA} &
  \textit{SMPA-MA} &
  \textit{SMPA-SPS} \\ \hline
\multirow{3}{*}{\rotatebox[origin=c]{90}{\textsc{\textbf{C23}}}}
& Rossler C23 & 98.0          & \textbf{96.5} & 88.6          & {91.0}          & 97.8          & 96.1          & 87.8          & 91.5          \\
                     & MAT         & 99.2          & 91.8          & 94.8          & \textbf{98.5} & 99.3          & 92.1          & \textbf{95.8} & \textbf{97.5} \\
                     & CADDM       & \textbf{99.8} & 96.3          & \textbf{95.0} & {92.0}          & \textbf{99.4} & \textbf{96.2} & 90.0          & 91.8          \\ \hline
\multirow{3}{*}{\rotatebox[origin=c]{90}{\textsc{\textbf{C40}}}} & Rossler C23 & 82.5          & \textbf{99.6}          & 85.7          & {97.0}          & 86.7          & 99.5          & 85.7          & 95.7          \\
                     & MAT         & \textbf{{98.0}} & 90.6          & \textbf{94.4} & 97.9          & \textbf{99.1} & 84.2          & \textbf{94.4} & {98.1}          \\
                     & CADDM       & 90.9          & {99.3} & 90.8          & \textbf{99.2} & 96.0          & \textbf{99.7} & 90.71          & \textbf{99.3} \\ 
\bottomrule
\end{tabular}
}
\end{table}

\textsc{\textbf{Compression Attacks (CA). }} In \autoref{table:baseline_C23_C40} we observe that for the CA baseline, methods such as Rossler C23, MAT, and CADDM show distinct accuracy ranges,
with Rossler C23 and CADDM achieving around 80.1–99.0\% and MAT lagging behind at 55.2–86.7\%. Following fine-tuning and retraining, however, the overall trend is upward, as detailed in \autoref{table:C23_C40}. Most notably, the MAT model’s accuracy surged to 99.2\% in the best case, effectively closing the gap and becoming competitive with the other methods. This indicates that fine-tuning or retraining models on specific datasets or with targeted adjustments can enhance their ability to adapt to Moiré patterns and compression artifacts, ultimately improving detection accuracy. This improvement suggests that models benefit from being updated to handle new types of distortions or patterns, which may not have been fully accounted for in the original training process. 
\textsc{\textbf{Image Distortion Attacks. }}
We evaluated the impact of Gaussian blurring and sharpening on deepfake detection by applying these techniques to the original datasets. Gaussian blurring, implemented with OpenCV's GaussianBlur function~\cite{smooth_1}, smooths images by reducing fine details, while sharpening, using a high-pass filter via filter2D, enhances edges~\cite{sharp}. This systematic approach ensures consistent application, allowing direct comparison of detection performance. In Appendix~\autoref{table:GaussianBlurringSharpening}, we present AUC scores before and after applying these transformations.

\begin{wrapfigure}{r}{0.3\textwidth}  
  \centering
  \includegraphics[width=1\linewidth]{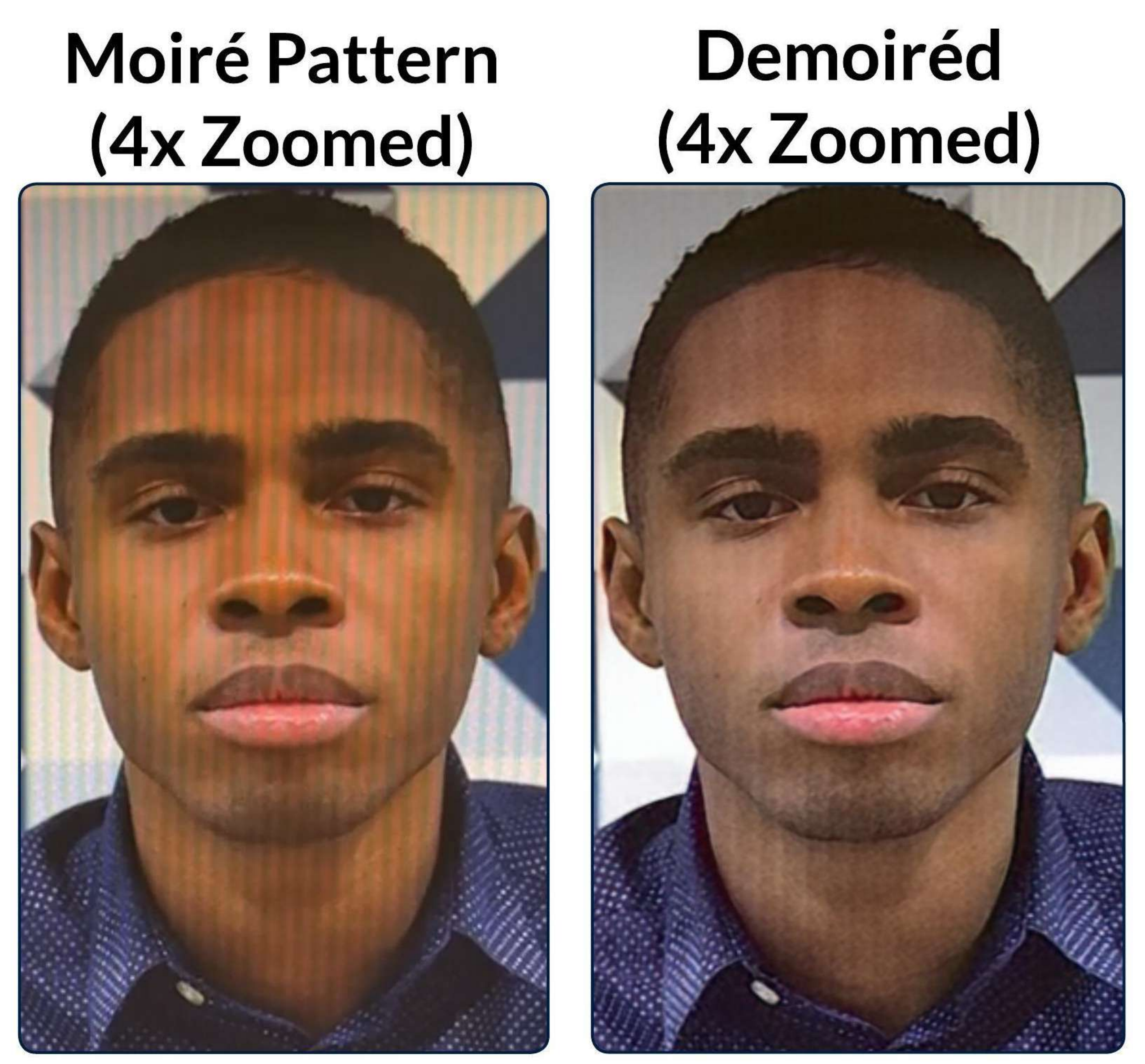}
  \caption{Moiré vs. Demoiréd}
    \label{fig:Demoiréd}
    \vspace{-10pt}
\end{wrapfigure}

\noindent \textsc{\textbf{Mitigation Strategies -- Performance After Demoiréing. }} The top-performing deepfake detectors across all demoiréing techniques were CCViT~\cite{coccomini2022combining}, CADDM~\cite{dong2023implicit},~and Rossler C23~\cite{faceforensics++}, consistently ranking 1st, 2nd, and 3rd, respectively (see Appendix~\autoref{table:Demoiring} for detailed results). CCViT achieved the highest average score of 79.2\%, maintaining superior performance across original, Moiré-affected, and demoiréd images. Among demoiréing methods, ESDNet~\cite{ESDNet}, trained on the FHDMi dataset, exhibited the lowest performance loss, indicating its effectiveness in mitigating Moiré-induced degradation. Conversely, DDA~\cite{DDA} demonstrated the highest performance loss, likely due to its optimization for mobile devices, which compromises its detection capabilities compared to other techniques. A significant finding from this experiment was that while demoiréing methods effectively removed most Moiré patterns from the images (see \autoref{fig:Demoiréd}), they also eliminated certain deepfake artifacts that detectors rely on for classification. As a result, performance on demoiréd images dropped more than on images with Moiré patterns. Specifically, Moiré patterns caused an average performance drop of 10.1 percentage points for detectors. In contrast, demoiréd images resulted in an average drop of 14.7 percentage points. This underscores the need for advanced mitigation strategies to address Moiré patterns without inadvertently removing critical deepfake artifacts, ensuring robust detection performance. We conducted additional experiments by processing Moiré videos using VD-Moiré~\cite{dai2022video} and FPANet~\cite{oh2025fpanet} demoiréing methods are outlined in \autoref{table:videoDemoiring} and denoising and deblurring from NAFNet~\cite{chen2022simple}, detailed results from these experiments are provided in (see Appendix \autoref{table:Denoising} and \autoref{table:Deblurring}).


\begin{table}[H]
\centering
\vspace{-10pt}
\caption{Overview of detectors with Fine-Tune and Retrain.} 
\label{table:fine-tune_retraining}
\resizebox{0.9\linewidth}{!}{ 
\begin{tabular}{l|cccc|cccc} 
\toprule
\multirow{2}{*}{\textsc{\textbf{Detectors}}} & \multicolumn{4}{c|}{\textsc{\textbf{Fine-tune}}} & \multicolumn{4}{c}{\textsc{\textbf{Retrain}}} \\ 
\cline{2-9} 
 & \textit{OG} & \textit{CMPA} & \textit{SMPA-MA} & \textit{SMPA-SPS} & \textit{OG} & \textit{CMPA} & \textit{SMPA-MA} & \textit{SMPA-SPS} \\ 
\hline

Rossler C23 & 77.0 & 80.6 & 94.4 & 81.1  & 87.9 & 84.7 & \textbf{94.9} & 79.5  \\ 
\hline
MAT & \textbf{94.5} & \textbf{85.4} & 70.3 & \textbf{95.6}  & \textbf{97.9} & \textbf{89.0} & 71.3 & \textbf{96.5}  \\ 
\hline
CADDM & 86.3 & 84.6 & \textbf{94.4} & 95.0  & 85.1 & 81.9 & 92.9 & 95.4  \\ 
\bottomrule
\end{tabular}%
} 
\vspace{-10pt}
\end{table}

\textsc{\textbf{Mitigation Strategies -- Performance After Fine-Tuning and Retraining. }} For fine-tuning, we utilized pretrained weights derived from the original dataset, which were also employed to assess the model's performance on the same data. The test dataset for fine-tuning and retraining comprises original data, captured Moiré data, and synthetic Moiré data. In the case of MAT, performance after retraining exhibited an improved score (see \autoref{table:fine-tune_retraining}). However, for CADDM, fine-tuning demonstrated superior performance compared to retraining.



\textsc{\textbf{Additional Analysis of Moiré Impact and Mitigation. }} We evaluate eight image detectors on original datasets (CelebDF, DFD, DFDC, FF++, and UADFV) and under the most severe Moiré distortion (LED screen) with multiple variations (light on/off, iPhone 13/Samsung S22+). We also assess the performance after demoiréing, denoising, and deblurring effects. The corresponding ROC curves are presented in (see appendix~\autoref{fig:enter-label}---\autoref{fig:label-h}), showing varying performance on image detectors and random guess prediction when impacted by Moiré patterns. Furthermore, our investigation extends to evaluating the impact of Moiré patterns on frequency analysis, Appendix~\autoref{fig:og_moire_frequency}, and deepfake generative models, with results provided in (see Appendix~\autoref{fig:gan_nongan} and~\autoref{fig:generative_methods}), with non-GAN and GAN showing distinct frequency patterns.

\textsc{\textbf{Remarks.}} These results demonstrate that just preprocessing methods (e.g., demoiréing) are insufficient to address the challenge posed by deepfake videos containing Moiré patterns or other artifacts. This highlights the need for more robust detection models capable of handling such distortions. In this context, our DMF dataset provides a valuable addition to public deepfake datasets for training these detectors. 


%% file: tex_files/6_discussion.tex
\section{Discussion}
\label{sec:discussion}




\noindent \textbf{Challenges in Data Collection.} 
Capturing Moiré patterns in real-world conditions required careful consideration of screen types, lighting variations, angles, and smartphone camera differences. Our dataset comprises 12,832 videos spanning 35.64 hours, sourced from CelebDF, DFD, DFDC, FF++, and UADFV, ensuring diverse representation. Differences in screen pixel structures influenced the intensity of Moiré artifacts. Additionally, smartphone cameras introduced variability in artifact appearance, further complicating the data collection process. These challenges highlight the complexity of generating a dataset that accurately represents Moiré-induced distortions in deepfake detection.

\noindent \textbf{Limitation and Future work.} 
While we acknowledge that real-world Moiré-inducing conditions span a wide range of factors, including variations in camera and display hardware, and dynamic motion, this work focuses on analyzing the impact of Moiré patterns on deepfake detection. Our experimental setup was intentionally designed to control these variables in a reproducible environment, enabling a focused investigation of Moiré-related effects. Broader scenarios involving diverse hardware configurations, motion artifacts, and platform-specific filters (e.g., beautification or AR effects on apps like TikTok and Instagram) remain essential directions for future work.

%% file: tex_files/7_conclusion.tex
\section{Conclusion}
\label{sec:conclusion}


In this paper, we investigated the impact of Moiré patterns on deepfake detection, exposing a significant vulnerability in current methods. Our experiments showed that both Authentic and Synthetic Moiré patterns can degrade detector performance, reducing accuracy by up to 25.4\%. This issue is further exacerbated by compression artifacts, where the combined effect leads to even greater performance deterioration. These findings highlight that existing models, often designed for clean, high-quality inputs, struggle with real-world artifacts introduced by screen captures and digital processing. While demoiréing techniques can mitigate these distortions, they may also inadvertently weaken detection performance. This underscores the need for more resilient deepfake detection systems capable of handling practical distortions like Moiré patterns and compression without significant accuracy loss.

\textsc{\textbf{Social Impact. }}  
Our work highlights the need for advanced deepfake detection to mitigate real-world artifacts. The dataset we share contains the real and deepfake videos captured with different mobile devices. The package also contains detailed documentation with all relevant metadata specified to users. We recommend using DMF as a training dataset to enhance detector robustness, aiding efforts to curb the spread of malicious deepfakes. To promote responsible, impactful use of the DMF dataset and to discourage misuse aimed at bypassing detectors, we provide access through a DOI-based request system. This process enhances security and ensures the dataset is used strictly for legitimate academic research.



\section*{Acknowledgement}
This work was partly supported by Institute for Information \& communication Technology Planning \& evaluation (IITP) grants funded by the Korean government MSIT: (RS-2022-II221199, RS-2022-II220688, RS-2019-II190421, RS-2023-00230337, RS-2024-00437849, RS-2021-II212068, and RS-2025-02263841). Also, this work was supported by the Cyber Investigation Support Technology Development Program (No.RS-2025-02304983) of the Korea Institute of Police Technology (KIPoT), funded by the Korean National Police Agency. Lastly, this work was supported by the National Research Foundation of Korea (NRF) grant funded by the Korea government (MSIT) (No. RS-2024-00356293).

%% file: tex_files/appendix-new.tex

\begin{center}
\rule{\linewidth}{2.5pt}\\[0.8em]
{\LARGE \bfseries Through the Lens: Benchmarking Deepfake Detectors\\
Against Moiré-Induced Distortions}\\[0.8em]
\rule{\linewidth}{0.5pt}\\[1.5em]

\begin{tabular}{@{}c@{\hspace{2em}}c@{\hspace{2em}}c@{\hspace{2em}}c@{}}
\textbf{Razaib Tariq}$^{*}$ & 
\textbf{Minji Heo}$^{*}$ & 
\textbf{Simon S. Woo}$^{\dagger}$ &
\textbf{Shahroz Tariq}$^{\dagger}$ \\[0.3em]

\multicolumn{3}{c}{Sungkyunkwan University, South Korea} &
CSIRO’s Data61, Australia \\[0.3em]

\multicolumn{3}{c}{\texttt{\{razaibtariq,minji.h0224,swoo\}@g.skku.edu}} &
\texttt{shahroz.tariq@data61.csiro.au}
\end{tabular}

\begin{minipage}{0.7\linewidth}
\centering
{\footnotesize ${}^{*}$Equal contribution.
${}^{\dagger}$Corresponding author.}
\end{minipage}

\end{center}

\rule{\linewidth}{1.5pt}\\[0.8em]
{\huge\textbf{Appendix  }}

\hspace{0.5pt}\textbf{Table of Contents}

\appendix

\hypersetup{linkcolor=black}
\startcontents[appendices] 
\printcontents[appendices]{}{0}{} 
\hypersetup{linkcolor=red} 
\renewcommand{\thesection}{\Alph{section}}

\noindent\rule{1\textwidth}{2pt}
\newpage

\section{Real World Examples of Moiré Pattern}
\begin{figure}[h]
  \centering
  \includegraphics[width=0.70\linewidth]{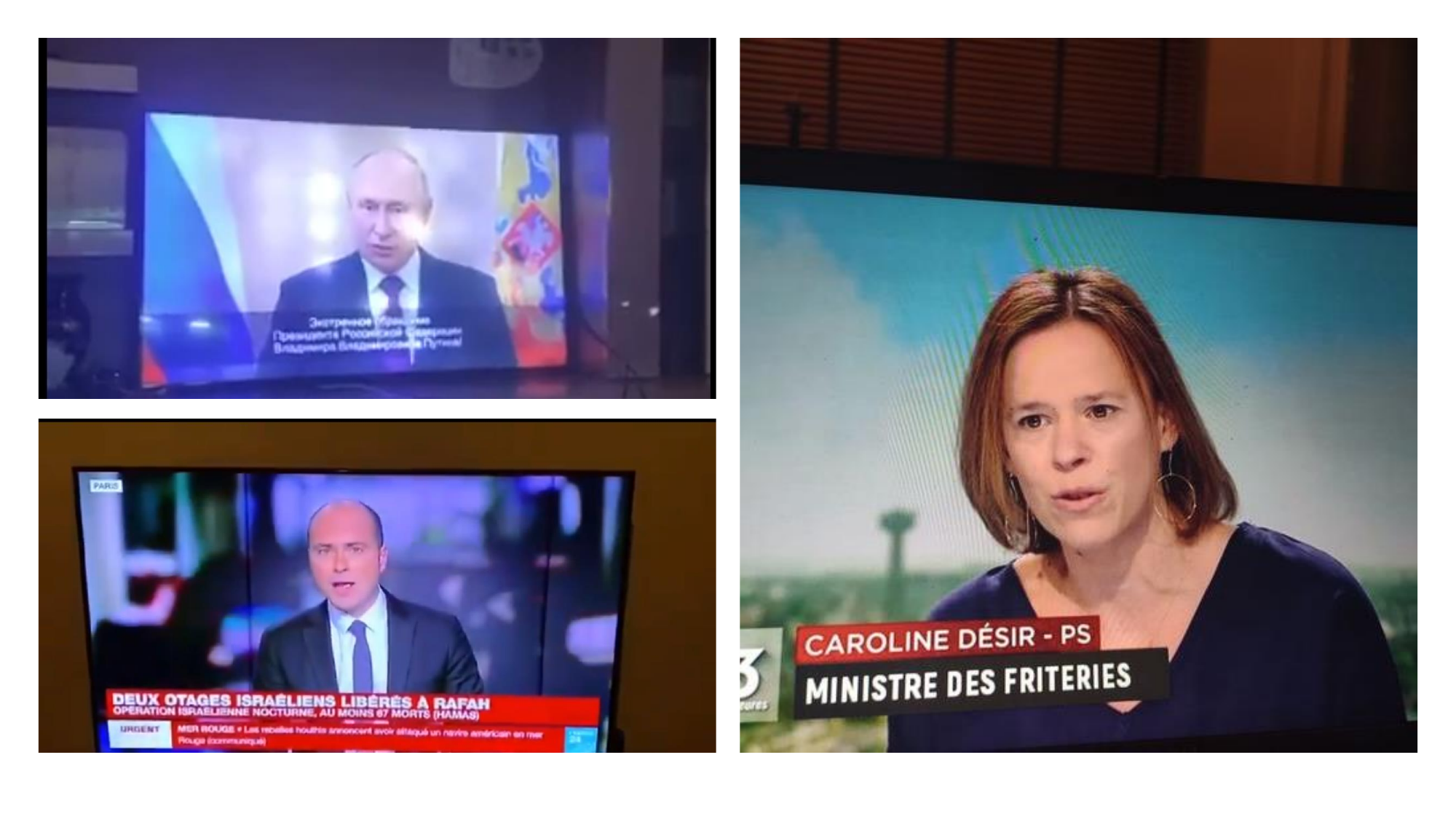}
  \caption{\textbf{\textsc{The examples on both sides show video on Live Broadcast captured by a smartphone camera introducing the Moiré Pattern.}}}
    \label{fig:moire-social}
\end{figure}

Deepfake videos are captured on the television screen using a smartphone device and distributed to different social networking services. The~\autoref{fig:moire-social} showcases how a novice tries to capture a deepfake when the platform will not allow the user to download it.

\section{Visualization of Dataset Samples}
Visualized examples in this section thoroughly compare the original images and those with Moiré patterns acquired on two separate monitors: LG (LED) and BenQ (LED), respectively, where the BenQ monitor tends to exhibit more pronounced Moiré patterns. The comparison involves displaying images before and after the implementation of ESDNet on different pre-trained weights, UHDM, and FHDMi for demoiréing, demonstrating how each method reduces Moiré patterns, and deblurring and denoising methods illustrate their efficacy in lowering Moiré patterns. This enables an evaluation of the effectiveness of each strategy on various displays.

\subsection{Examples of Original vs. Moiré Pattern}
\begin{figure}[h]
    \centering
    \frame{\includegraphics[keepaspectratio,width=0.32\textwidth,trim={23cm 19.5cm 19cm 1.5cm},clip]{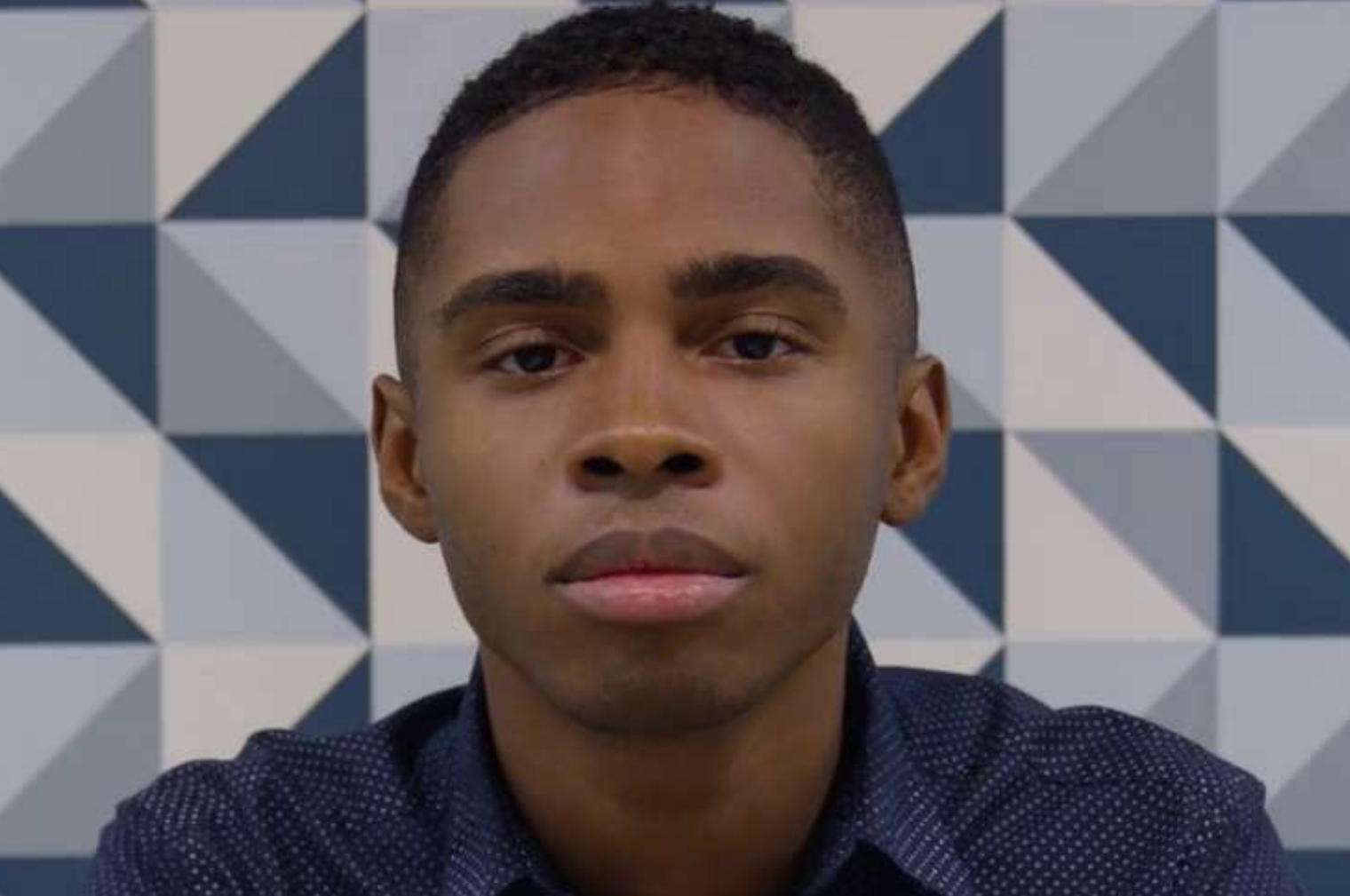}}
    \frame{\includegraphics[keepaspectratio,width=0.32\textwidth,trim={23cm 19.5cm 19cm 1.5cm},clip]{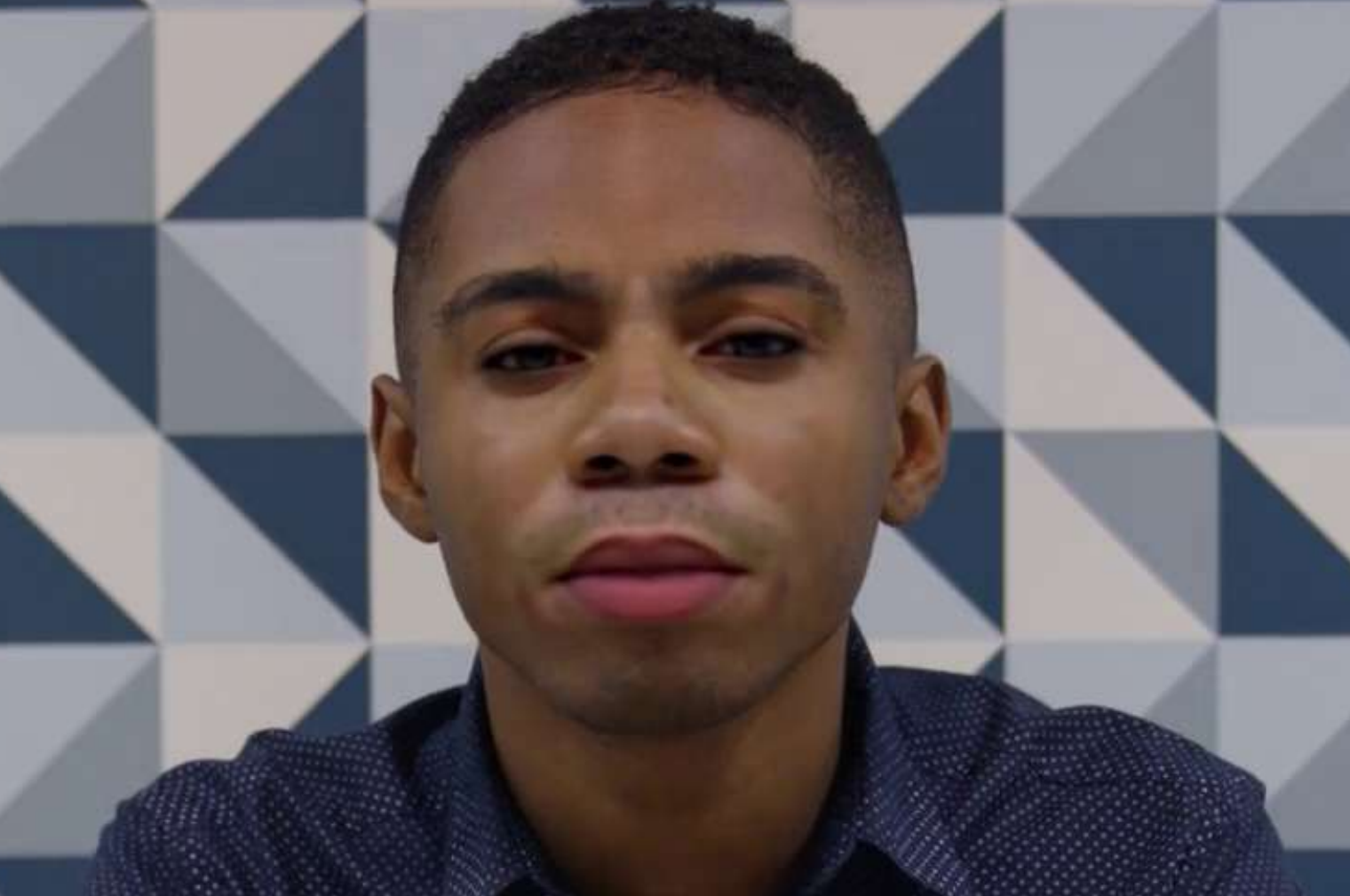}}\\\vspace{1.5pt}
    \frame{\includegraphics[keepaspectratio,width=0.32\textwidth,trim={24cm 19.5cm 19cm 1.5cm},clip]{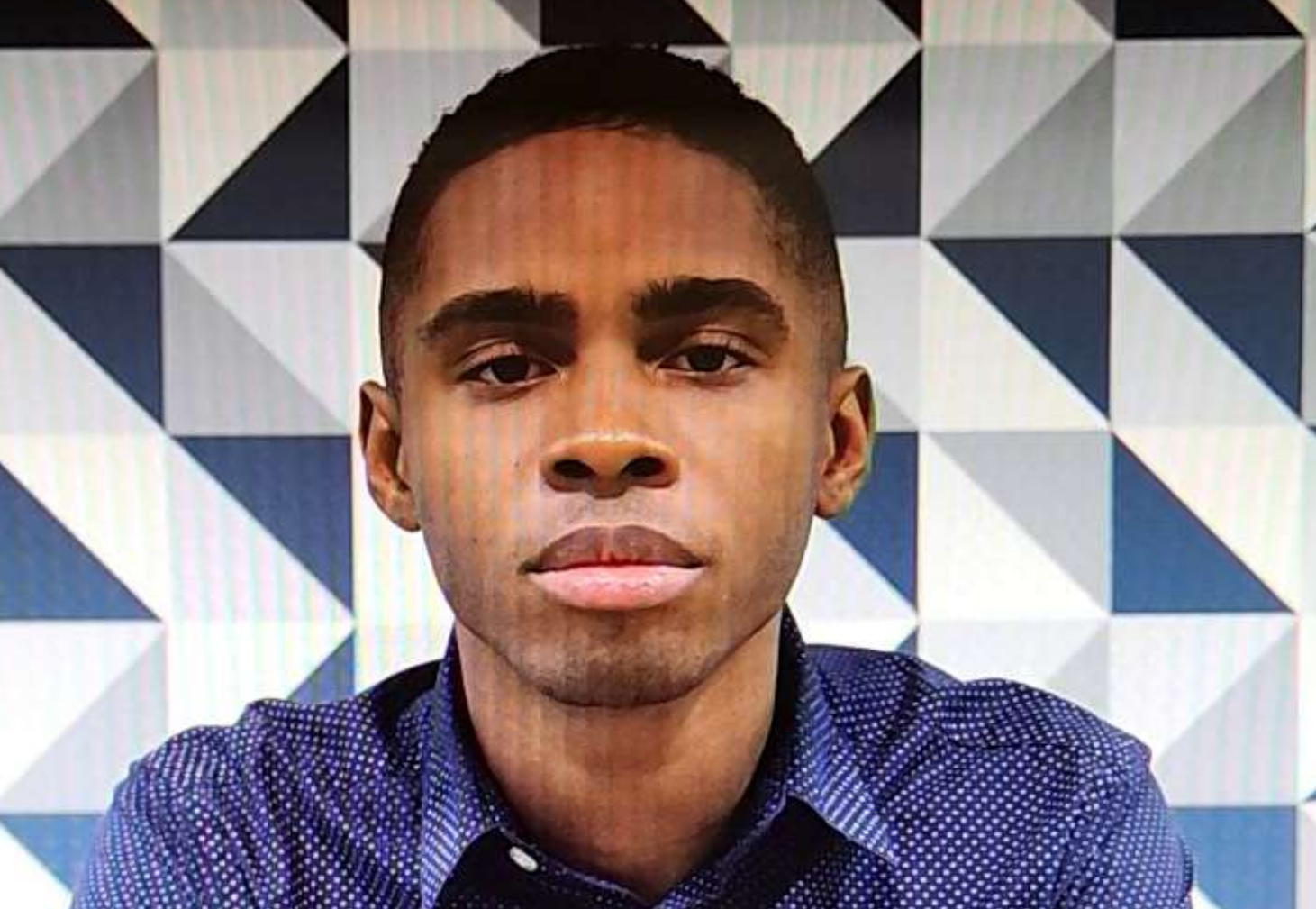}}
    \frame{\includegraphics[keepaspectratio,width=0.32\textwidth,trim={24cm 19.5cm 19cm 1.5cm},clip]{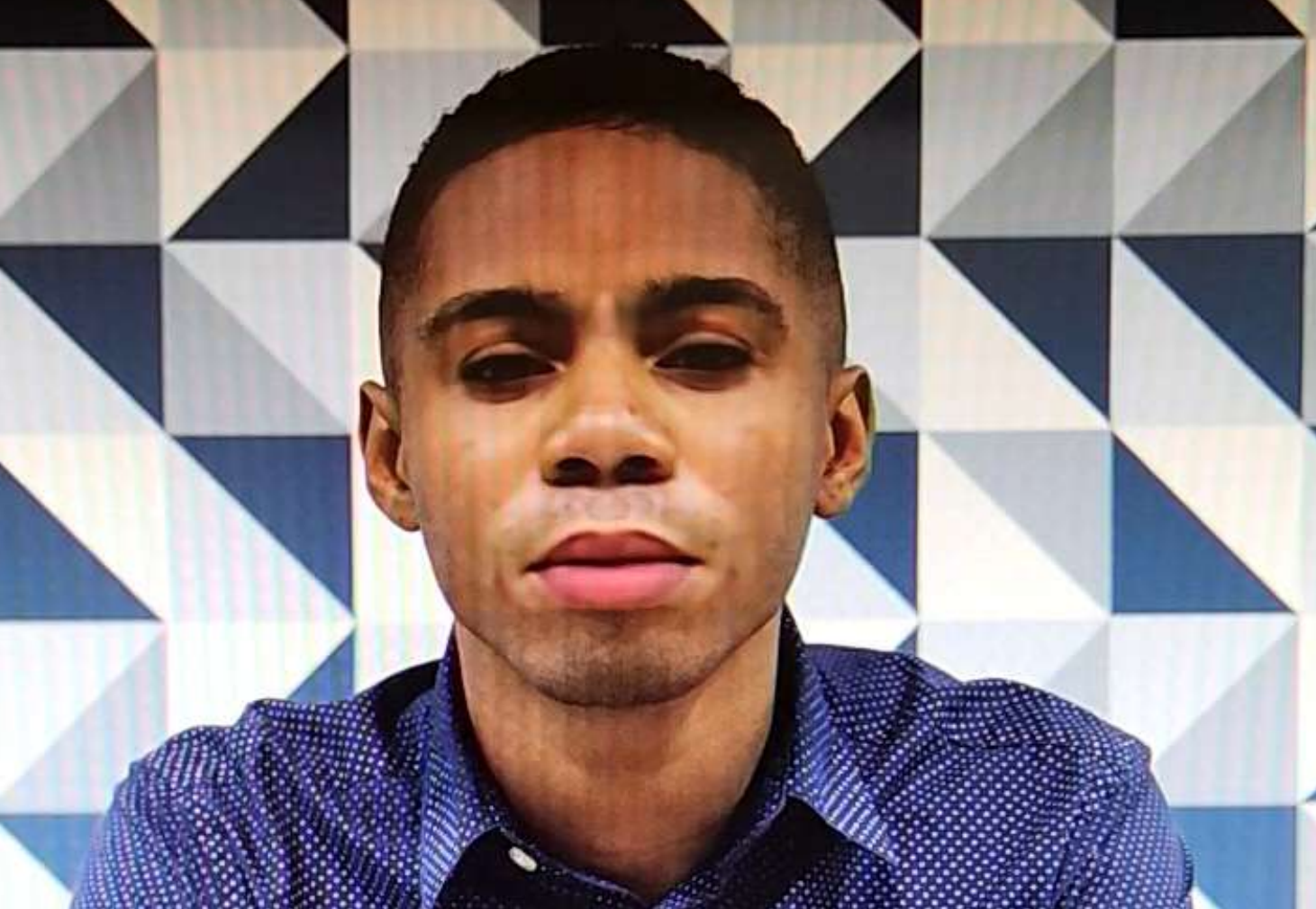}}
    \caption{\textbf{\textsc{Original vs. Moiré Pattern. }} \textbf{Top-left:} Real image without Moiré Pattern. \textbf{Top-right:} Deepfake without Moiré Pattern. \textbf{Bottom-left:} Real image with Moiré Pattern. \textbf{Bottom-right:} Deepfake with Moiré Pattern.}
\end{figure}
\newpage
\subsection{Examples of Moiré Patterns in Different Settings}
\begin{figure}[h]
    \centering
    \frame{\includegraphics[keepaspectratio,width=0.45\textwidth,trim={24cm 22cm 20cm 2cm},clip]{figures/appendix/Moire/Moire_Sam_LG_Fake_M.pdf}}
    \frame{\includegraphics[keepaspectratio,width=0.45\textwidth,trim={22cm 22cm 22cm 2cm},clip]{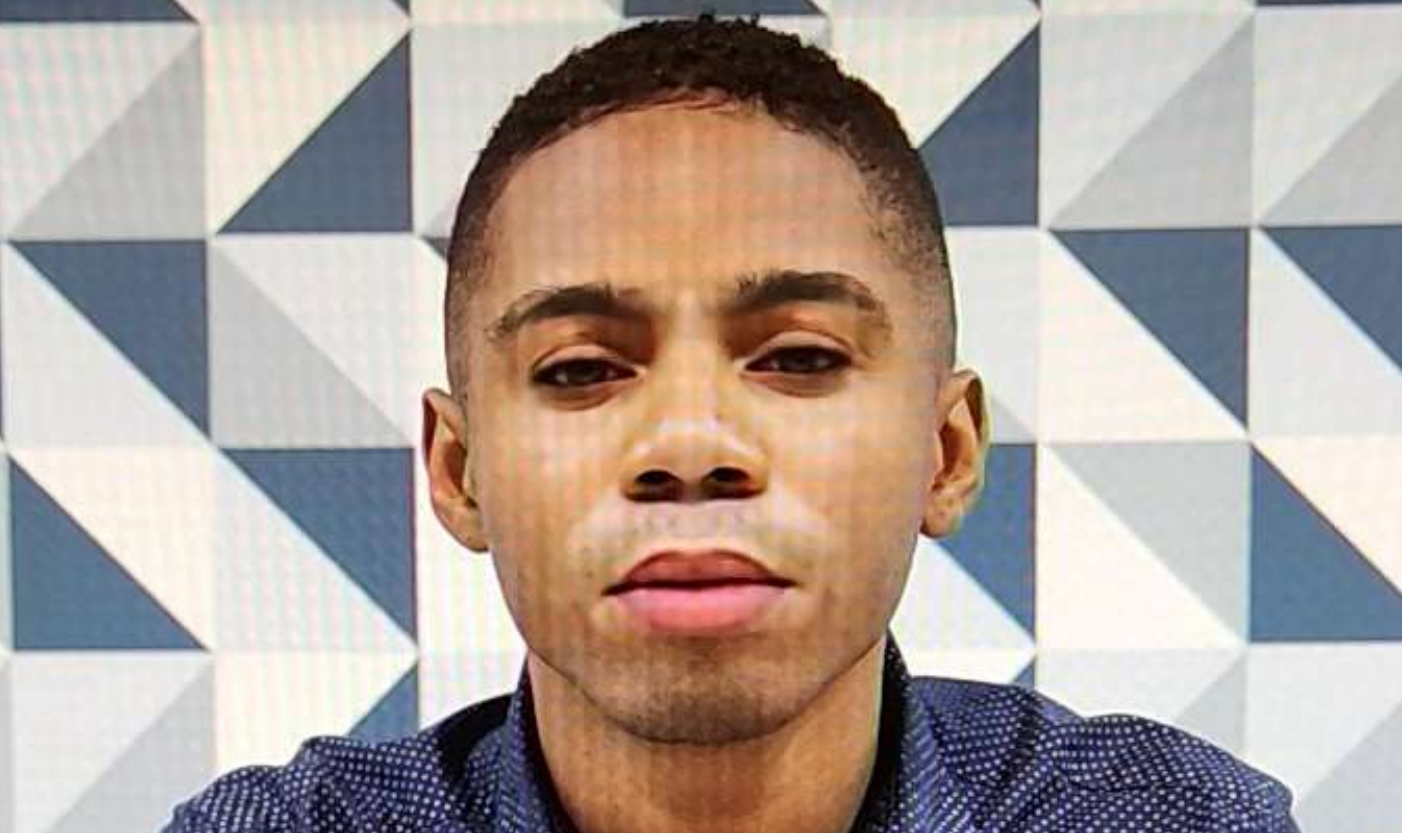}}\\\vspace{1.5pt}
    \frame{\includegraphics[keepaspectratio,width=0.45\textwidth,trim={24cm 21cm 20cm 3cm},clip]{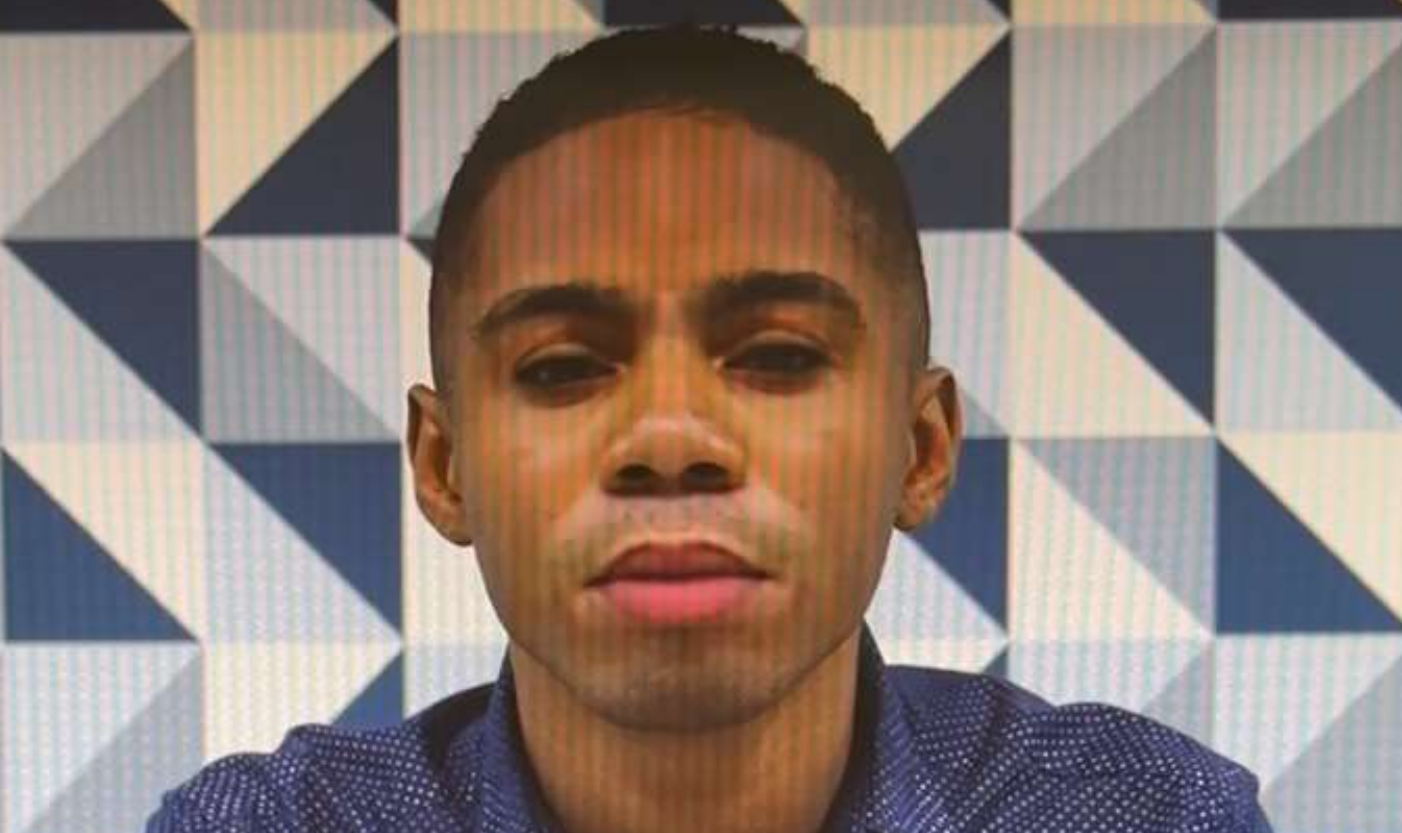}}
    \frame{\includegraphics[keepaspectratio,width=0.45\textwidth,trim={24cm 22cm 20cm 2cm},clip]{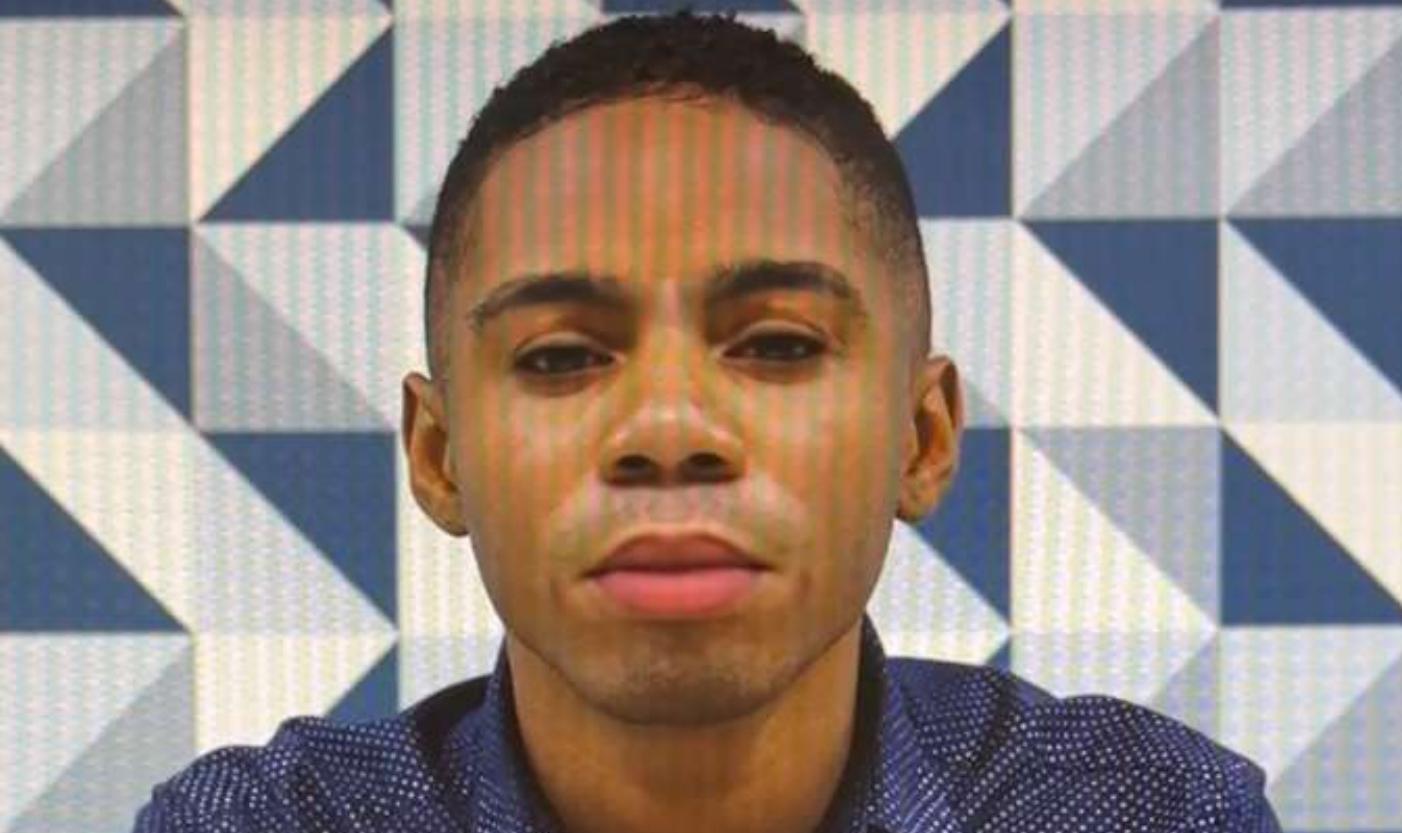}}
    \caption{\textbf{\textsc{Moiré Patterns on Different Monitors captured by Different Cameras. }}\textbf{Top-left:} On LG Monitor captured by Samsung S22 Plus. \textbf{Top-right:} On BenQ Monitor captured by Samsung S22 Plus. \textbf{Bottom-left:} On LG Monitor captured by iPhone 13. \textbf{Bottom-right:} On BenQ Monitor captured by iPhone 13.}
\end{figure}

\subsection{Examples of Demoiréing using ESDNet (UHDM) Methods}
\begin{figure}[h]
    \centering
    \frame{\includegraphics[keepaspectratio,width=0.45\textwidth,trim={24cm 22cm 20cm 2cm},clip]{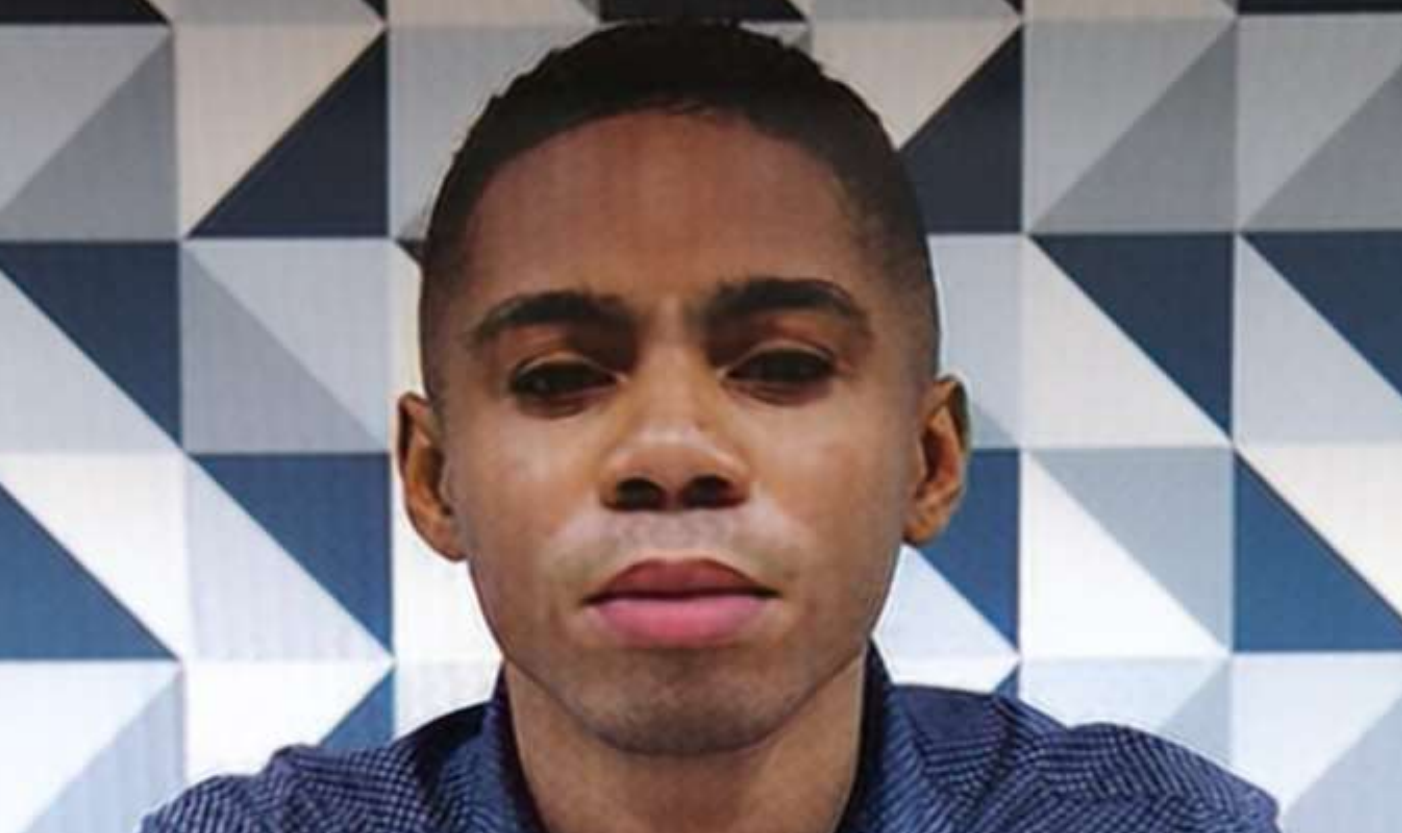}}
    \frame{\includegraphics[keepaspectratio,width=0.45\textwidth,trim={22cm 22cm 22cm 2cm},clip]{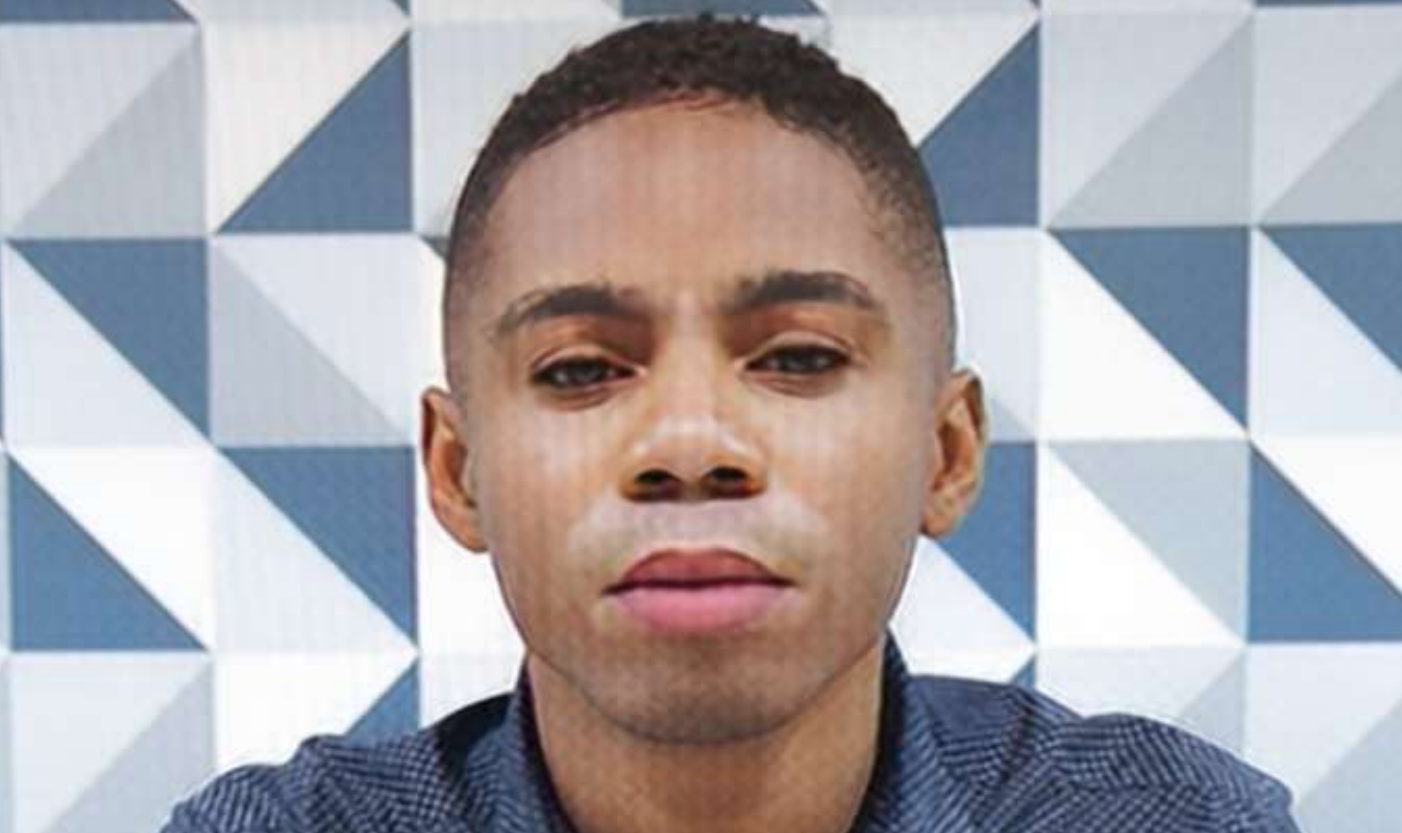}}
    \frame{\includegraphics[keepaspectratio,width=0.45\textwidth,trim={24cm 21cm 20cm 3cm},clip]{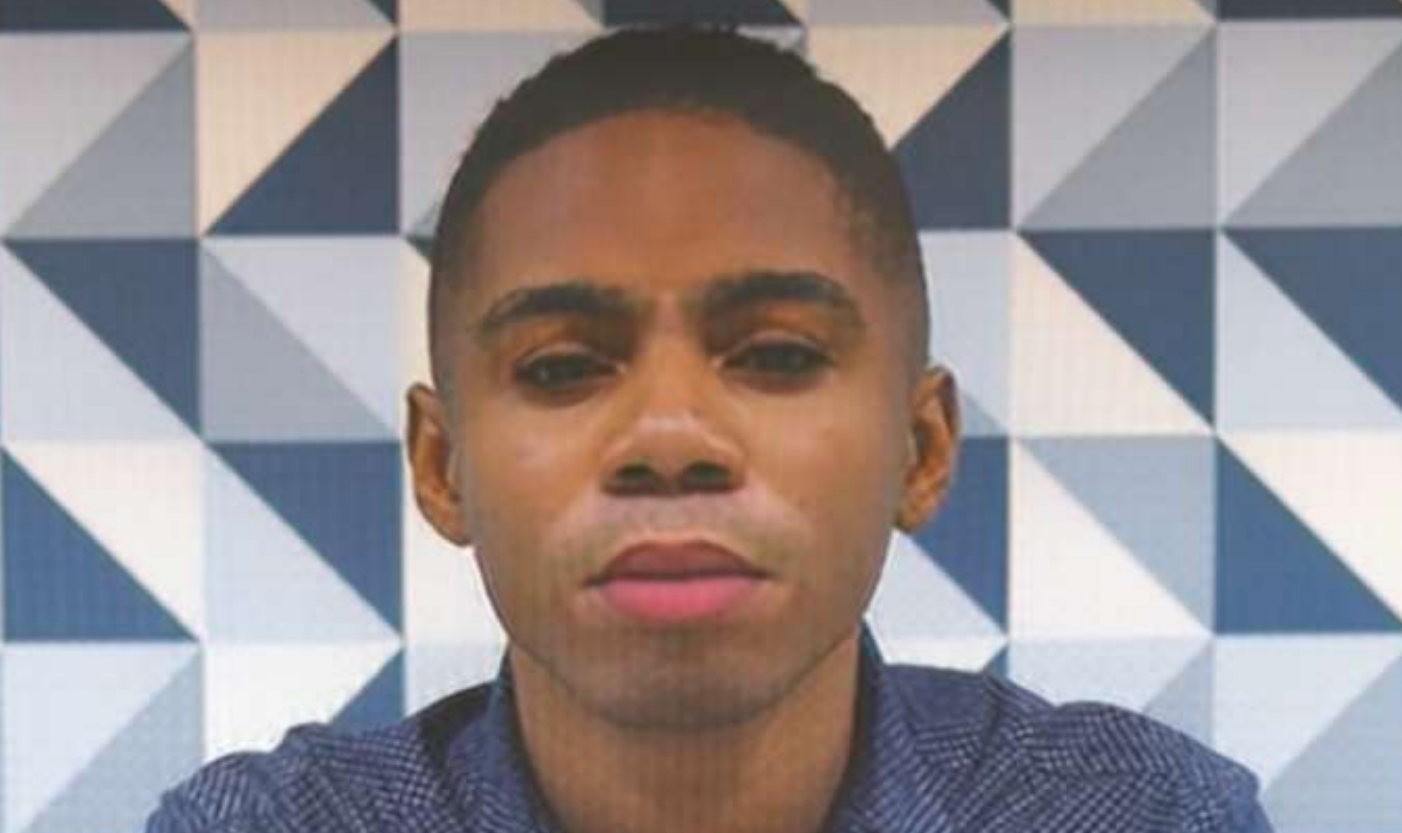}}
    \frame{\includegraphics[keepaspectratio,width=0.45\textwidth,trim={24cm 22cm 20cm 2cm},clip]{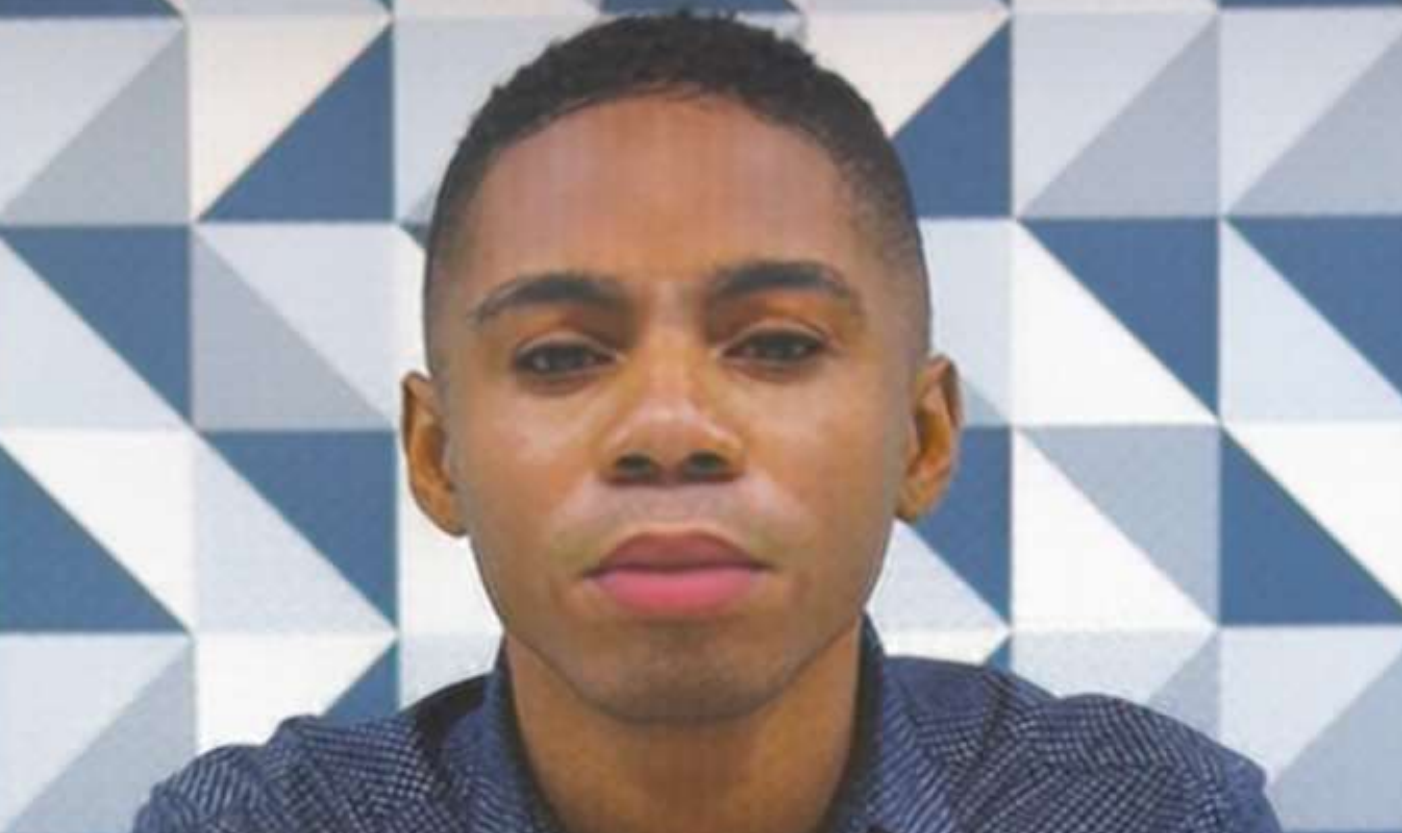}}
    \caption{\textbf{\textsc{Demoiré using the ESDNet (UHDM) Method. }}\textbf{Top-left:} On LG Monitor captured by Samsung S22 Plus. \textbf{Top-right:} On BenQ Monitor captured by Samsung S22 Plus. \textbf{Bottom-left:} On LG Monitor captured by iPhone 13. \textbf{Bottom-right:} On BenQ Monitor captured by iPhone 13.}
\end{figure}
\newpage

\subsection{Examples of Demoiréing using ESDNet (FHDMi) Methods}
\begin{figure}[h]
    \centering
    \frame{\includegraphics[keepaspectratio,width=0.45\textwidth,trim={24cm 22cm 20cm 2cm},clip]{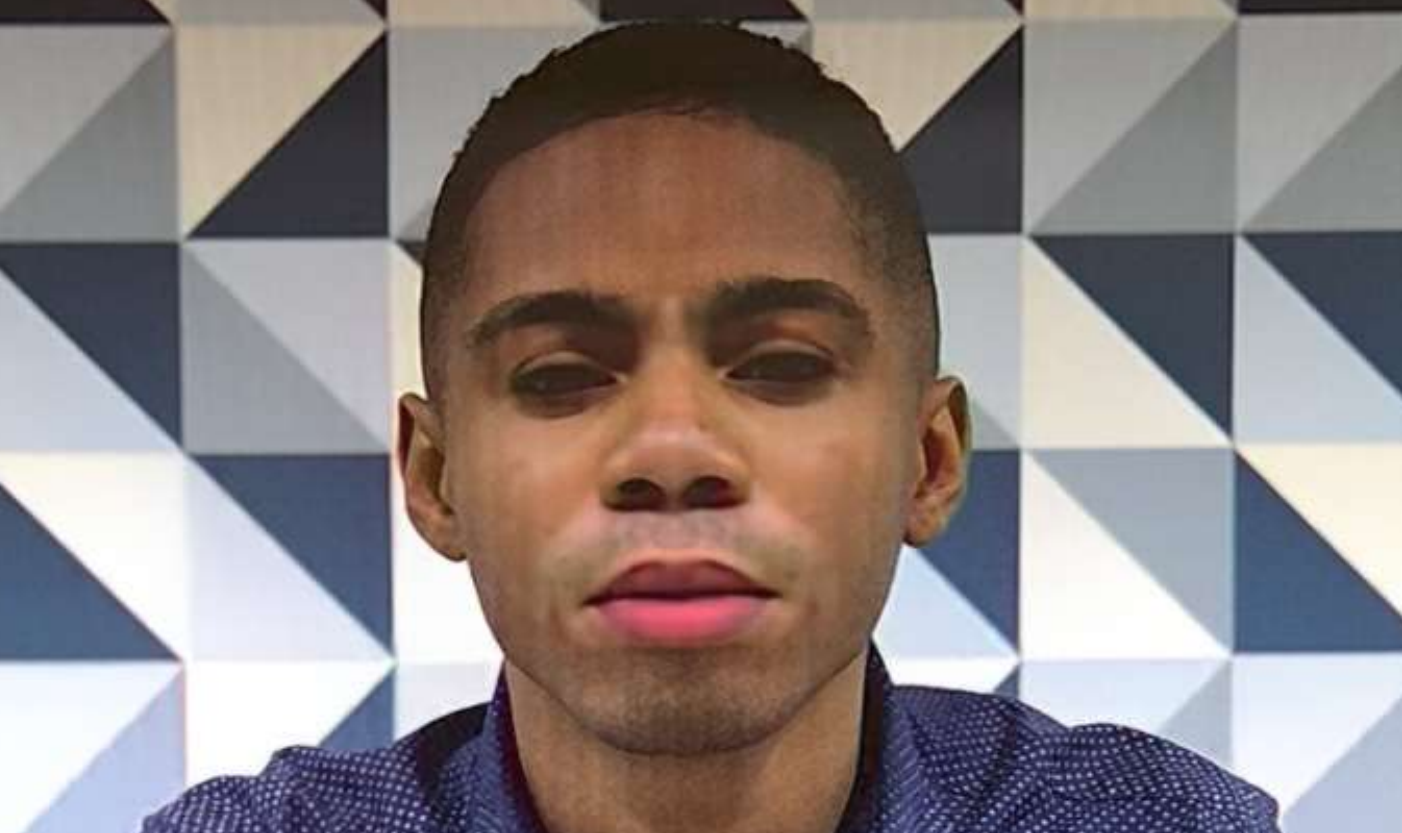}}
    \frame{\includegraphics[keepaspectratio,width=0.45\textwidth,trim={22cm 22cm 22cm 2cm},clip]{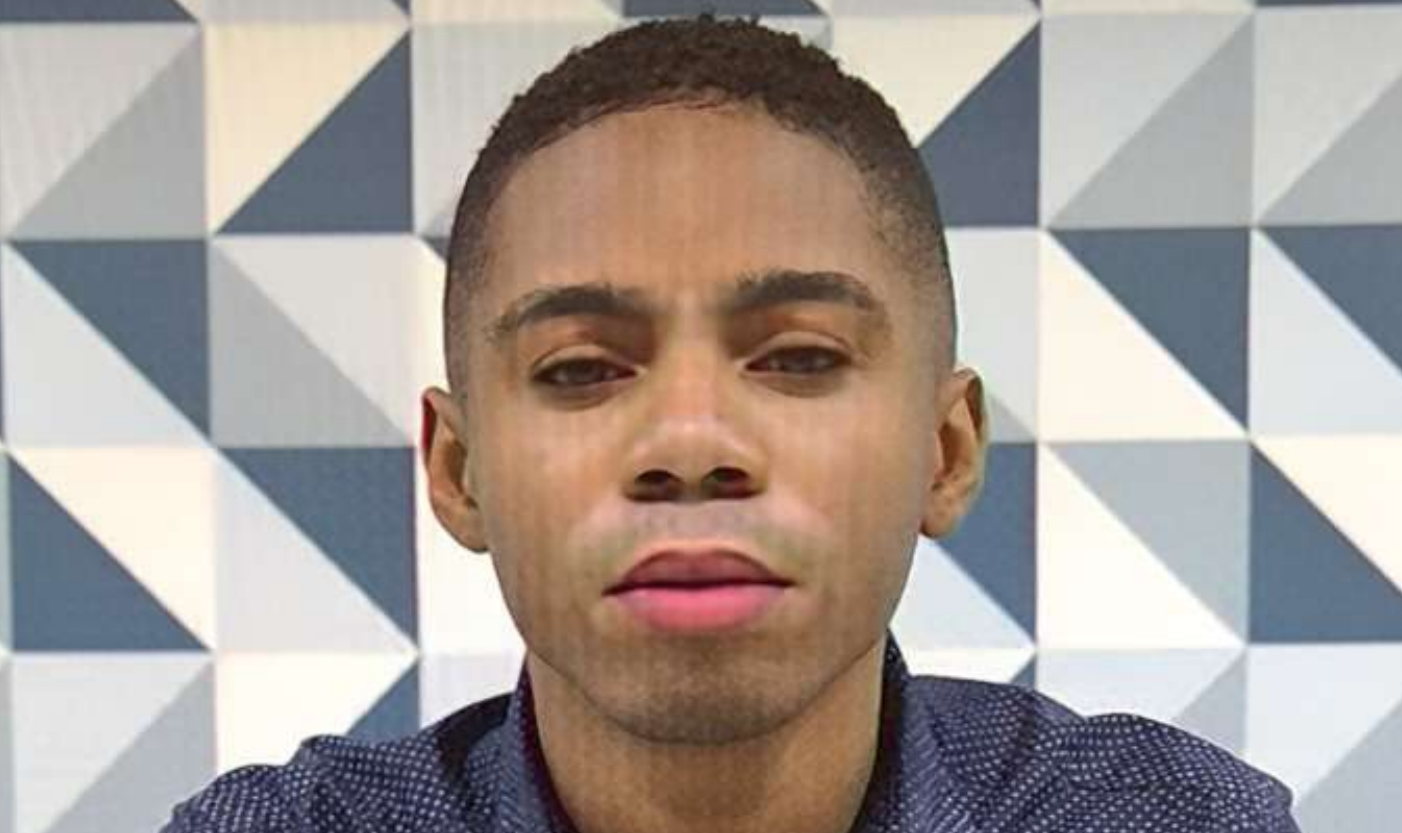}} 
    \frame{\includegraphics[keepaspectratio,width=0.45\textwidth,trim={24cm 21cm 20cm 3cm},clip]{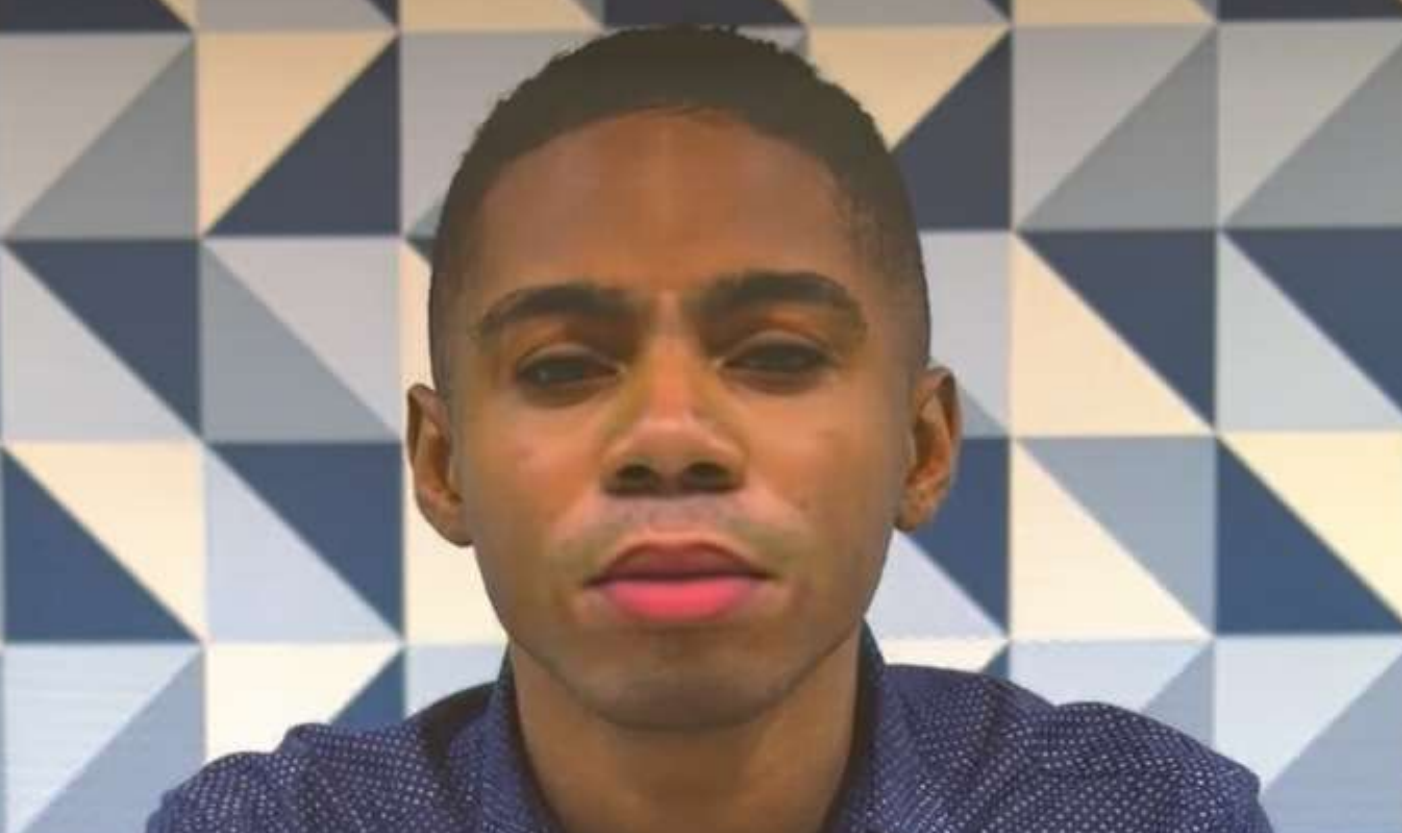}}
    \frame{\includegraphics[keepaspectratio,width=0.45\textwidth,trim={24cm 22cm 20cm 2cm},clip]{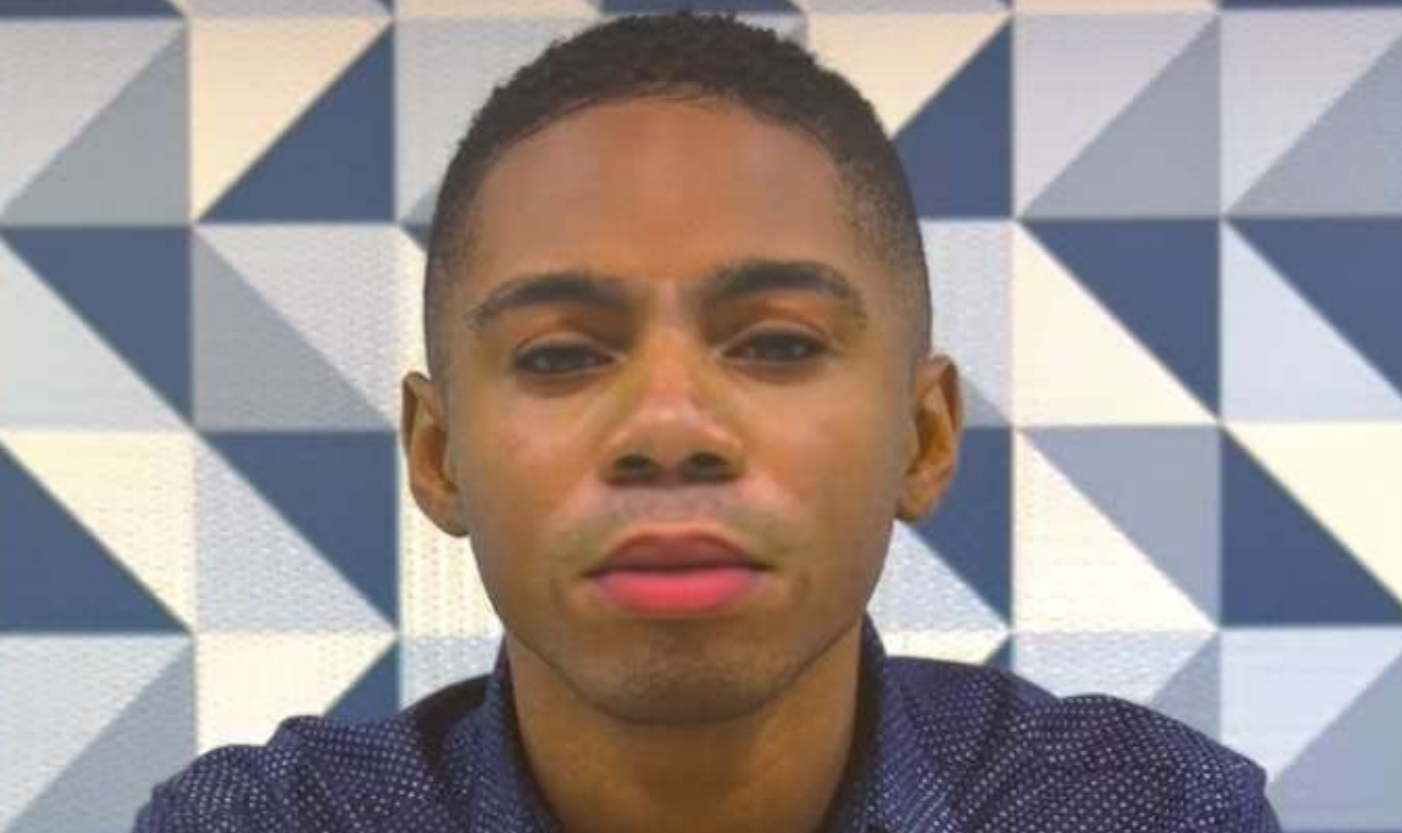}}
    \caption{\textbf{\textsc{Demoiré using the ESDNet (FHDMi) Method. }}\textbf{Top-left:} On LG Monitor captured by Samsung S22 Plus. \textbf{Top-right:} On BenQ Monitor captured by Samsung S22 Plus. \textbf{Bottom-left:} On LG Monitor captured by iPhone 13. \textbf{Bottom-right:} On BenQ Monitor captured by iPhone 13.}

\end{figure}

\subsection{Examples of Deblurring and Denoising Methods.}
\begin{figure}[ht!]
    \centering
    \centering
    \frame{\includegraphics[keepaspectratio,width=0.45\textwidth,trim={24cm 22cm 20cm 2cm},clip]{figures/appendix/Moire/Moire_i13_BenQ_Fake_M.pdf}}
    \frame{\includegraphics[keepaspectratio,width=0.45\textwidth,trim={22cm 22cm 22cm 2cm},clip]{figures/appendix/Demoire/Demoire_i13_FHDMI_BenQ_Fake_M.pdf}} 
    \frame{\includegraphics[keepaspectratio,width=0.45\textwidth,trim={24cm 21cm 20cm 3cm},clip]{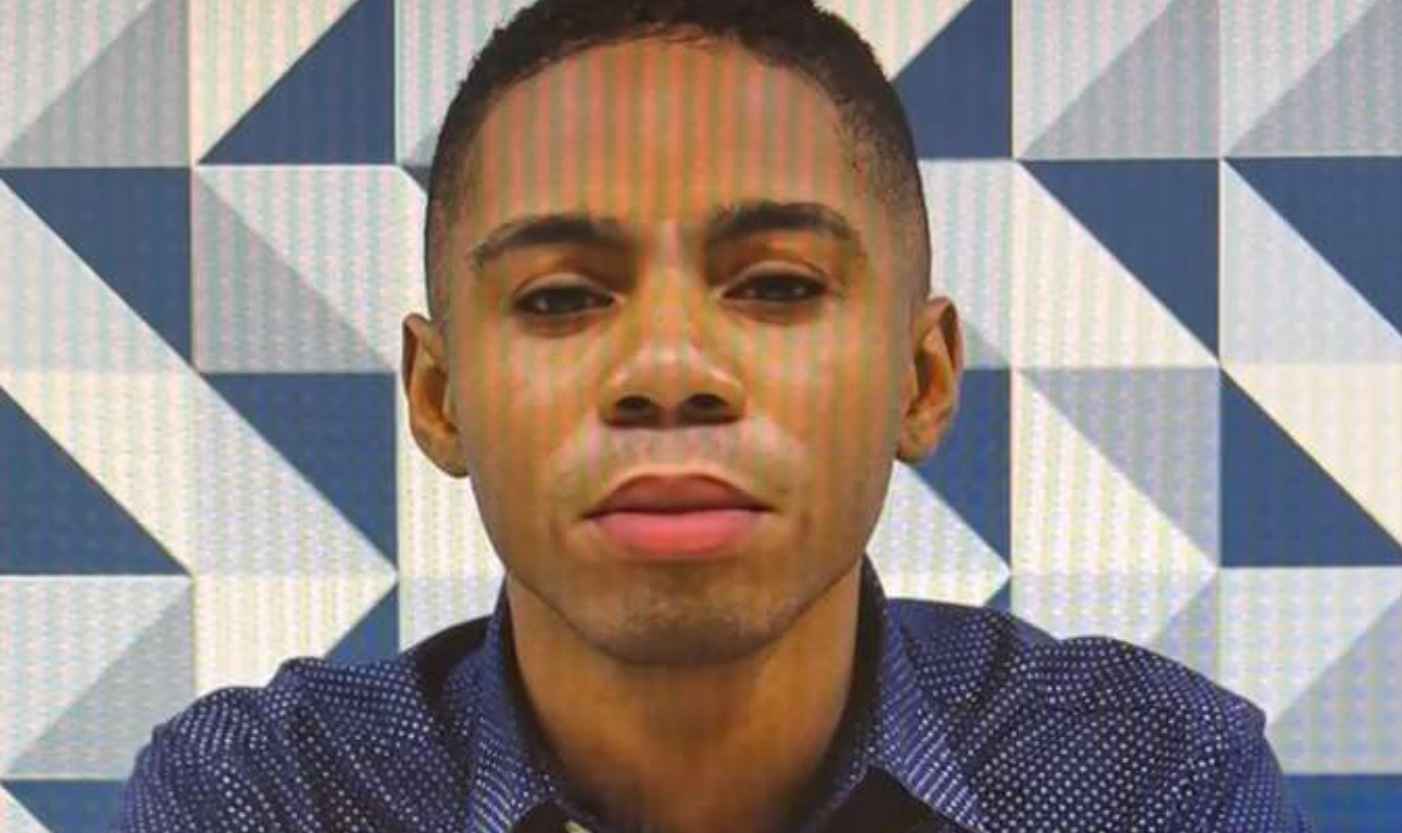}}
    \frame{\includegraphics[keepaspectratio,width=0.45\textwidth,trim={24cm 22cm 20cm 2cm},clip]{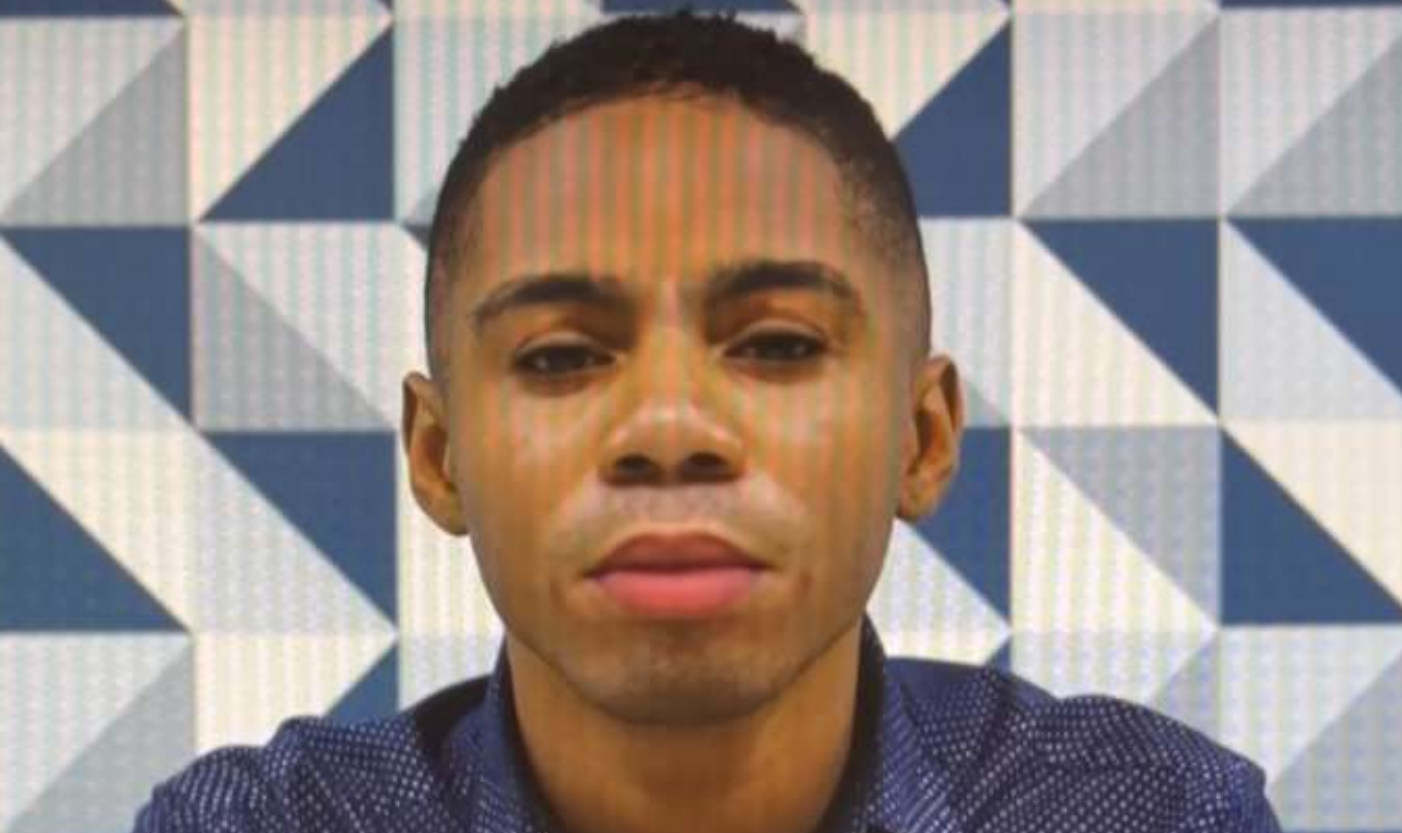}}
    \caption{\textbf{\textsc{Demoiréing vs. Deblurring vs. Denoising. }}\textbf{Top-left:} Deepfake with Moiré Pattern. \textbf{Top-right:} Moiré image processed with the demoiréing (FHDMi) method. \textbf{Bottom-left:} Moiré image processed with the deblurring (GoPro-64) method. \textbf{Bottom-right:} Moiré image processed with the Denoising (SSID-64) method.}
\end{figure}
\newpage
\subsection{Comparison of Original, Moiré Pattern, and Demoiréd Images.}

\begin{figure}[h]
    \centering
        \centering
        \includegraphics[scale=0.2, width=\textwidth]{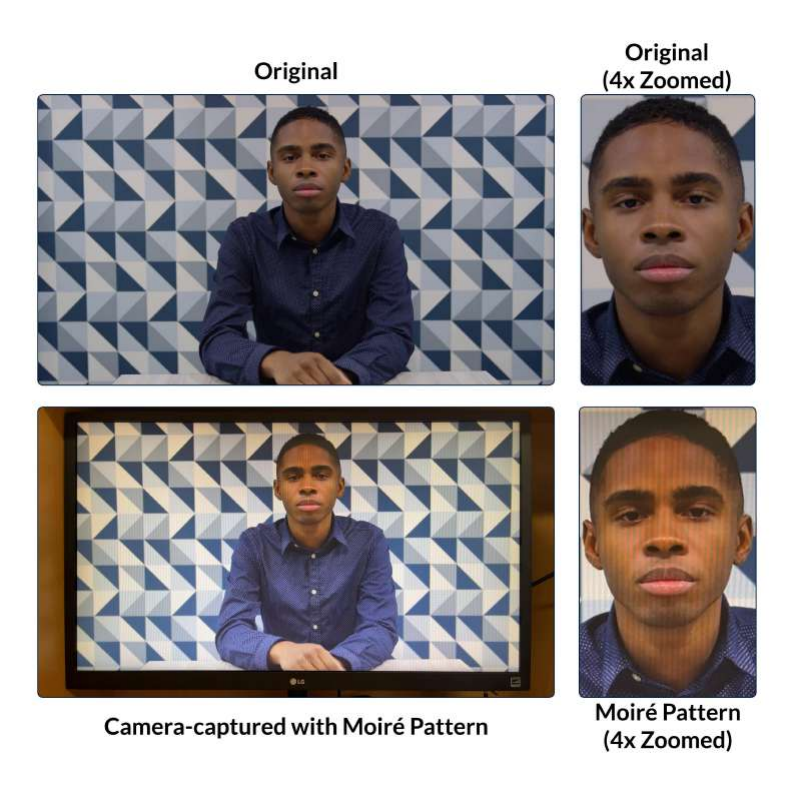}
        \caption{\textbf{\textsc{Original vs. Moiré Pattern Images:}} The top row displays original images without Moiré artifacts, with the leftmost image showing a full-frame original and the rightmost a 4× zoomed-in view. The bottom row illustrates how smartphone capture on a monitor introduces Moiré patterns, with the leftmost image showing the full-frame effect and the rightmost highlighting its distortions.}
        \label{fig:Demoire1}
\end{figure}
\newpage
\begin{figure}[h]
        \centering
        \includegraphics[scale=0.2, width=\textwidth]{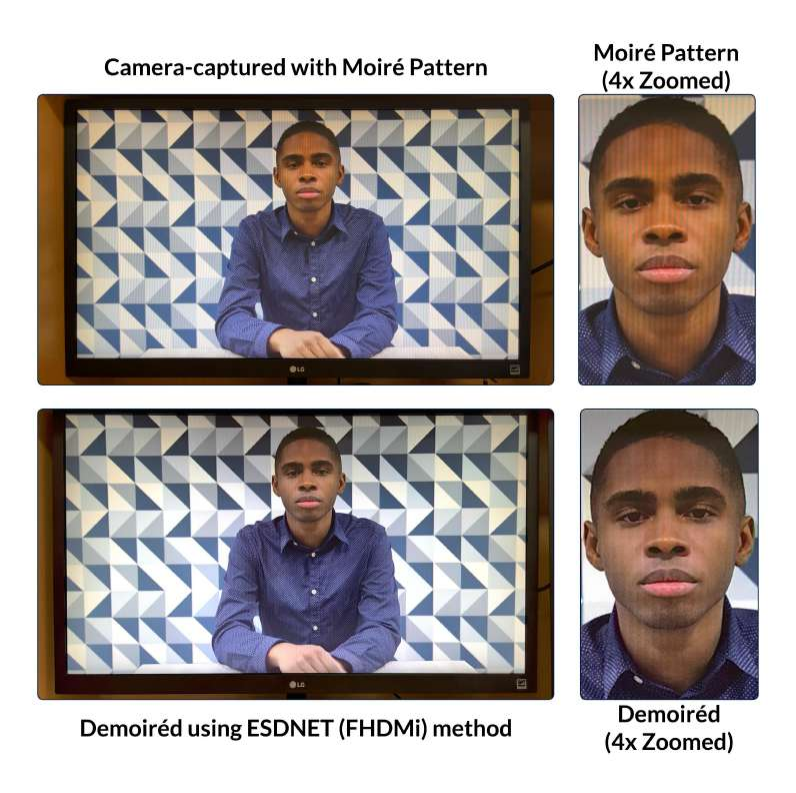}
        \caption{\textbf{\textsc{Moiré vs. Demoiréd Images:}} The top row shows smartphone-captured images with Moiré artifacts, with the leftmost image presenting the full-frame pattern and the rightmost a 4× zoomed-in view. The bottom row demonstrates the effect of ESDNet (FHDMi) in removing Moiré patterns, with the leftmost image showing the demoiréed result and the rightmost a zoomed-in comparison.}
        \label{fig:Demoire2}
    \label{fig:comparison_demoire}
\end{figure}
\newpage
\section{Impact of Lighting Conditions on Moiré-Captured Data}

\begin{figure}[h] 
  \centering
  \includegraphics[width=1\textwidth]{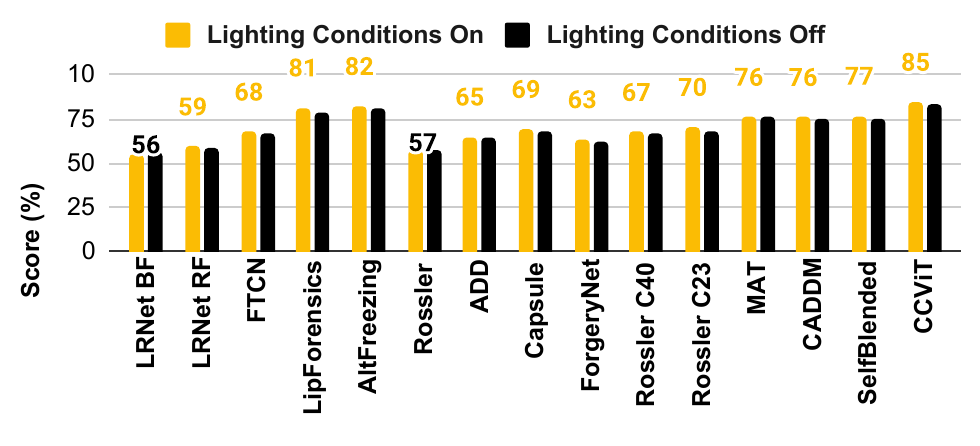}
    \caption{\textsc{\textbf{Lighting Conditions:}} Performance scores were similar across both conditions, though the `lights on' (yellow) setting showed a slight improvement.\textsc{\textbf{Lightning Conditions vs Performance for Each Detection Method.}}}
    \label{fig:lights}
\end{figure}

Lighting conditions can influence how Moiré patterns appear in captured videos, potentially affecting the performance of deepfake detection models. To evaluate this, we compared detection accuracy under two ambient lighting setups: lights on and lights off. As shown in Figure~\ref{fig:lights}, most detection models exhibited minimal variation in performance across the two conditions.
Interestingly, models such as MAT, CCViT, and CADDM showed slightly higher scores when the lights were on. This may be attributed to increased ambient reflections and contrast enhancement, which can amplify the visibility of Moiré artifacts and make them easier for detectors to exploit. However, the overall difference was relatively small, suggesting that most models are robust to moderate changes in lighting during video capture.
These results indicate that while lighting does have a minor effect on performance, the core challenge remains the presence of Moiré artifacts themselves rather than illumination conditions alone.

\section{Other Type of Distortions}
\autoref{table:GaussianBlurringSharpening} shows the AUC performance of Rossler C23 and MAT detectors under Gaussian blurring and sharpening. Both detectors experienced performance degradation as the blur kernel size increased, indicating vulnerability to low-pass filtering. Notably, MAT showed a sharper decline under severe blurring (15×15 kernel), suggesting greater sensitivity to the removal of fine-grained features.
Conversely, sharpening generally improved detection performance, particularly for MAT, which achieved its highest AUC (98.2\%) with a sharpening kernel value of 3. This suggests that enhancement of local edges and high-frequency details can accentuate manipulation artifacts, benefiting attention-based detectors.
These results highlight the asymmetric effects of low-level image transformations on deepfake detectors and the importance of evaluating robustness under diverse perturbation types.

\begin{table}[h!]

\centering
\caption{\textsc{\textbf{Performance Under Gaussian Blurring and Sharpening:}} AUC scores of Rossler C23 and MAT detectors on original, blurred, and sharpened images.}
\label{table:GaussianBlurringSharpening}
\small
\resizebox{\linewidth}{!}{
\begin{tabular}{l|c|cc|cc}
\toprule
\multirow{2}{*}{\textsc{\textbf{Detectors}}} & \multirow{2}{*}{\textsc{\textbf{Original}}} & \multicolumn{2}{c|}{\begin{tabular}[c]{@{}c@{}}\textbf{Gaussian Blurring }\\\textbf{(Kernel Size)}\end{tabular}} & \multicolumn{2}{c}{\begin{tabular}[c]{@{}c@{}}\textbf{Sharpening}\\\textbf{(Kernel Value)}\end{tabular}}\\ 
\cline{3-6}
& & \textsc{\textbf{7×7}} & \textsc{\textbf{15×15}} & \textsc{\textbf{3}} & \textsc{\textbf{5}} \\ 

\hline
Rossler C23 & 86.5 & \textbf{74.5} & \textbf{67.5} & 88.5 & 76.7 \\
MAT         & 87.0 & 71.3 & 57.7 & \textbf{98.2} & \textbf{88.6} \\ 
\bottomrule
\end{tabular}
}
\end{table}

\newpage
\section{Image Demoiréing Methods}
\noindent
We assessed the impact of applying four state-of-the-art demoiréing methods DMCNN, MBCNN, ESDNet (two variants), and DDA on deepfake detection performance using the DMF dataset. As these techniques are primarily designed for image-based restoration, evaluations were conducted using image-based detectors. To extend the analysis to video-level settings, LipForensic was additionally employed due to its capability to operate on both image and video modalities. The results in the following table reflect detection performance after demoiréing.

\begin{table}[h]
\centering
 \caption{\textsc{\textbf{Different Demoiréing Methods:}} We tested four state-of-the-art methods, finding up to a 16\% performance decline. This may be due to Demoiréing removing key deepfake artifacts needed for detection. Here, OG denotes Original.}
 \label{table:Demoiring}
\resizebox{\linewidth}{!}{%
\begin{tabular}{l|cc|ccccc|c} 
\toprule
\multicolumn{1}{l|}{\multirow{3}{*}{\textsc{\textbf{Detectors}}}} & \multirow{3}{*}{\begin{tabular}[c]{@{}c@{}}\textsc{\textbf{AUC}}\\\textsc{\textbf{on OG}}\end{tabular}} & \multirow{3}{*}{\begin{tabular}[c]{@{}c@{}}\textsc{\textbf{AUC}}\\\textsc{\textbf{on Moiré}}\end{tabular}} & \multicolumn{6}{c}{\textsc{\textbf{Demoiréing Methods Performance}}} \\ 
\cline{4-9}
\multicolumn{1}{c|}{} &  &  & \begin{tabular}[c]{@{}c@{}}\textit{ESDNet}\\\textit{(FHDMi)}\end{tabular} & \begin{tabular}[c]{@{}c@{}}\textit{ESDNet}\\\textit{(UHDM)}\end{tabular} & \textit{MBCNN} & \textit{DMCNN} & \textit{DDA} & \textit{Average} \\ 
\hline
Rossler & \textit{67.7} & \textit{58.5} & 58.1 & 53.7 & 55.9 & 57.1 & 54.4 & 55.8 \\
ADD & \textit{69.7} & \textit{67.6} & 68.5 & 65.5 & 66.1 & 65.7 & 64.8 & 66.1 \\
Capsule & \textit{71.3} & \textit{70.4} & 69.4 & 60.7 & 60.5 & 62.0 & 59.9 & 62.5 \\
ForgeryNet & \textit{76.9} & \textit{64.3} & 64.4 & 60.4 & 54.6 & 61.3 & 51.2 & 58.4 \\
Rossler C40 & \textit{77.0} & \textit{68.2} & 69.1 & 66.6 & 64.2 & 66.5 & 63.3 & 65.9 \\
Rossler C23 & \textit{86.5} & \textit{72.8} & 76.9 & 69.2 & 67.3 & 71.5 & 66.5 & 70.3 \\
MAT & \textit{87.0} & \textit{75.2} & 75.5 & 66.0 & 63.6 & 65.9 & 63.4 & 66.9 \\
CADDM & \textit{87.1} & \textit{78.5} & 79.5 & 73.4 & 72.7 & 75.0 & 72.3 & 74.6 \\
SelfBlended & \textit{88.8} & \textit{78.4} & 73.6 & 60.7 & 70.1 & 70.6 & 69.3 & 68.9 \\
LipForensics & \textit{90.6} & \textit{83.3} & 67.2 & 66.1 & 66.1 & 71.6 & 65.9 & 67.3 \\
CCViT & \textbf{\textit{95.0}} & \textbf{\textit{85.8}} & \textbf{84.5} &\textbf{75.3} & \textbf{76.2} & \textbf{82.2} & \textbf{77.9} & \textbf{79.2} \\ 
\hline
\multicolumn{3}{c|}{\textbf{Avg. AUC loss (DeMoiré vs. Moiré)}} & -1.5\da & -7.8\da & -7.8\da & 4.9\da & 8.6\da & 6.1\da\ \\
\multicolumn{3}{c|}{\textbf{Avg. AUC loss (DeMoiré vs. OG)}} & -10.1\da & -16.4\da & -16.4\da & 13.5\da & 17.2\da & 14.7\da\ \\
\bottomrule
\end{tabular}
}

\end{table}

\section{Video Demoiréing Methods} In \autoref{table:videoDemoiring}, we present the AUC performance of three video-based deepfake detectors AltFreezing, FTCN, and LipForensics on original (clean), Moiré-affected, and demoiréd videos processed using VD-Moiré\footnote{\url{https://github.com/CVMI-Lab/VideoDemoireing}} and FPANet\footnote{\url{https://github.com/kuai-lab/nn24_FPANet}}. Overall, the table highlights the negative impact of Moiré patterns on detection performance, with all models experiencing performance drops when tested on Moiré-affected videos. Notably, AltFreezing's AUC fell significantly from 100.0\% to 84.4\% and further decreased to 74.7\% with VD-Moiré, though it partially recovered to 92.9 with FPANet. FTCN showed inconsistent results, dropping from 56.3\% to 43.8\% on Moiré videos and failing to improve meaningfully with FPANet (40.6\%), though VD-Moiré slightly boosted its score to 68.8\%. LipForensics demonstrated the most resilience, with a minor drop from 100.0\% to 87.5\% and recovering to 90.6\% with both demoiréing methods. These results suggest that while some models, like LipForensics, benefit modestly from demoiréing, others remain sensitive to artifacts even after processing, indicating the limited generalizability of current demoiréing techniques across different detector architectures.
\begin{table}[h!]
\centering
\caption{\textsc{\textbf{AUC performance of video-based detectors on clean, moiré, and demoiréd videos}}}
\label{table:videoDemoiring}
\begin{tabular}{l|c|c|c|c}
\toprule
\textbf{Detector} & \textbf{Original} & \textbf{Moiré Video} & \textbf{VD-Moiré (Demoiré)} & \textbf{FPANet (Demoiré)} \\
\hline
AltFreezing   & \textbf{100.0} & 84.4 & 74.7 & \textbf{92.9} \\
FTCN          & 56.3  & 43.8 & 68.8 & 40.6 \\
LipForensics  & \textbf{100.0} & \textbf{87.5} & \textbf{90.6} & \textbf{90.6} \\
\bottomrule
\end{tabular}
\end{table}

\newpage

\section{Denoising and Deblurring Methods}
We used the denoising and deblurring methods from the NAFNet\footnote{\url{https://github.com/megvii-research/NAFNet}}. The NAFNet models were trained on SSID and GoPro datasets, using widths of 32 and 64 to balance computational efficiency and accuracy. Experiments revealed that adjusting the lengths of specific components can enhance performance for tasks such as reducing image noise (SSID) and removing blurriness (GoPro). The utilization of a 32-bit width aims to optimize computing efficiency, while a 64-bit width seeks to achieve higher precision, enabling the model to deliver optimal results within the computational limitations.
\subsection{Evaluation of Denoising Method on Different Weights}
\begin{table}[h!]
\centering
\caption{\textsc{\textbf{Denoising Method:}} In the NAFNet (SSID) technique, we observed an unexpected decrease in performance of up to 19.5 percentage points due to the denoising process. This decline in detection accuracy is likely attributable to the denoising procedure. Furthermore, we found that the performance of the methods was less effective after denoising than after demoiréing.}

\label{table:Denoising}
\resizebox{\linewidth}{!}{
\begin{tabular}{l|c|c|c|c|c}
\toprule
\multicolumn{1}{l|}{\multirow{2}{*}{\textsc{\textbf{Detectors}}}} & \multirow{2}{*}
{\begin{tabular}[c]{@{}c@{}}\textsc{\textbf{AUC}}\\\textsc{\textbf{on Original}}\end{tabular}} & \multirow{2}{*}{\begin{tabular}[c]{@{}c@{}}\textsc{\textbf{AUC}}\\\textsc{\textbf{on Moiré}}\end{tabular}} & \multicolumn{3}
{c}{\textsc{\textbf{Denoising Method Performance}}} \\ 
\cline{4-6}
 &  &  & \textit{NAFNet (SSID-32)} & \textit{NAFNet (SSID-64)} & \textit{Average} \\ 
\hline
Rossler & \textit{67.7} & \textit{58.5} & 60.71 & 59.62 & 60.16 \\
ADD & \textit{69.7} & \textit{67.6} & 65.93 & 68.96 & 67.44 \\
Capsule & \textit{71.3} & \textit{70.4} & 54.53 & 56.73 &  55.63\\
ForgeryNet & \textit{76.9} & \textit{64.3} & 53.02 & 56.73  & 54.87\\
Rossler C40 & \textit{77.0} & \textit{68.2} & 48.63 & 50.82 & 49.72 \\
Rossler C23 & \textit{86.5} & \textit{72.8} & 66.76 & 67.58 & 67.17 \\
MAT & \textit{87.0} & \textit{75.2} & 84.89 & 85.16 & 85.02 \\
CADDM & \textit{87.1} & \textit{78.5} & 61.26 & 66.76 & 64.01 \\
SelfBlended & \textit{88.8} & \textit{78.4} & 75.55 & 72.25 & 73.90\\
LipForensics & \textit{90.6} & \textit{83.3} & 66.90 & 65.93 & 66.41\\
CCViT & \textit{95.0} & \textit{85.8} & 79.95 & 76.92 & 78.43 \\ 
\hline
\multicolumn{3}{c|}{\textbf{Avg. Performance loss (DeNoise vs. Moiré)}} & 7.7\da & 6.9\da & -7.3 \\
\multicolumn{3}{c|}{\textbf{Avg. Performance loss (DeNoise vs. Original)}} & 16.3\da & 15.4\da & -15.9 \\
\bottomrule
\end{tabular}
}
\end{table}

\subsection{Evaluation of Deblurring Method on Different Weights}
\begin{table}[h!]
\centering
\caption{\textsc{\textbf{Deblurring Method:}} We implemented the NAFNet (GoPro) technique for deblurring. Upon comparing the effectiveness of each method for demoiréing and denoising, we found that the deblurring technique exhibited the lowest performance of the two. The deblurring process led to an additional decrease in performance of up to 36.8 percentage points.}
\label{table:Deblurring}
\resizebox{\linewidth}{!}{
\begin{tabular}{l|c|c|c|c|c}
\toprule
\multicolumn{1}{l|}{\multirow{2}{*}{\textsc{\textbf{Detectors}}}} & \multirow{2}{*}
{\begin{tabular}[c]{@{}c@{}}\textsc{\textbf{AUC}}\\\textsc{\textbf{on Original}}\end{tabular}} & \multirow{2}{*}{\begin{tabular}[c]{@{}c@{}}\textsc{\textbf{AUC}}\\\textsc{\textbf{on Moiré}}\end{tabular}} & \multicolumn{3}
{c}{\textsc{\textbf{Deblurring Method Performance}}} \\ 
\cline{4-6}
 &  &  & \textit{NAFNet (GoPro-32)} & \textit{NAFNet (GoPro-64)} & \textit{Average} \\ 
\hline
Rossler & \textit{67.7} & \textit{58.5} & 48.63 & 47.66 &  48.14\\
ADD & \textit{69.7} & \textit{67.6} & 46.70 & 46.29 & 46.49 \\
Capsule & \textit{71.3} & \textit{70.4} & 52.06 & 50.14 & 51.10 \\
ForgeryNet & \textit{76.9} & \textit{64.3} & 46.43 & 46.98 & 46.70 \\
Rossler C40 & \textit{77.0} & \textit{68.2} & 45.33 & 44.23 & 44.78 \\
Rossler C23 & \textit{86.5} & \textit{72.8} & 60.03 & 55.08 & 57.55 \\
MAT & \textit{87.0} & \textit{75.2} &  77.75 & 74.31 & 76.03 \\
CADDM & \textit{87.1} & \textit{78.5} & 64.42 & 63.74 & 64.08 \\
SelfBlended & \textit{88.8} & \textit{78.4} & 59.34 & 60.44 & 59.89\\
LipForensics & \textit{90.6} & \textit{83.3} & 69.09 & 61.26 & 65.17\\
CCViT & \textit{95.0} & \textit{85.8} & 64.97 & 63.32 & 64.14  \\ 
\hline
\multicolumn{3}{c|}{\textbf{Avg. Performance loss (DeBlur vs. Moiré)}} & 17.3\da & 18.6\da & -17.1 \\
\multicolumn{3}{c|}{\textbf{Avg. Performance loss (DeBlur vs. Original)}} & 24.1\da & 25.1\da & -25.7 \\
\bottomrule
\end{tabular}
}
\end{table}

\newpage
\section{Performance on Original Dataset --- ROC Curve}
We conducted a comprehensive performance analysis of the original dataset, employing various methodologies on many datasets to evaluate the effectiveness of each method. This analysis aimed to assess the efficacy of four techniques moiré, demoiréing, denoising, and deblurring. Additionally, compare their performance to that of the original dataset.
\label{sec:Roc-curve}
\begin{figure}[h!]
    \centering
    \includegraphics[width=0.49\textwidth]{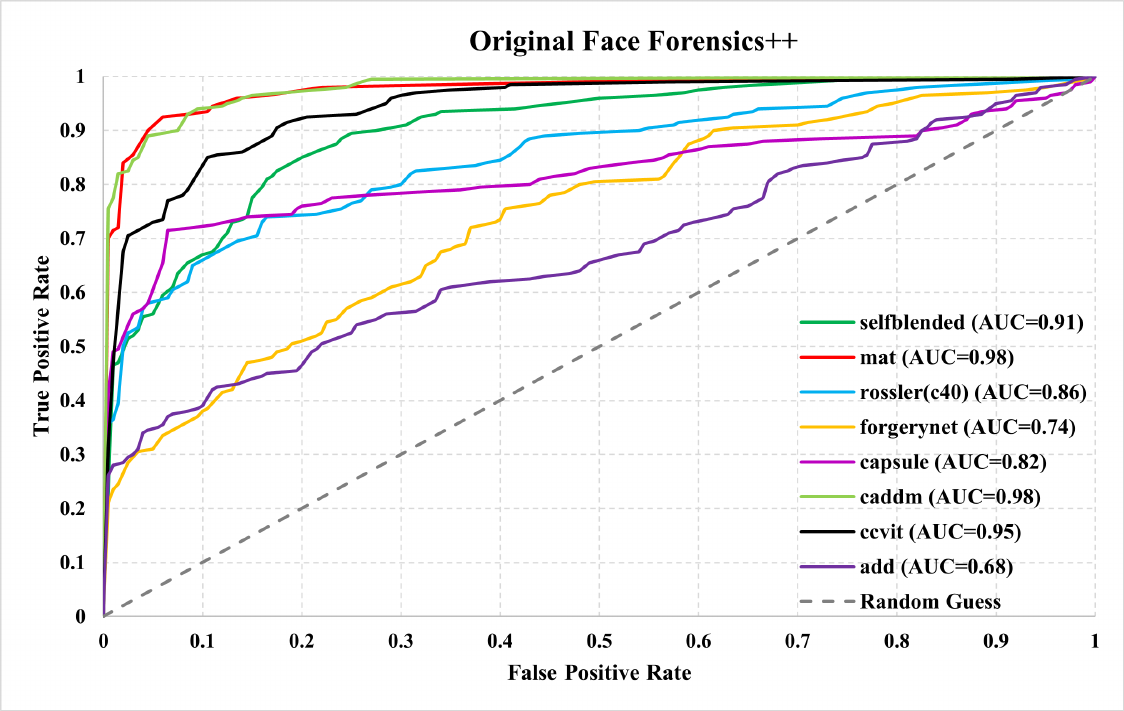}
    \includegraphics[width=0.49\textwidth]{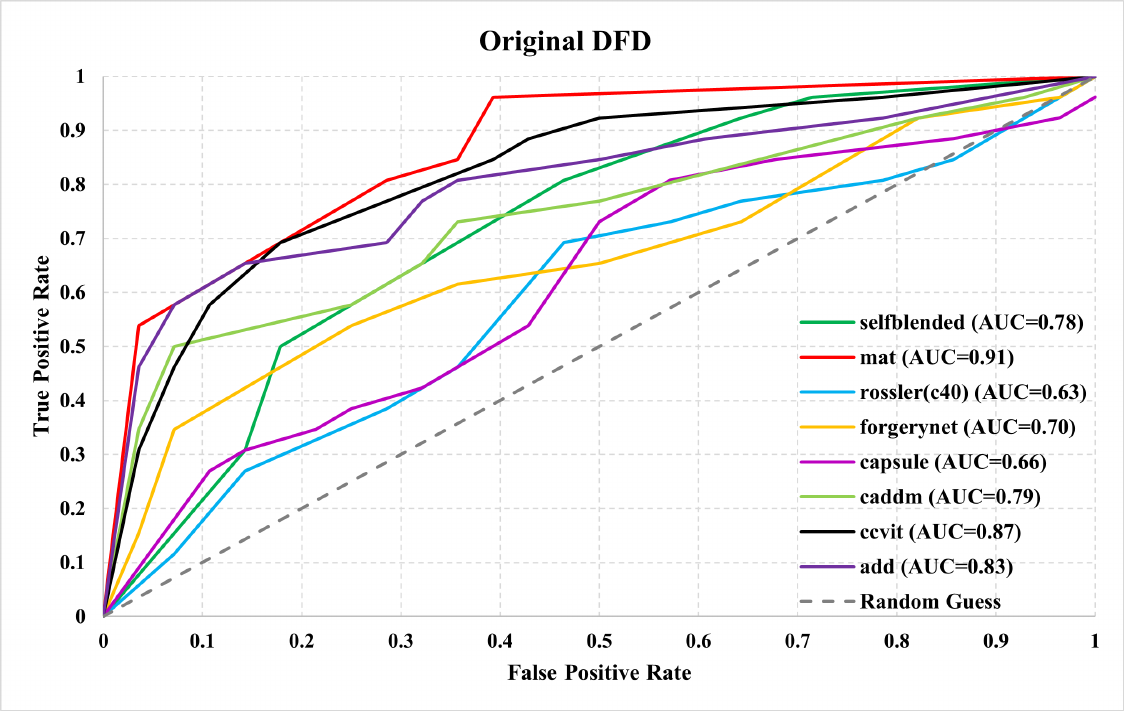}
    \includegraphics[width=0.49\textwidth]{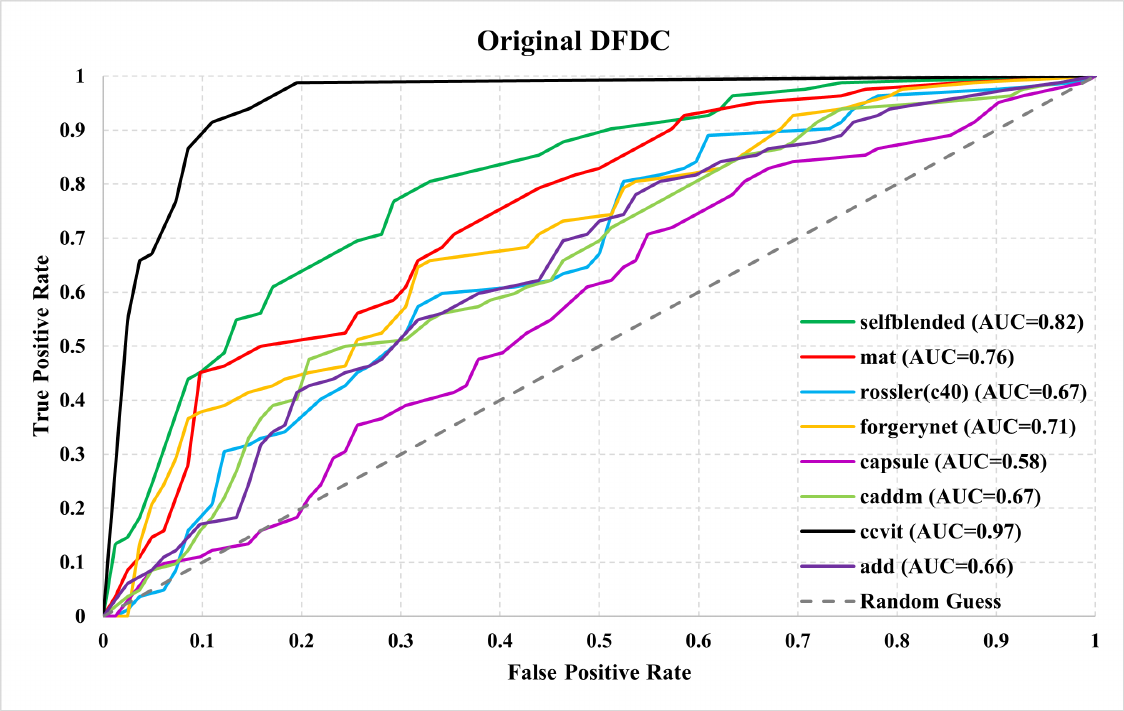}
    \includegraphics[width=0.49\textwidth]{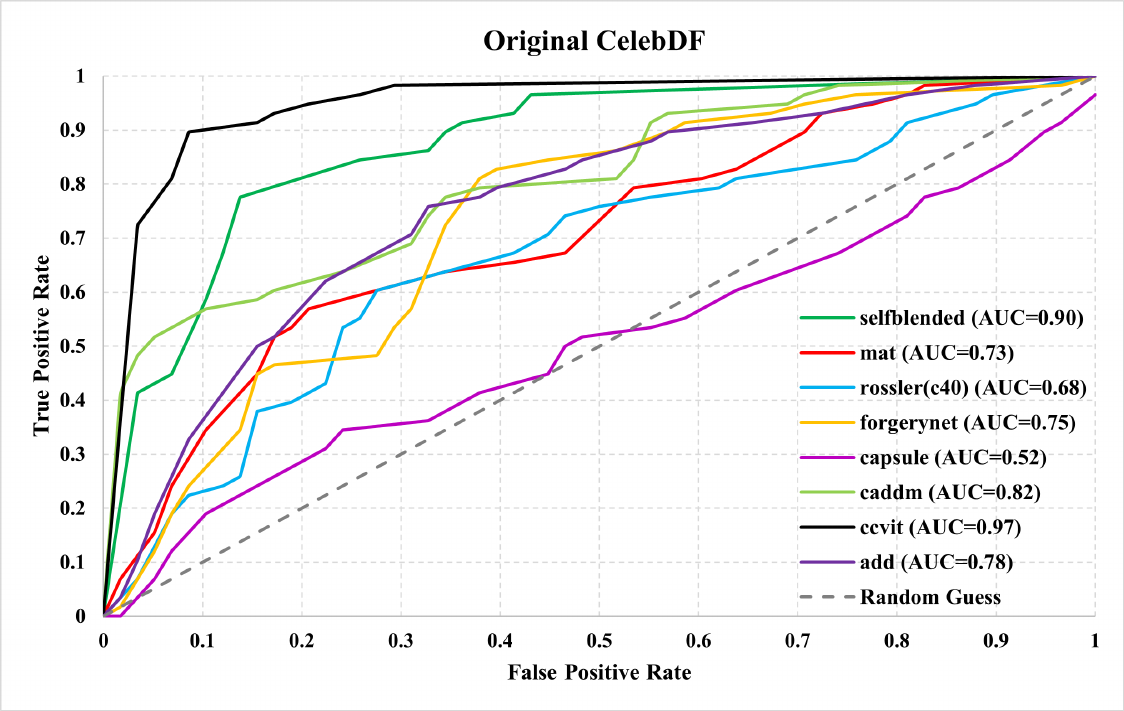}
    \includegraphics[width=0.49\textwidth]{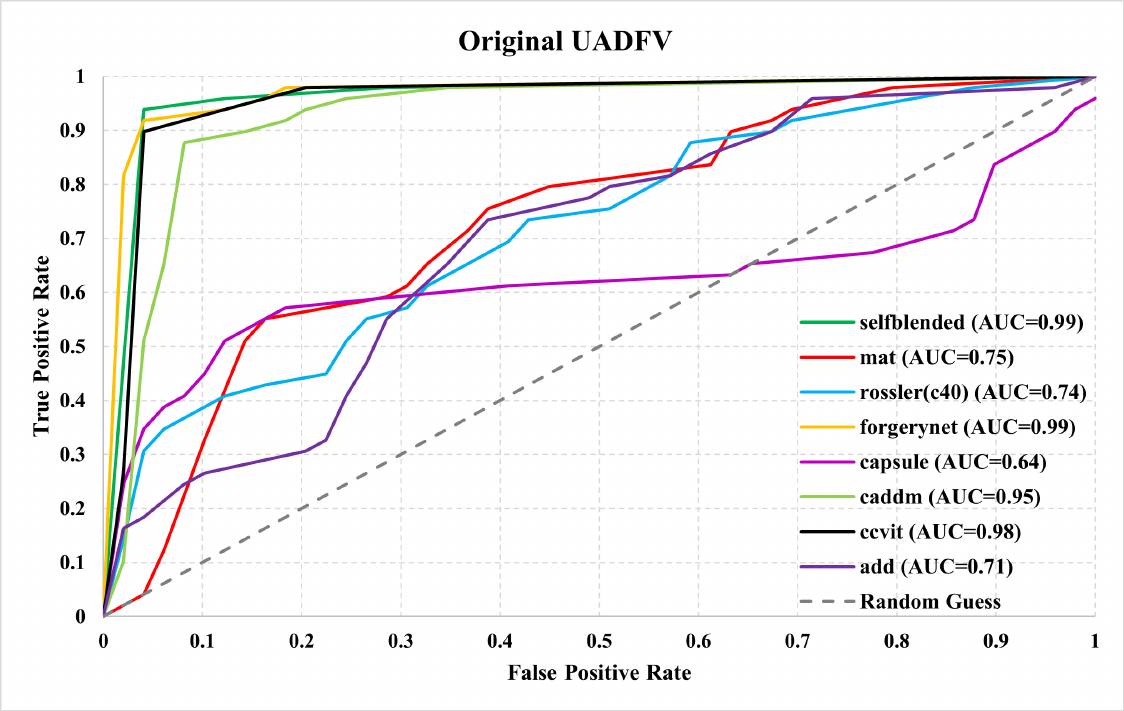}
    \caption{\textbf{\textsc{Performance on Original Datasets. }} The ROC AUC curves of different Deepfake Detectors on the Original deepfake datasets are unaffected by the Moiré-induced distortions. On FaceForensics++, MAT and CADDM models attain near-perfect AUCs of 0.98\%, with CCViT performing well at 0.95\%.  For DFD, the MAT model leads with an AUC of 0.91\%, followed by CCViT. In DFDC, CCViT again excels with an AUC of 0.97\%, while Capsule has the lowest AUC of 0.58\%. On CelebDF, it achieves an impressive AUC of 0.97\%, while the Capsule model performs poorly with an AUC of 0.52\%. For UADFV, SelfBlended and ForgeryNet models achieve AUCs of 0.99\%, closely followed by CCViT at 0.98\%. Capsule remains the least effective with an AUC of 0.64\%. CCViT consistently ranks in the top 3 across all datasets, showcasing its reliable performance.
}
    \label{fig:enter-label}
\end{figure}

\newpage

\section{Performance on \DatasetName\ (DMF) Dataset --- ROC Curve}
For the performance evaluation on the DMF Dataset, we mainly used a dataset obtained from a BenQ monitor, which displays the most pronounced Moiré patterns. The BenQ monitor was chosen to ensure the dataset presented substantial challenges, allowing us to effectively assess and measure the efficiency of our approaches in demanding scenarios. The Moiré patterns produced by the BenQ monitor provided a solid basis for evaluating the effectiveness of various methods on each dataset in distinguishing real and fake videos.
\subsection{Camera: Samsung S22 Plus}
\subsubsection{Lights Condition: ON}
\begin{figure}[h]
    \centering
    \includegraphics[width=0.45\textwidth]{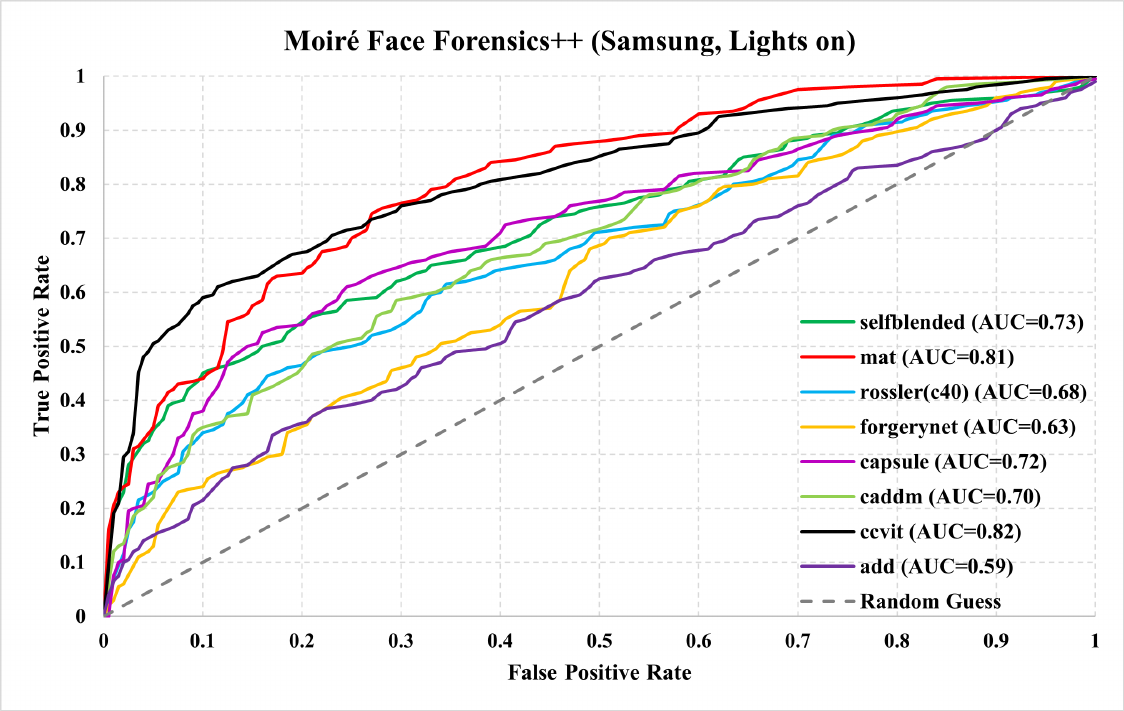}
    \includegraphics[width=0.45\textwidth]{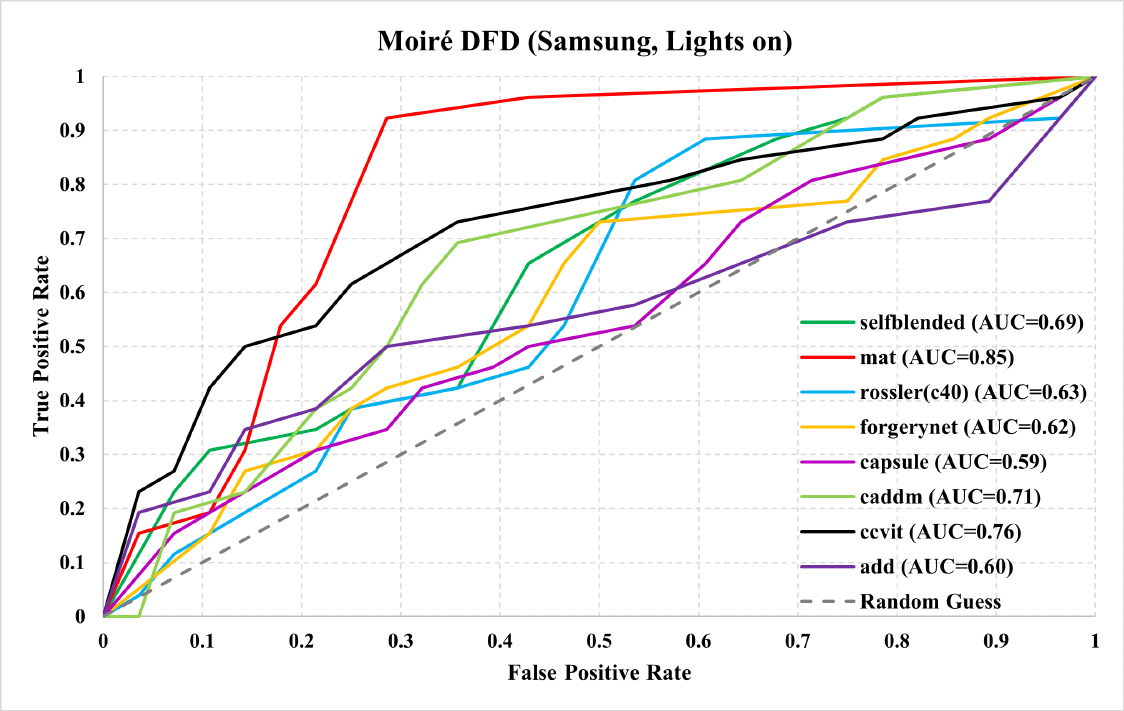}
    \includegraphics[width=0.45\textwidth]{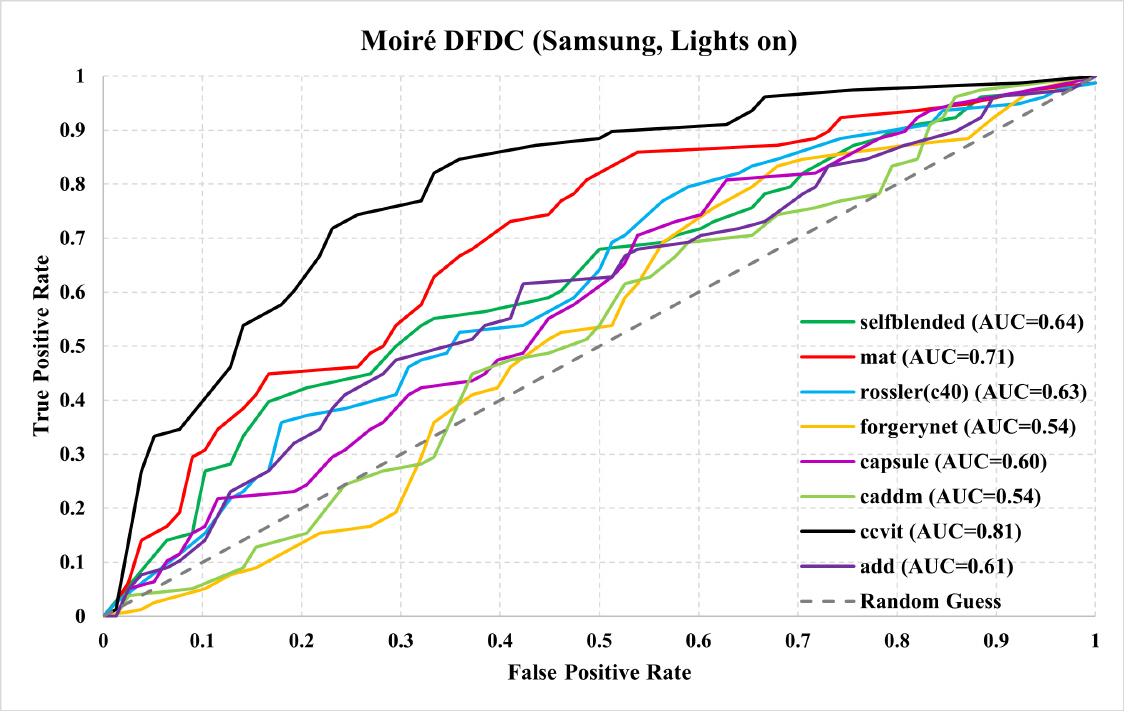}
    \includegraphics[width=0.45\textwidth]{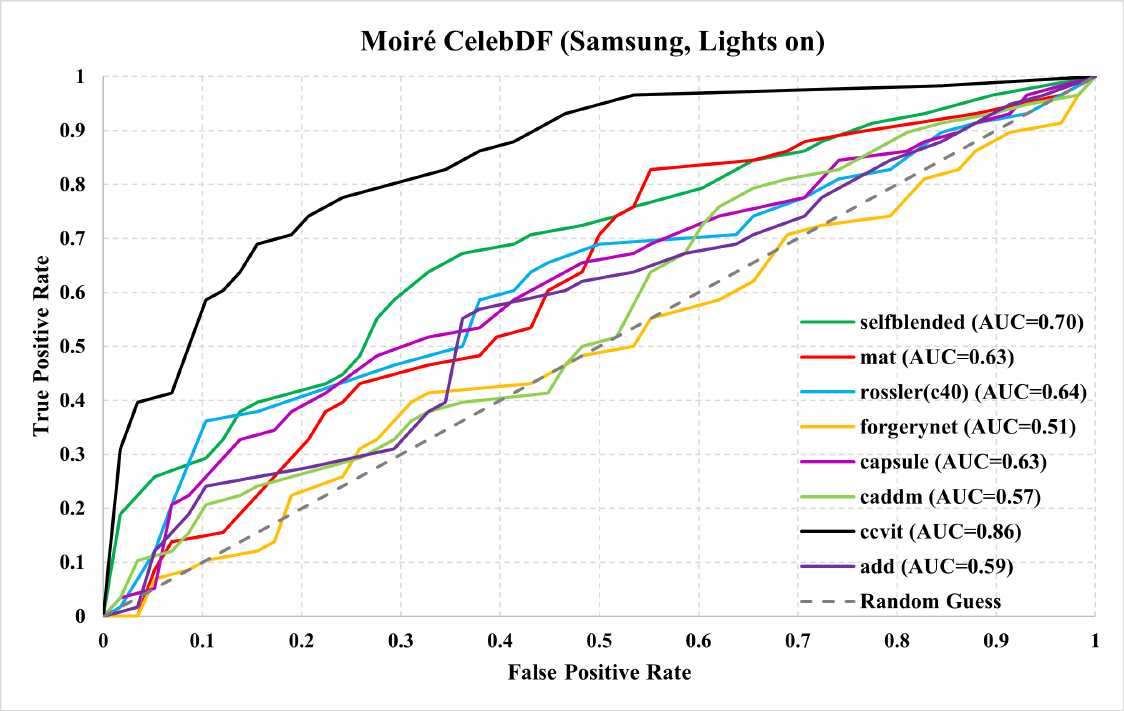}
    \includegraphics[width=0.45\textwidth]{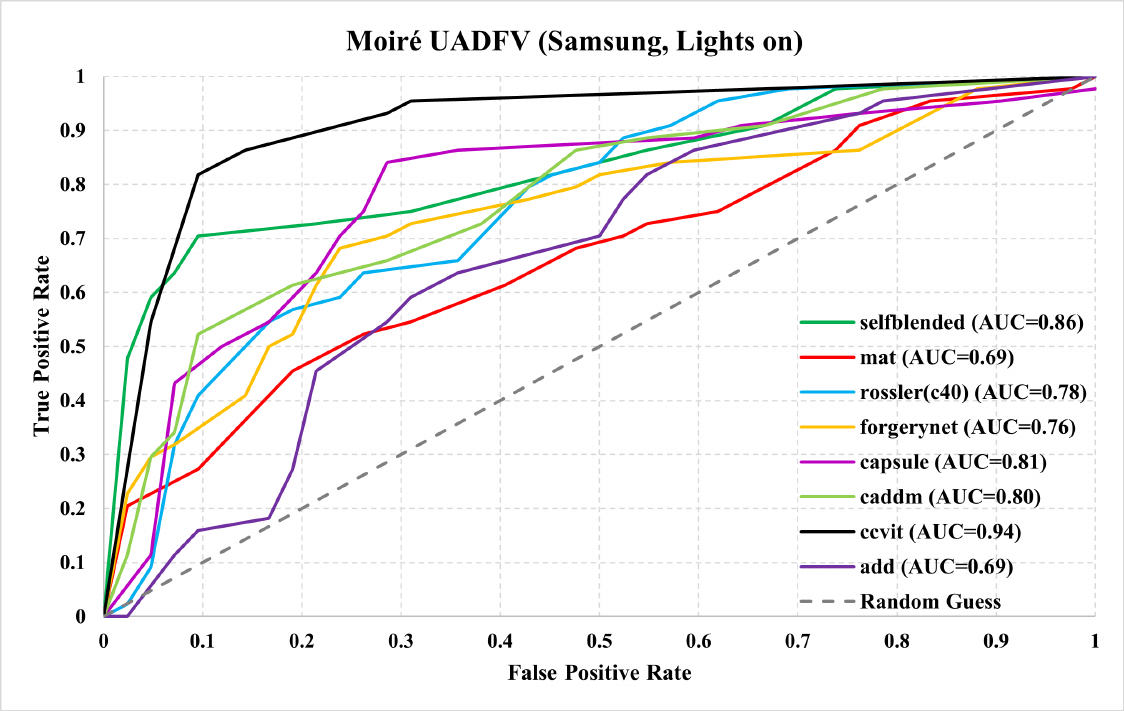}
    \caption{\textbf{\textsc{Performance on Moiré pattern-induced Datasets on BenQ Monitor with Lights on.}} The CCViT and MAT models generally perform best across different datasets. The performance of other models varies across datasets. On Moiré FaceForensics++, the CCViT and MAT models achieve AUC scores of 0.82\% and 0.81\%, respectively. In the Moiré DFD dataset, the MAT model leads with an AUC of 0.85\%, followed by CCViT with 0.76\%. For Moiré DFDC, CCViT scores 0.81\%, while MAT achieves 0.71\%. In Moiré CelebDF, CCViT leads with an AUC of 0.86\%. In Moiré UADFV, CCViT excels with an AUC of 0.94\%, followed by Capsule with 0.81\%.}
\end{figure}
\newpage
\subsubsection{Lights Condition: OFF}
\begin{figure}[h]
    \centering
    \includegraphics[width=0.49\textwidth]{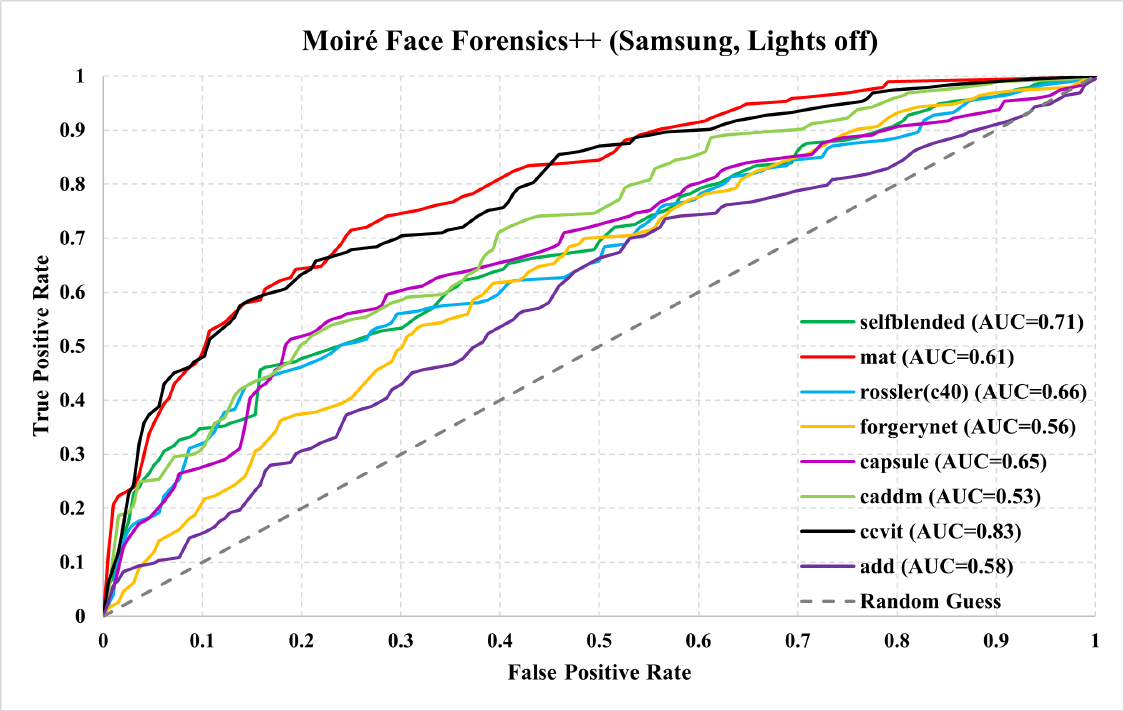}
    \includegraphics[width=0.49\textwidth]{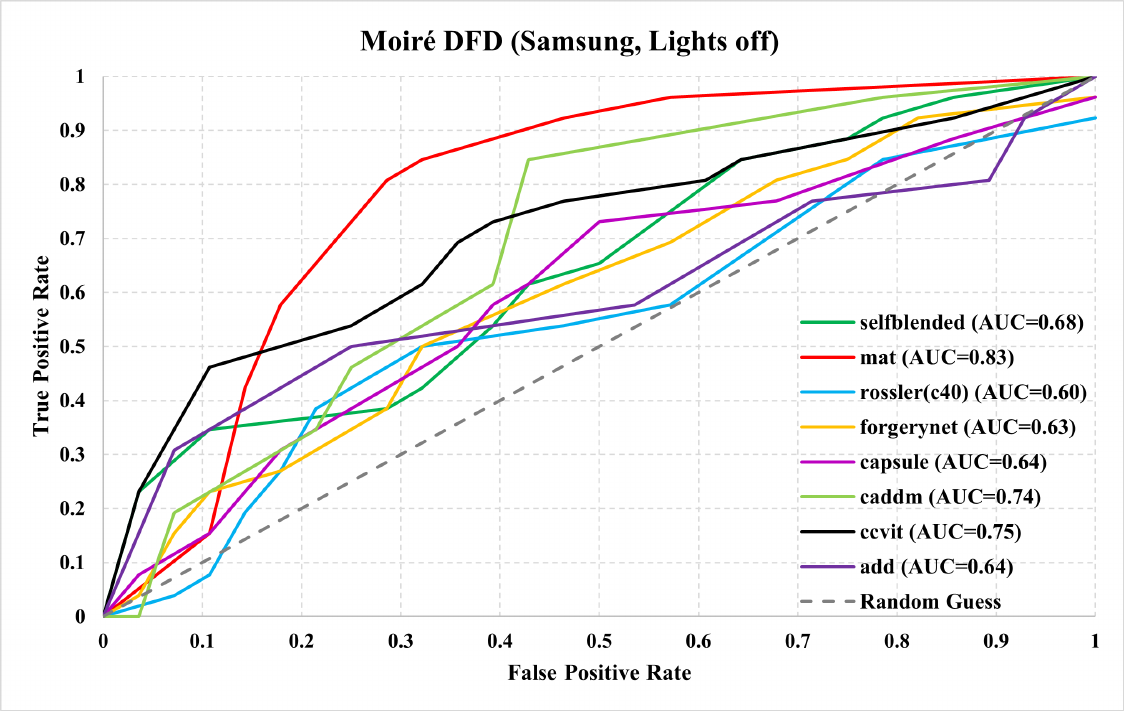}
    \includegraphics[width=0.49\textwidth]{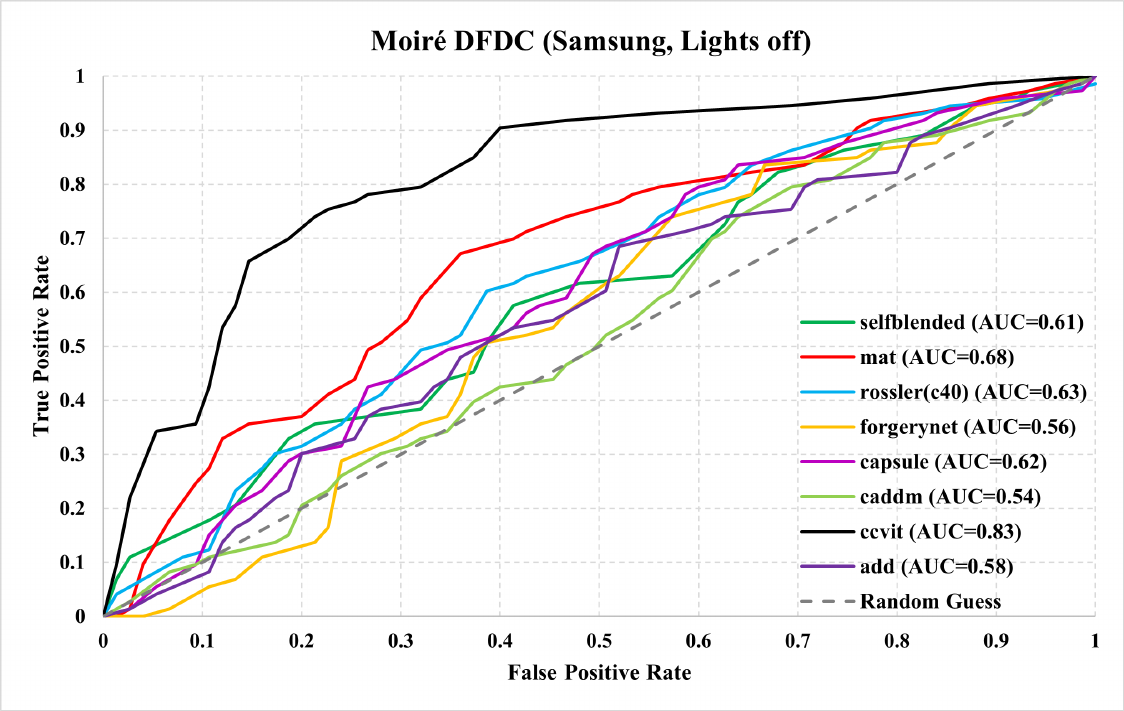}
    \includegraphics[width=0.49\textwidth]{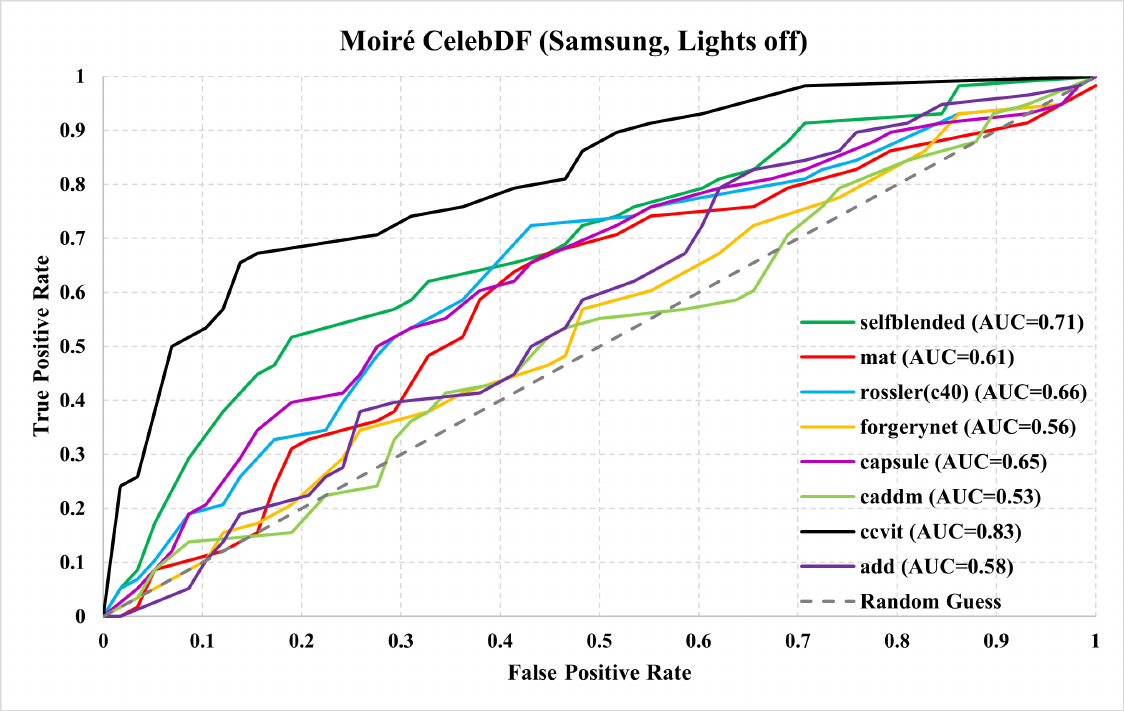}
    \includegraphics[width=0.49\textwidth]{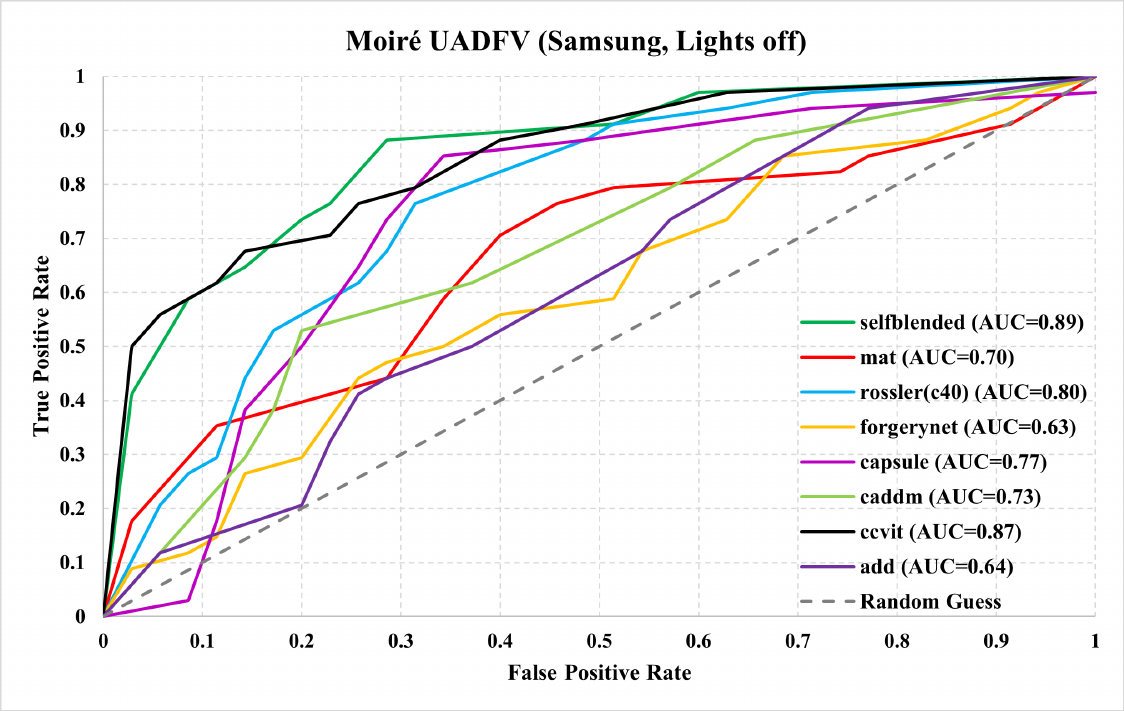}
    \caption{\textbf{\textsc{Performance on Moiré pattern-induced Datasets on BenQ Monitor with Lights off. }}For Moiré FaceForensics++, MAT leads with an AUC of 0.61\%, closely followed by CCViT at 0.83\%. In the Moiré DFD dataset, the MAT model leads with an AUC of 0.83\%, followed by CCViT with 0.75\%. In the Moiré DFDC dataset, CCViT performs best with an AUC of 0.83\%, followed by the MAT model with 0.68\%. For the Moiré CelebDF dataset, CCViT achieves the highest AUC of 0.83\%, followed by SelfBlended at 0.71\%. In the Moiré UADFV dataset, CCViT excels with an AUC of 0.87\% and SelfBlended scores of 0.89\%.}
\end{figure}

\newpage

\subsection{Camera: iPhone 13}
\subsubsection{Lights Condition: ON}
\begin{figure}[h]
    \centering
    \includegraphics[width=0.49\textwidth]{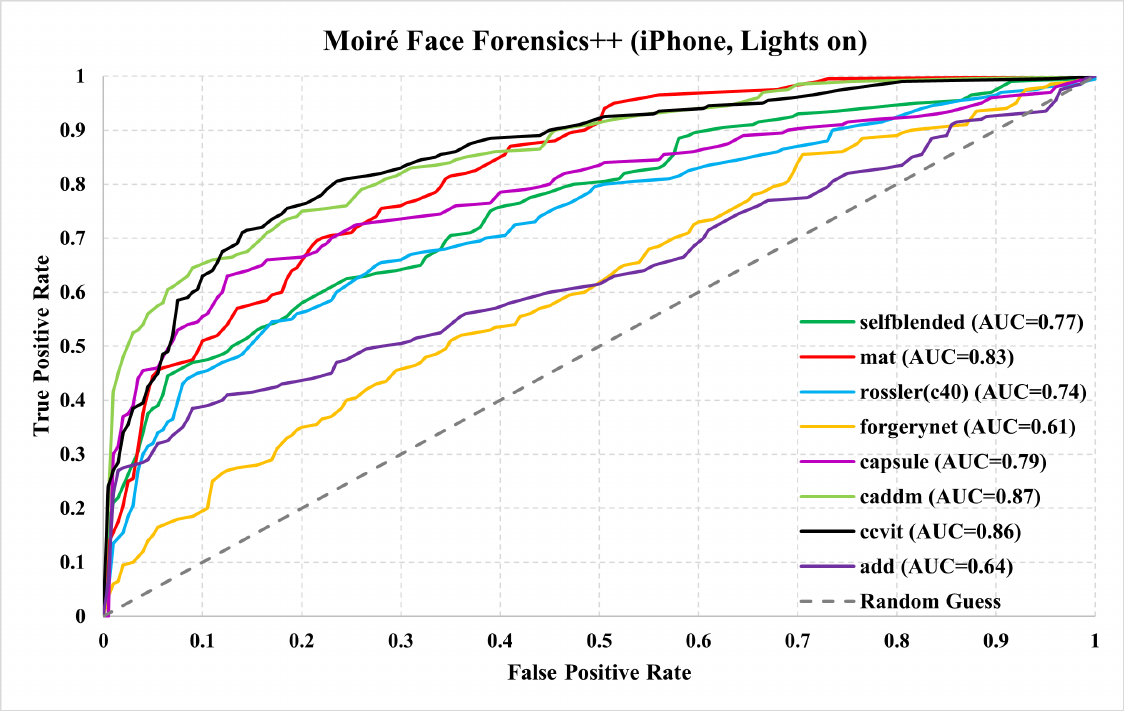}
    \includegraphics[width=0.49\textwidth]{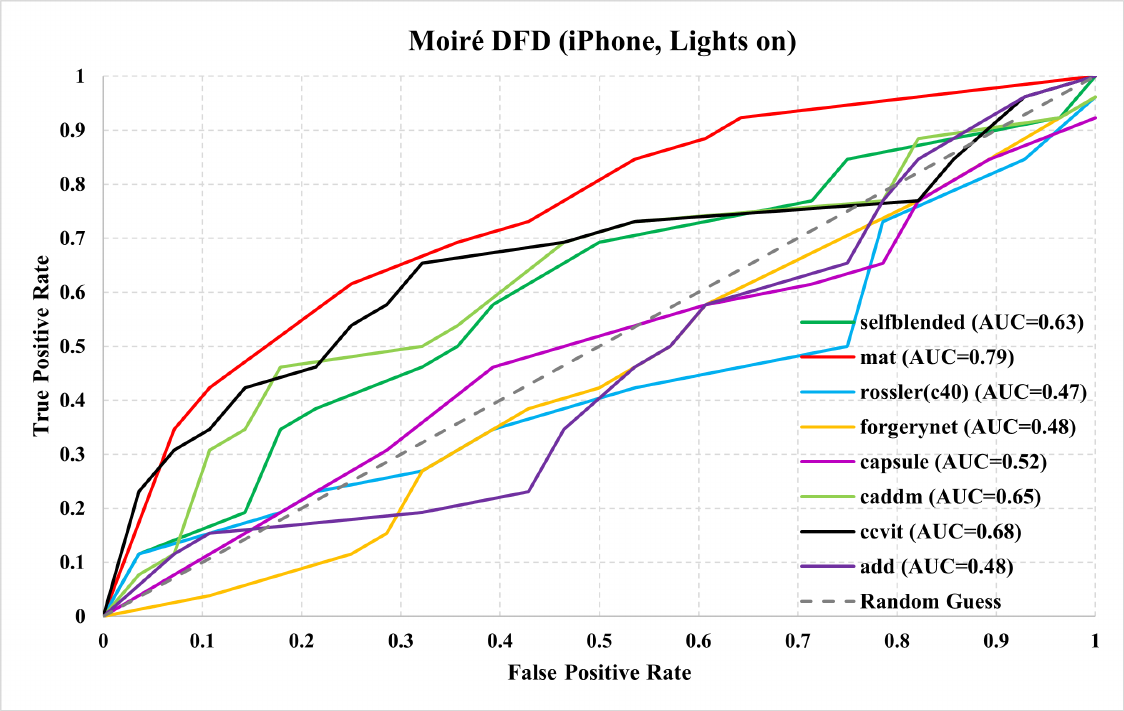}
    \includegraphics[width=0.49\textwidth]{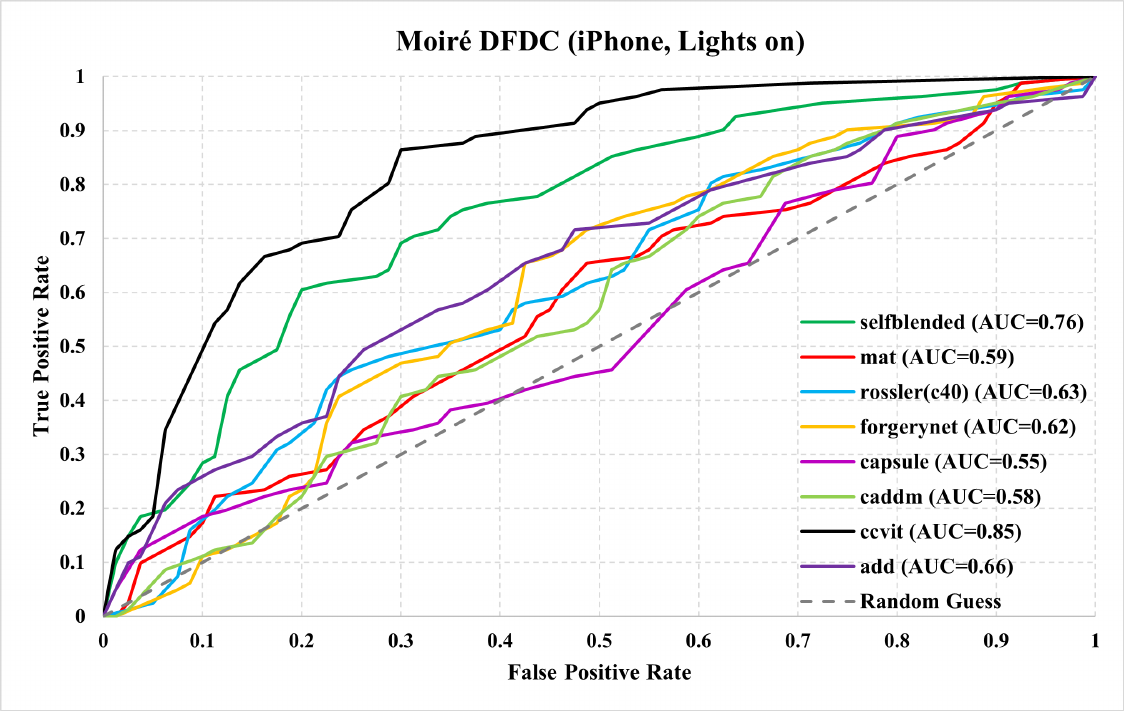}
    \includegraphics[width=0.49\textwidth]{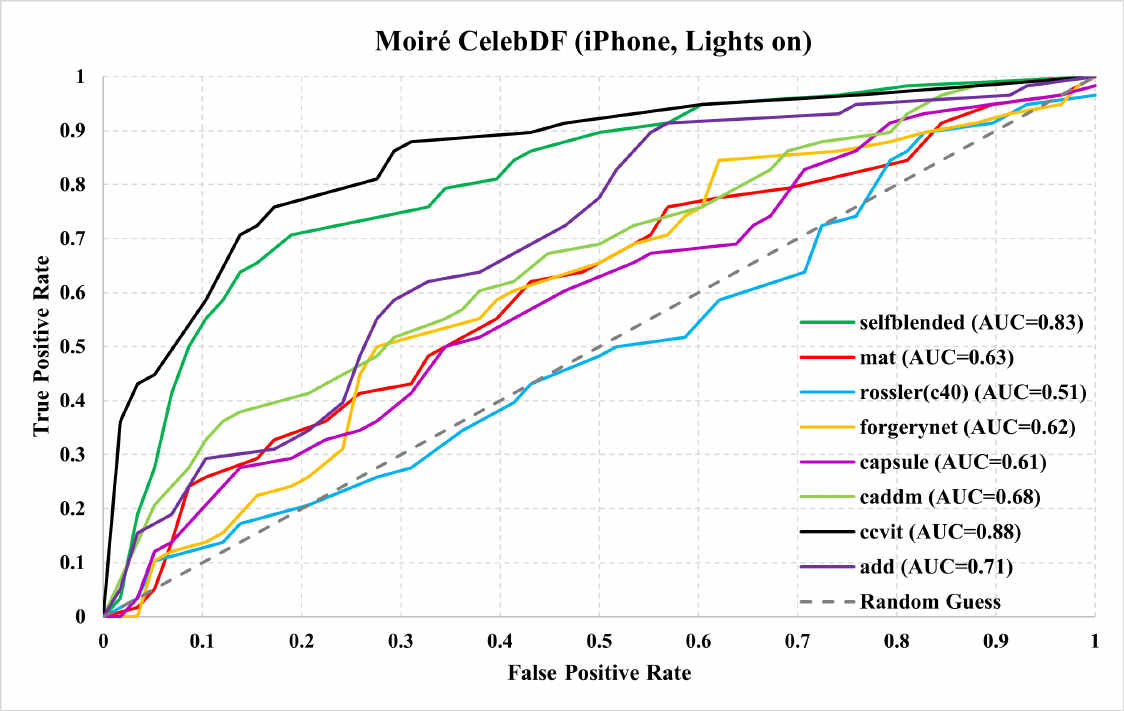}
    \includegraphics[width=0.49\textwidth]{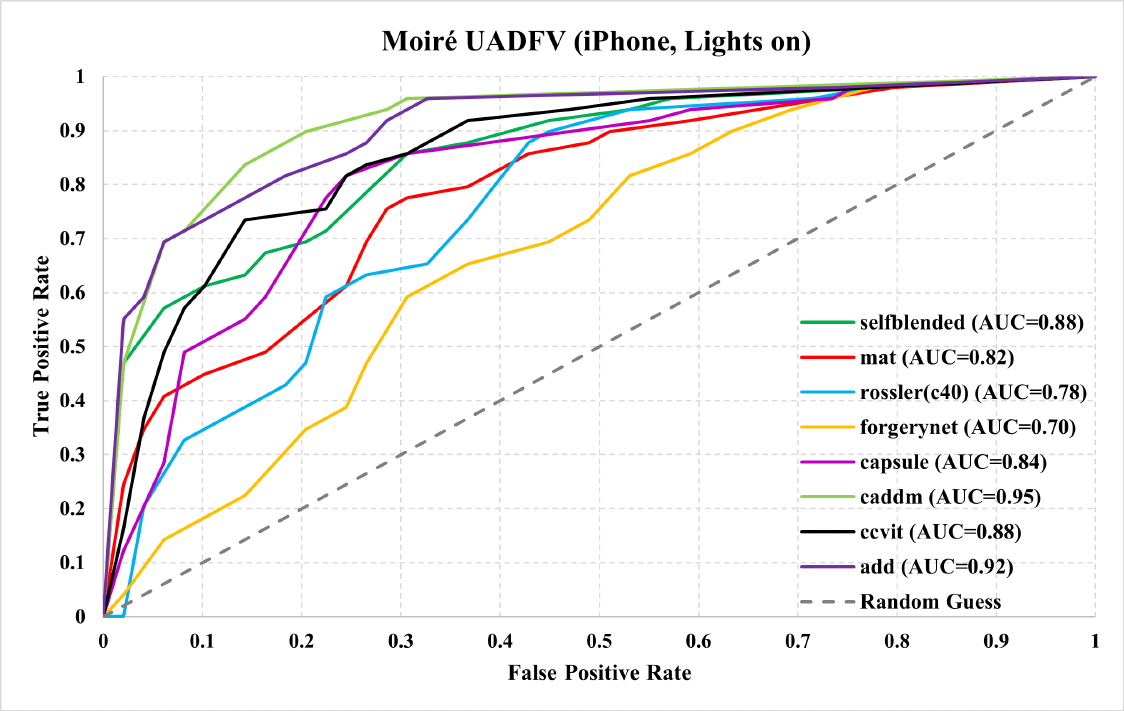}
    \caption{\textbf{\textsc{Performance on Moiré pattern-induced Datasets on BenQ Monitor with Lights on. }}The ROC curves display the performance of various deepfake detection methods on datasets captured with an iPhone in a light-on environment using a BenQ monitor, where moiré patterns were prominent. Across multiple datasets, the CCViT model consistently showed strong performance with AUCs of 0.86\% in Moiré FaceForensics++, 0.85\% in Moiré DFDC, and 0.88\% in Moiré CelebDF. The MAT model also performed well, with AUCs of 0.83\% in Moiré FaceForensics++ and 0.79\% in Moiré DFD. Additionally, SelfBlended and CADDM models demonstrated notable performance, particularly in the Moiré UADFV dataset, achieving AUCs of 0.88\% and 0.95\%, respectively. CCViT and MAT models were the most reliable across various datasets under these conditions.}
\end{figure}
\newpage
\subsubsection{Lights Condition: OFF}
\begin{figure}[h]
    \centering
    \includegraphics[width=0.49\textwidth]{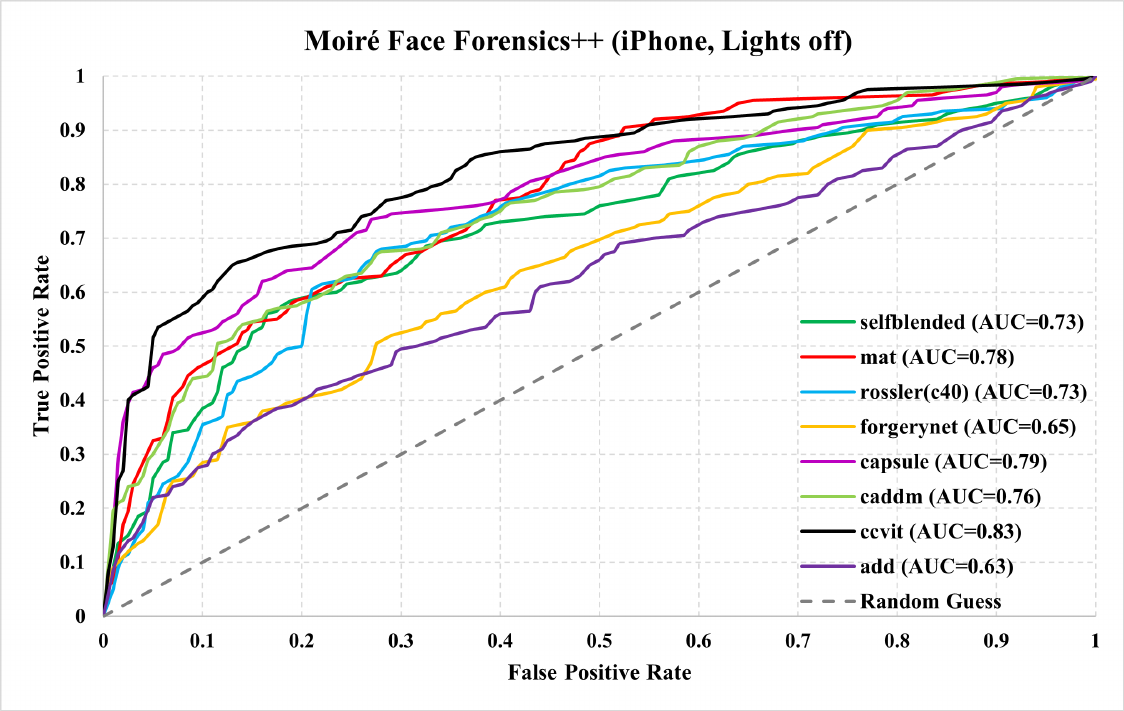}
    \includegraphics[width=0.49\textwidth]{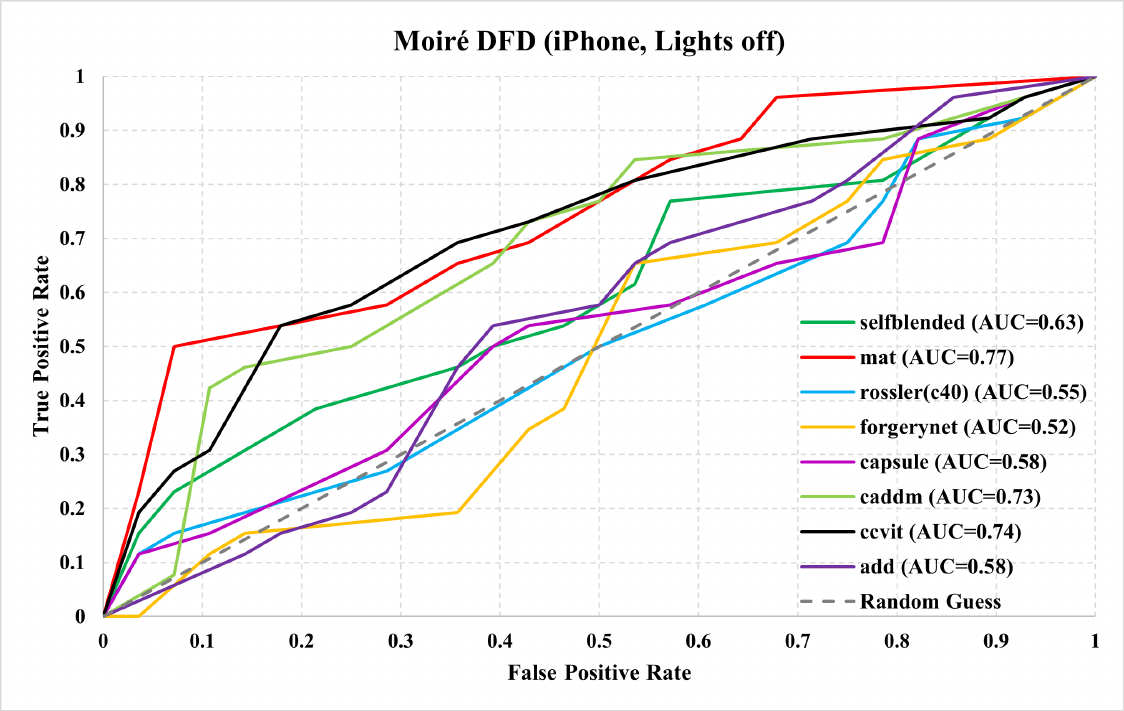}
    \includegraphics[width=0.49\textwidth]{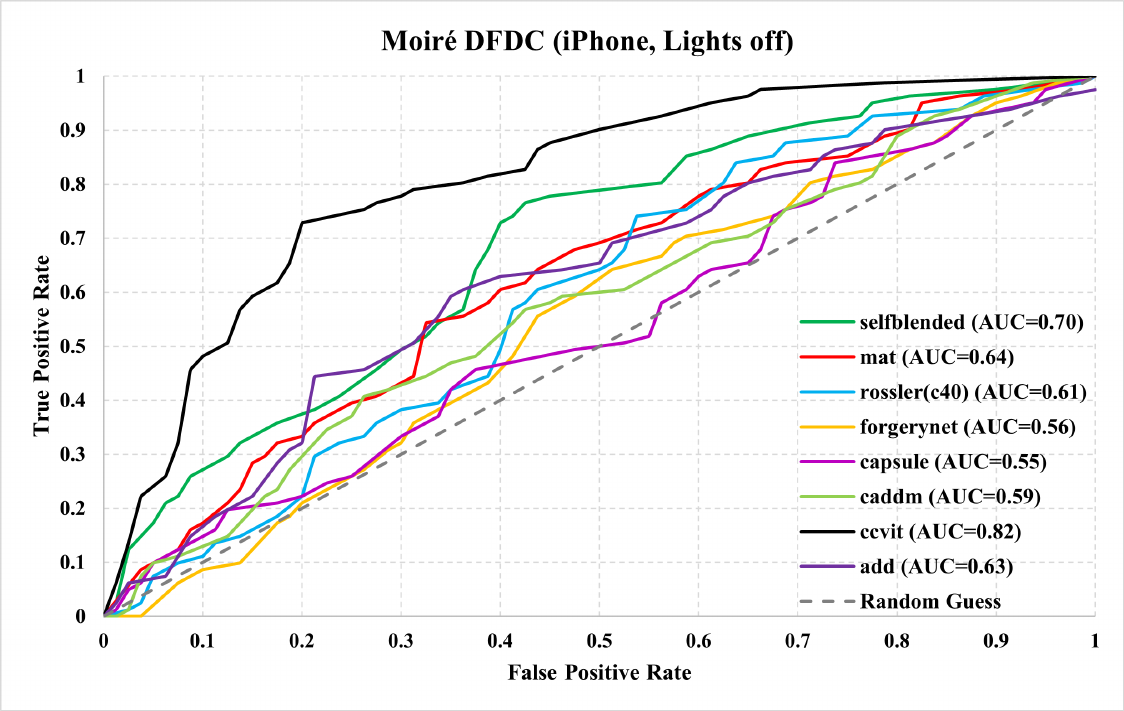}
    \includegraphics[width=0.49\textwidth]{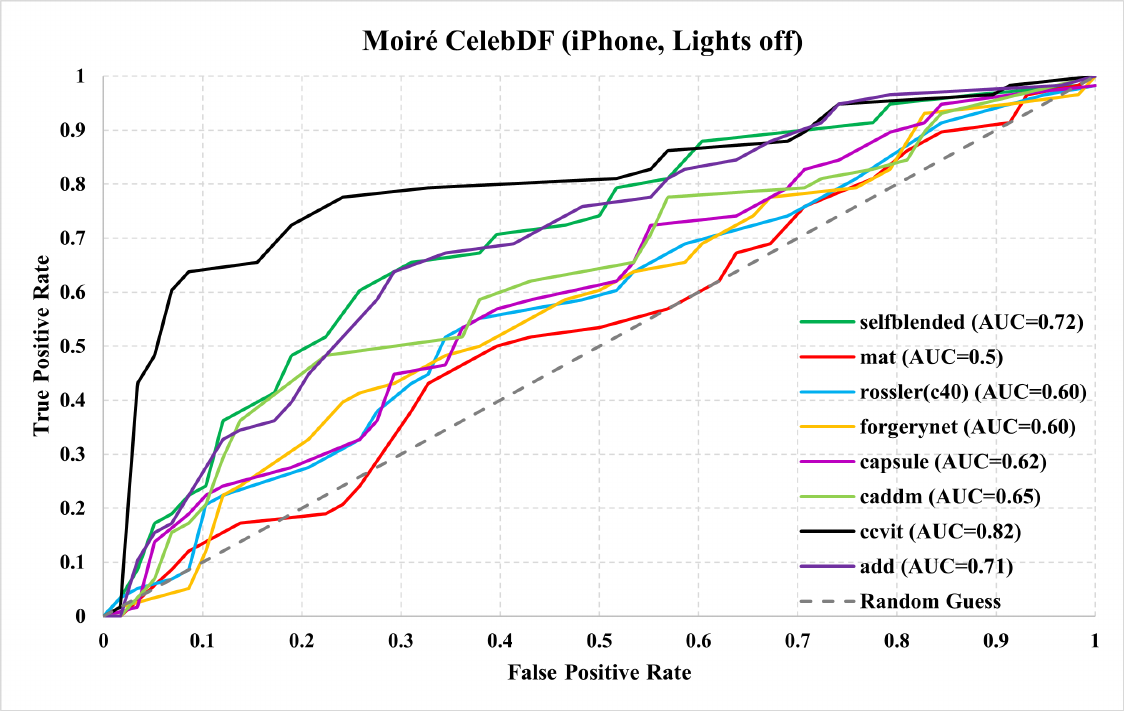}
    \includegraphics[width=0.49\textwidth]{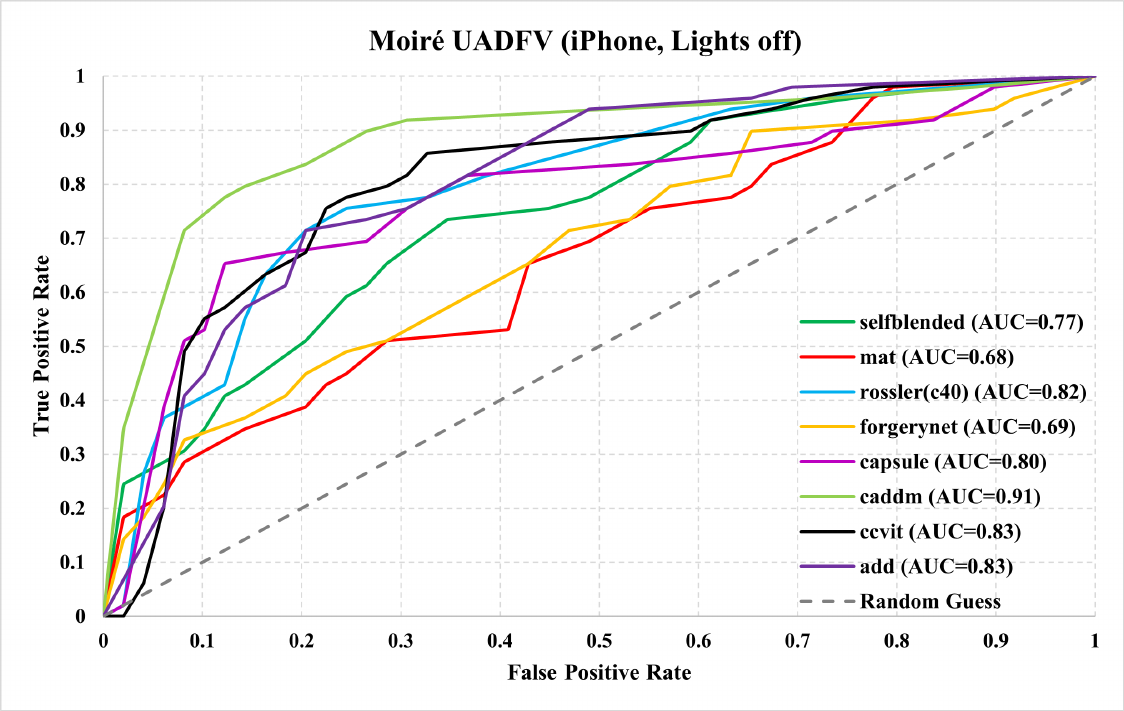}
    \caption{\textbf{\textsc{Performance on Moiré pattern-induced Datasets on BenQ Monitor with Lights off. }}CCViT consistently performs well across different datasets, achieving the highest AUCs of 0.83\% in Moiré FaceForensics++ and 0.82\% in both Moiré DFDC and CelebDF. MAT also shows strong performance with an AUC of 0.77\% in Moiré DFD and 0.78\% in Moiré FaceForensics++. Other notable performances include SelfBlended with an AUC of 0.72\% in Moiré CelebDF and CADDM with the highest AUC of 0.91\% in Moiré UADFV. Forgerynet generally shows lower performance across datasets. CCViT and MAT are the most reliable models for detecting deepfakes in these settings.}
\end{figure}

\newpage
\section{Performance after Demoiréing on DMF Dataset --- ROC Curve}
To compare performance on demoiré datasets, we compare detection results on deepfake datasets captured with a BenQ monitor, where the moiré pattern was most pronounced. This comparison utilized the ESDNet (FHDMi) method, the best-performing demoiré technique, and the evaluation was carried out using ROC curves.
\subsection{Camera: Samsung S22 Plus}
\subsubsection{Lights Condition: ON}
\begin{figure}[h]
    \centering
    \includegraphics[width=0.45\textwidth]{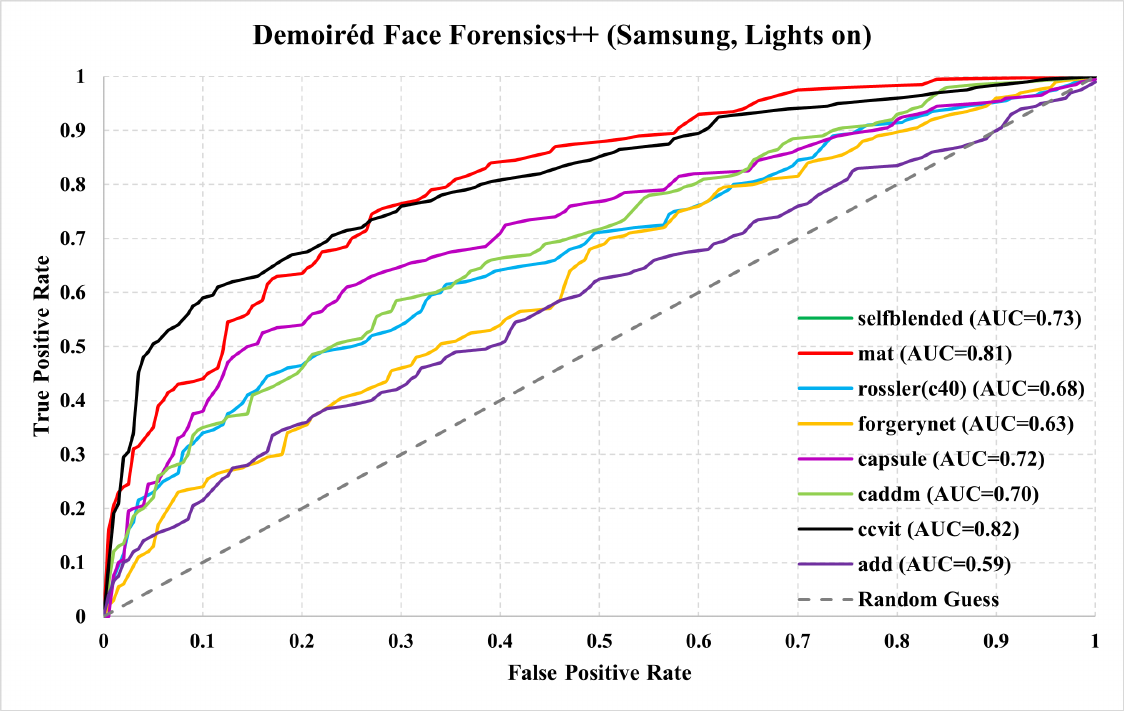}
    \includegraphics[width=0.45\textwidth]{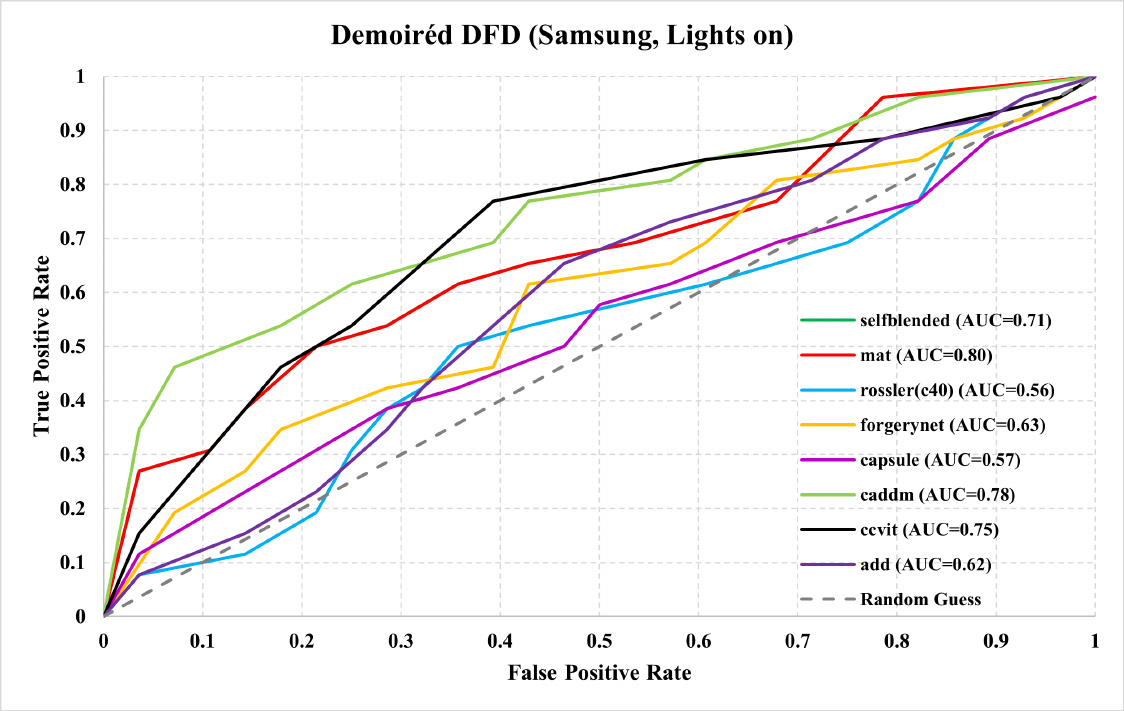}
    \includegraphics[width=0.45\textwidth]{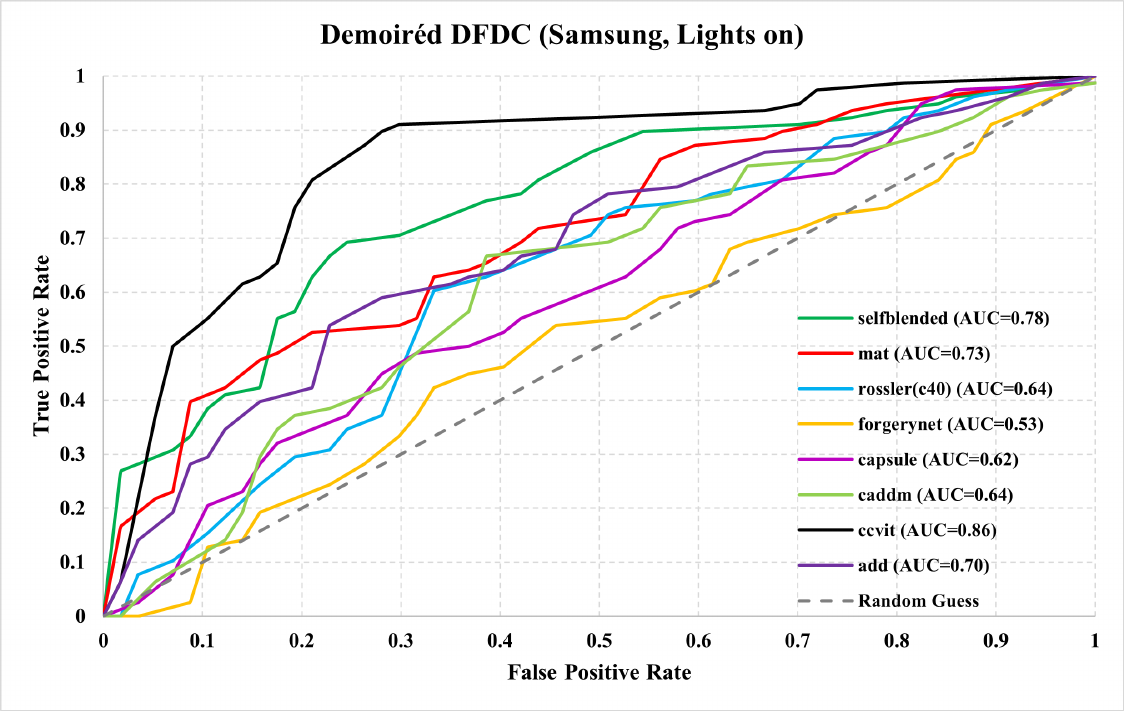}
    \includegraphics[width=0.45\textwidth]{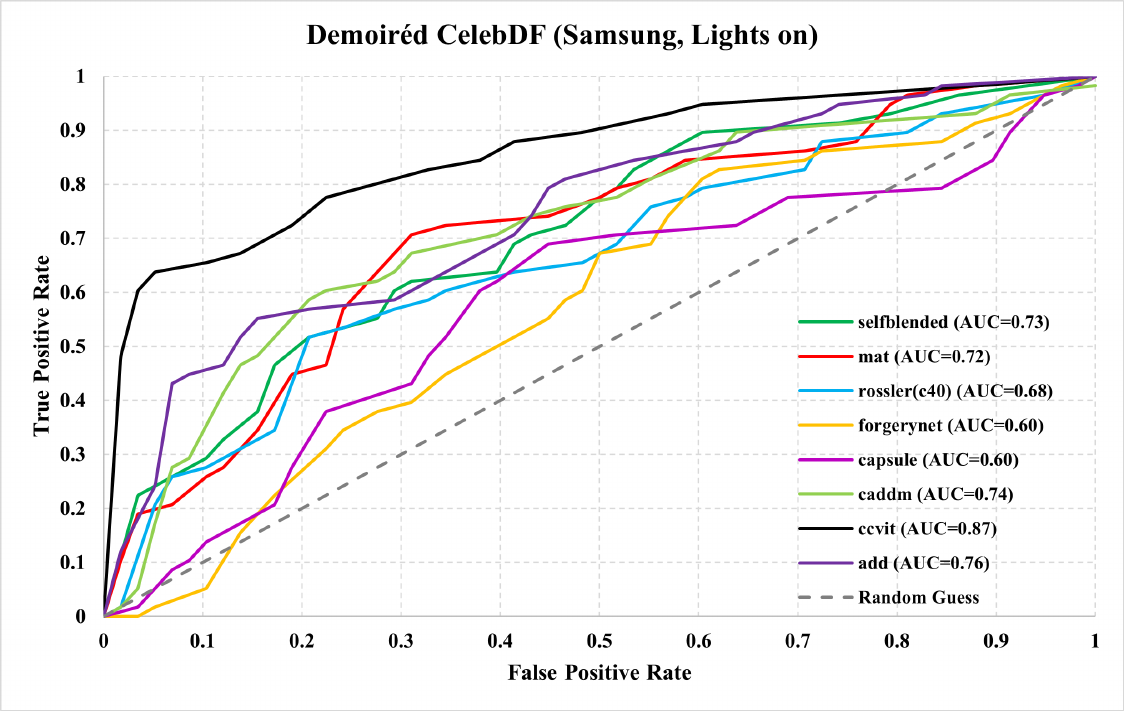}
    \includegraphics[width=0.45\textwidth]{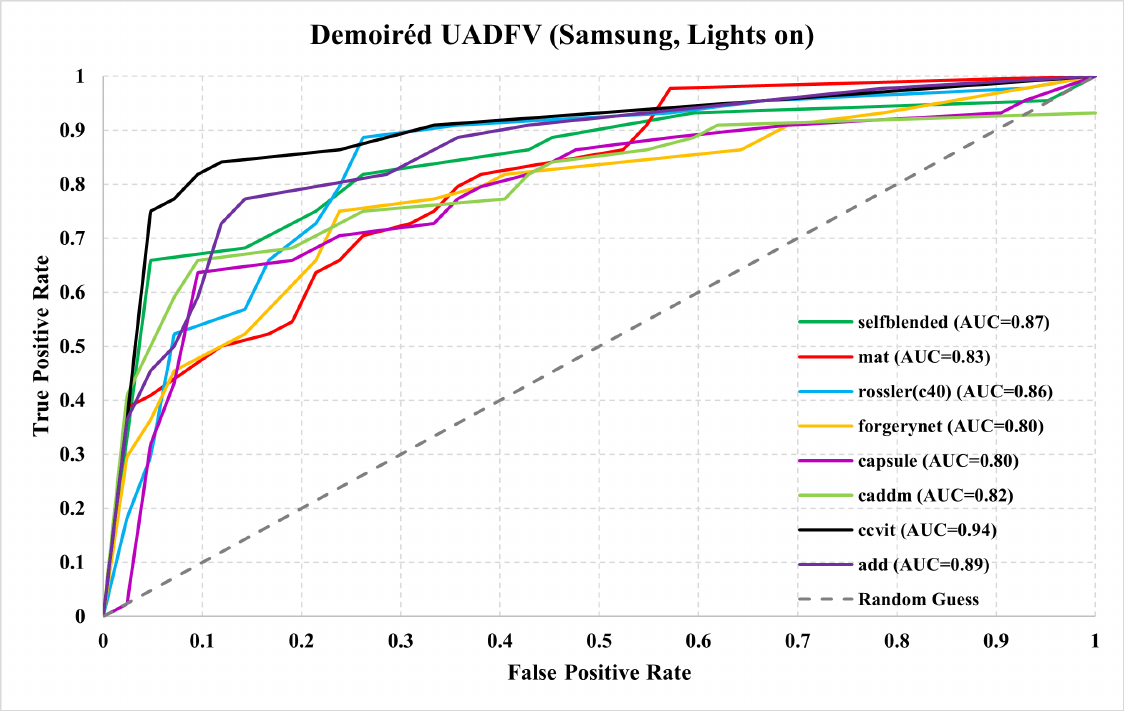}
    \caption{\textbf{\textsc{Performance on Demoiré Datasets (BenQ Monitor, Lights On) }}For the FaceForensics++ dataset, the CCViT model achieves the highest AUC of 0.82\%, closely followed by the MAT model with 0.81\%. MAT leads with an AUC of 0.80\% in the DFD dataset, while CCViT follows with 0.75\%. In the DFDC dataset, CCViT outperforms other models with an AUC of 0.86\%. For the CelebDF dataset, CCViT again leads with an AUC of 0.87\%. In the UADFV dataset, CCViT excels with an AUC of 0.94\%, followed by the ADD model with 0.89\%.}
\end{figure}
\newpage
\subsubsection{Lights Condition: OFF}

\begin{figure}[h]
    \centering
    \includegraphics[width=0.49\textwidth]{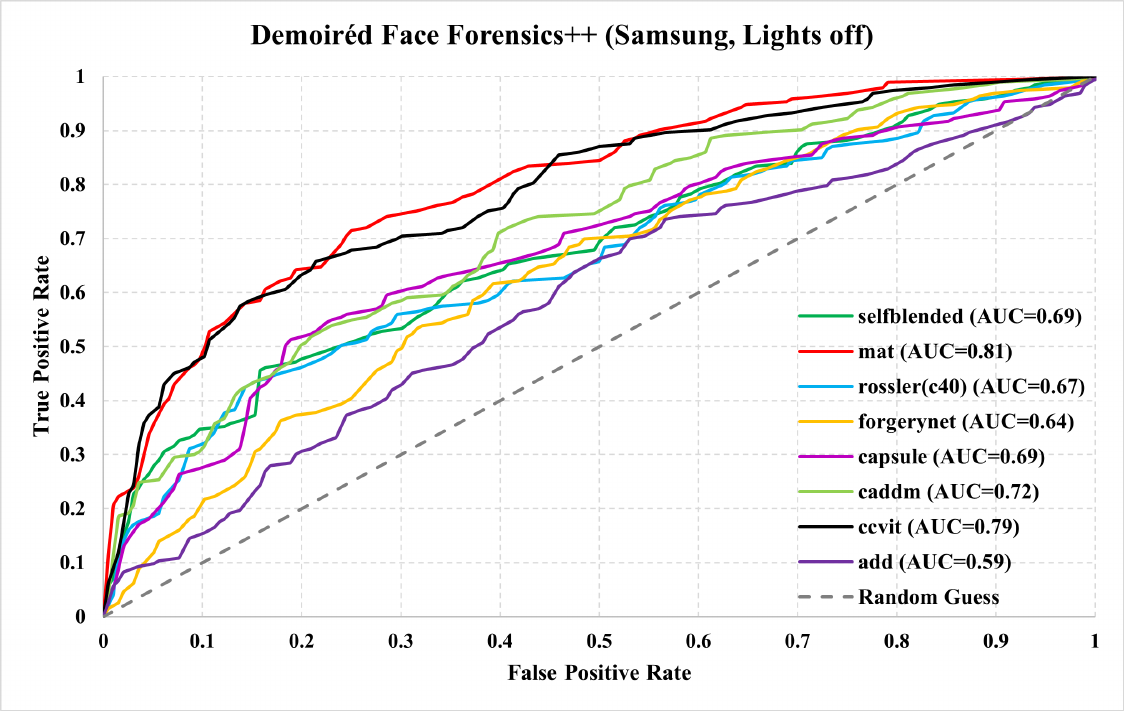}
    \includegraphics[width=0.49\textwidth]{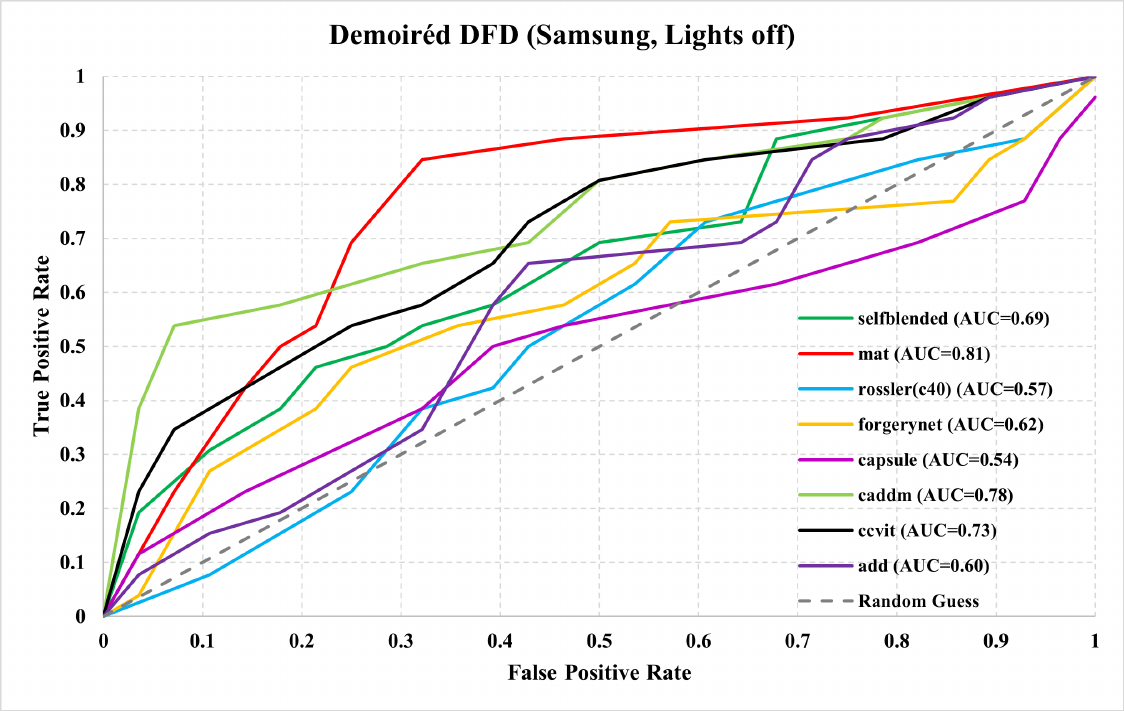}
    \includegraphics[width=0.49\textwidth]{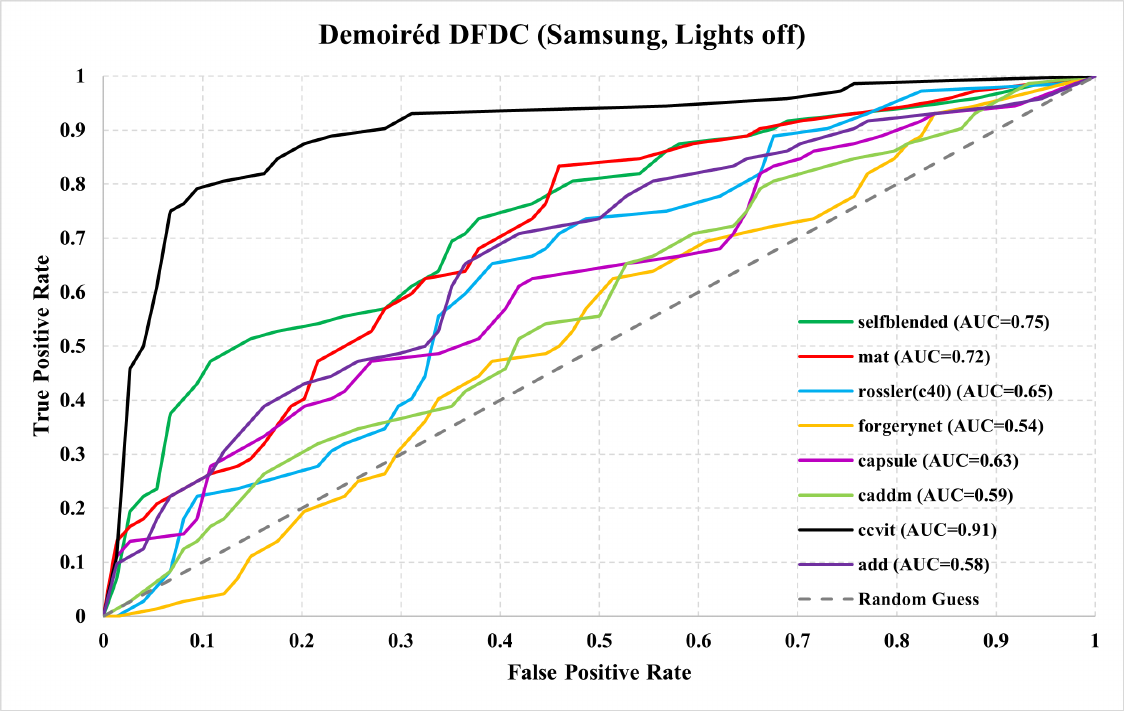}
    \includegraphics[width=0.49\textwidth]{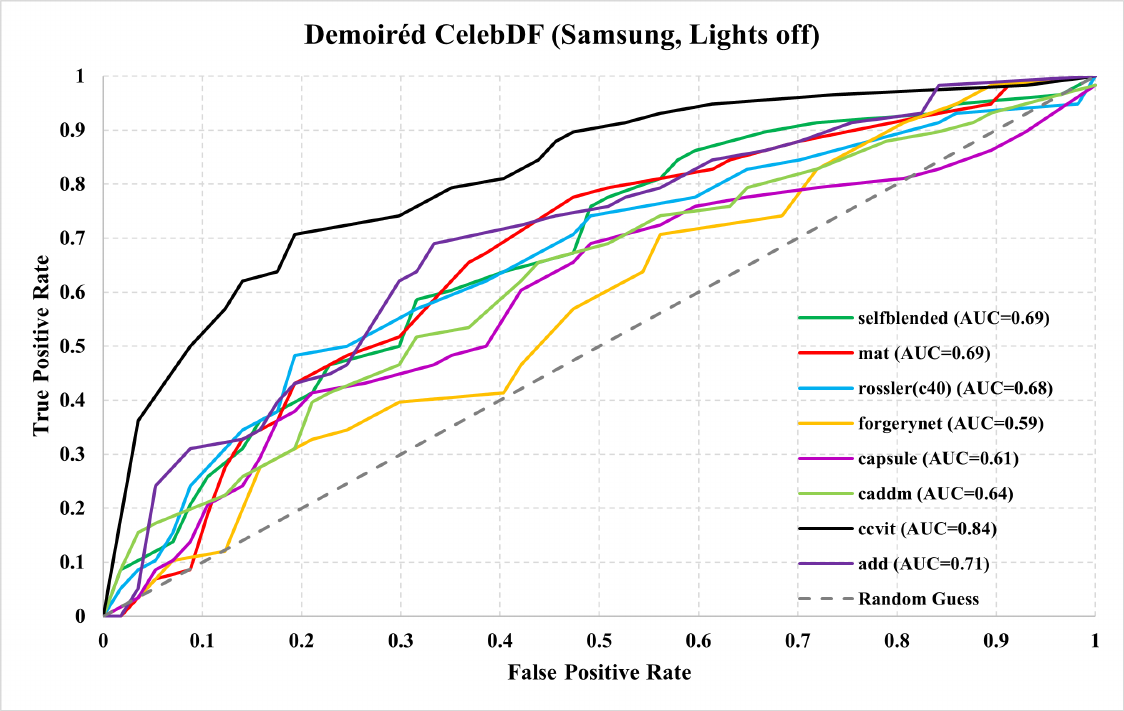}
    \includegraphics[width=0.49\textwidth]{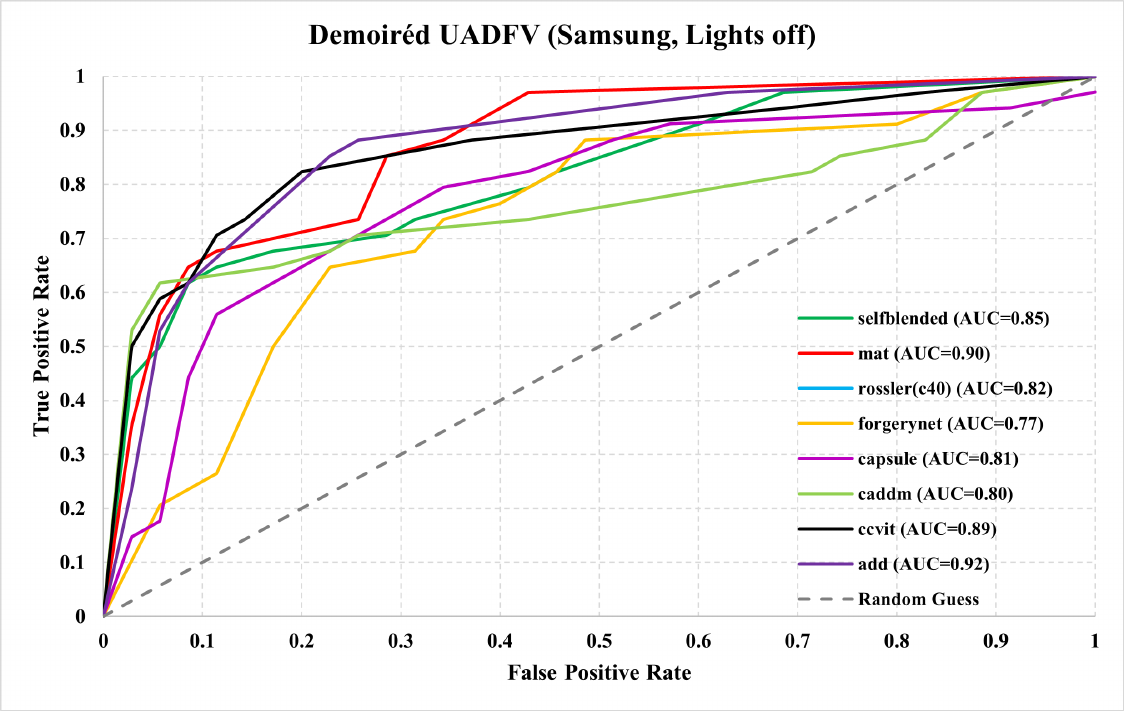}
    \caption{\textbf{\textsc{Performance on Demoiré Datasets (BenQ Monitor, Lights Off) }}In the FaceForensics++ dataset, MAT and CCViT demonstrate the highest AUC scores of 0.81\% and 0.79\%, respectively, indicating strong detection capabilities. MAT and CCViT again lead with AUC scores of 0.81\% and 0.73\% for the DFD dataset. In the DFDC dataset, CCViT performs best with an AUC of 0.91\%, significantly outperforming other methods. The CelebDF dataset shows CCViT as the top performer with an AUC of 0.84\%, followed by ADD at 0.71\%. For the UADFV dataset, ADD achieves the highest AUC of 0.92\%, with MAT following at 0.90\%. These results suggest that MAT and CCViT are consistently effective across various datasets.}
\end{figure}
\newpage
\subsection{Camera: iPhone 13}
\subsubsection{Lights Condition: ON}
\begin{figure}[h]
    \centering
    \includegraphics[width=0.49\textwidth]{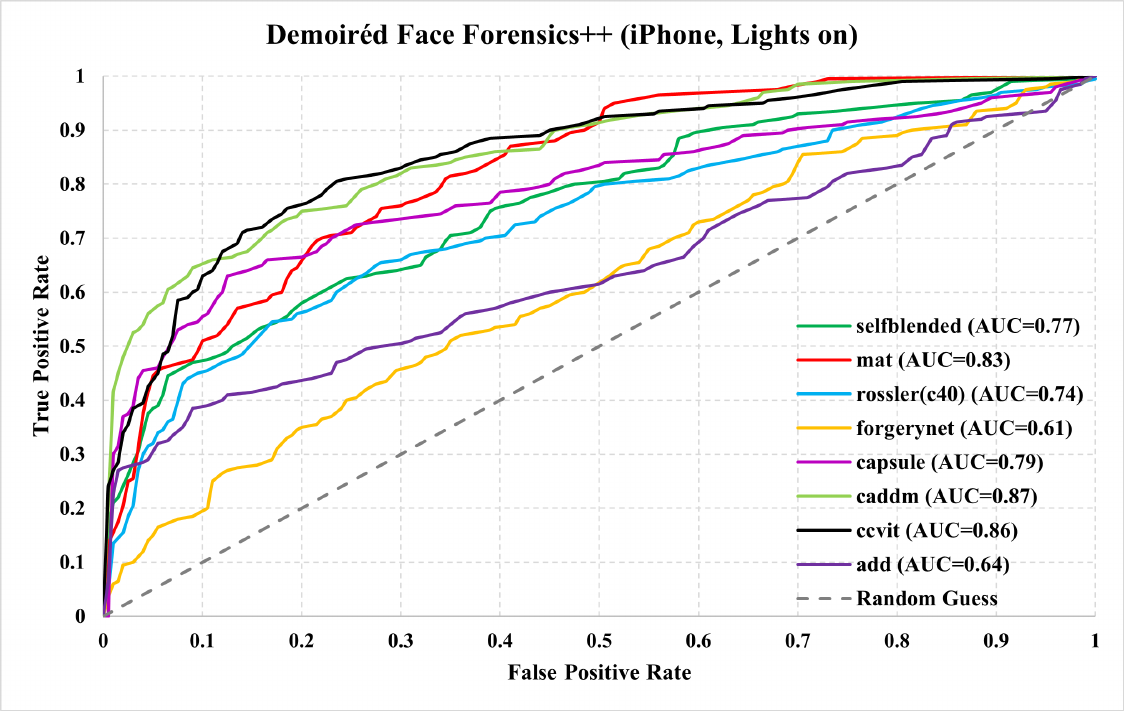}
    \includegraphics[width=0.49\textwidth]{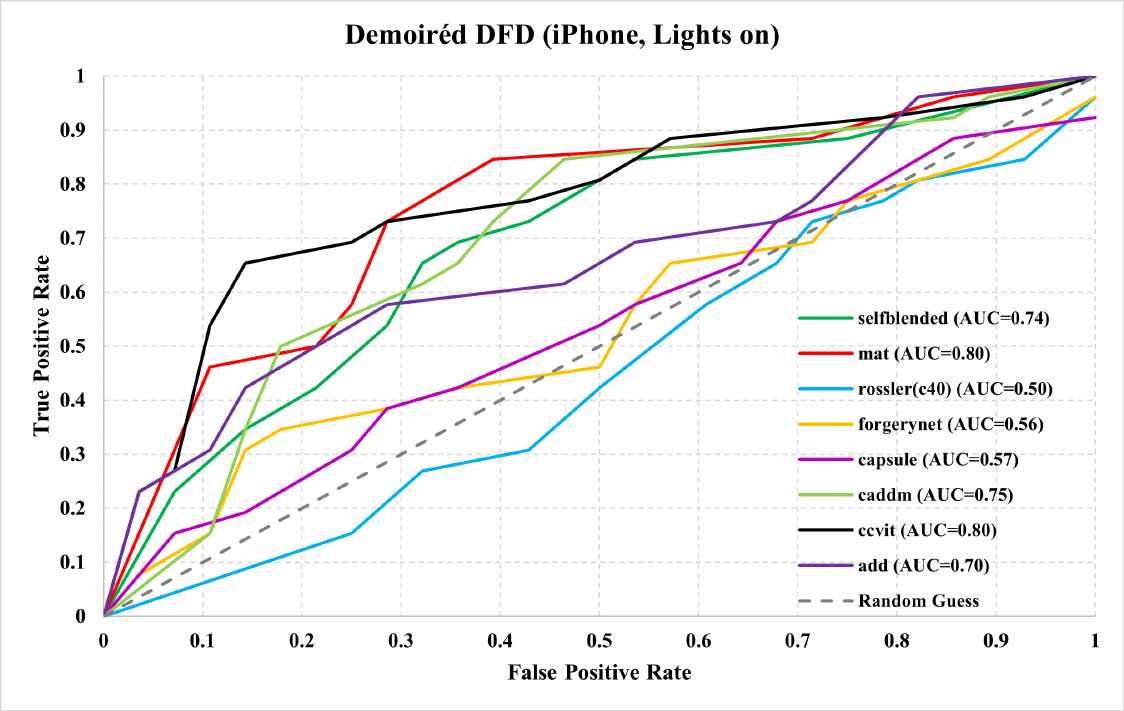}
    \includegraphics[width=0.49\textwidth]{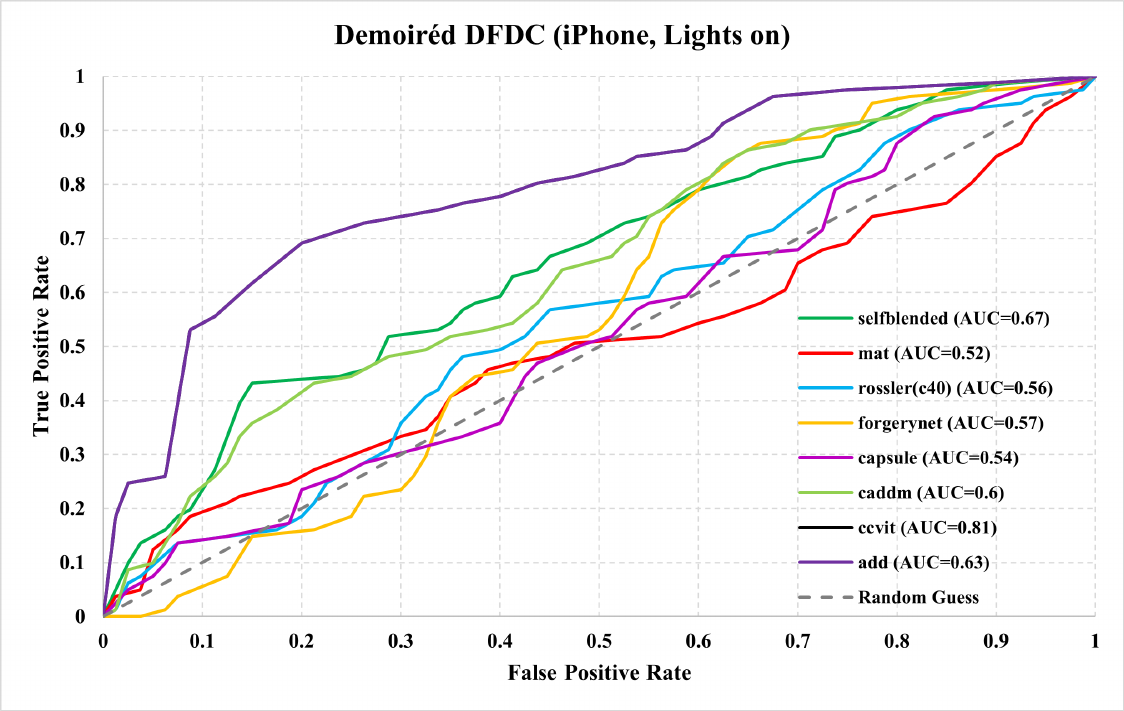}
    \includegraphics[width=0.49\textwidth]{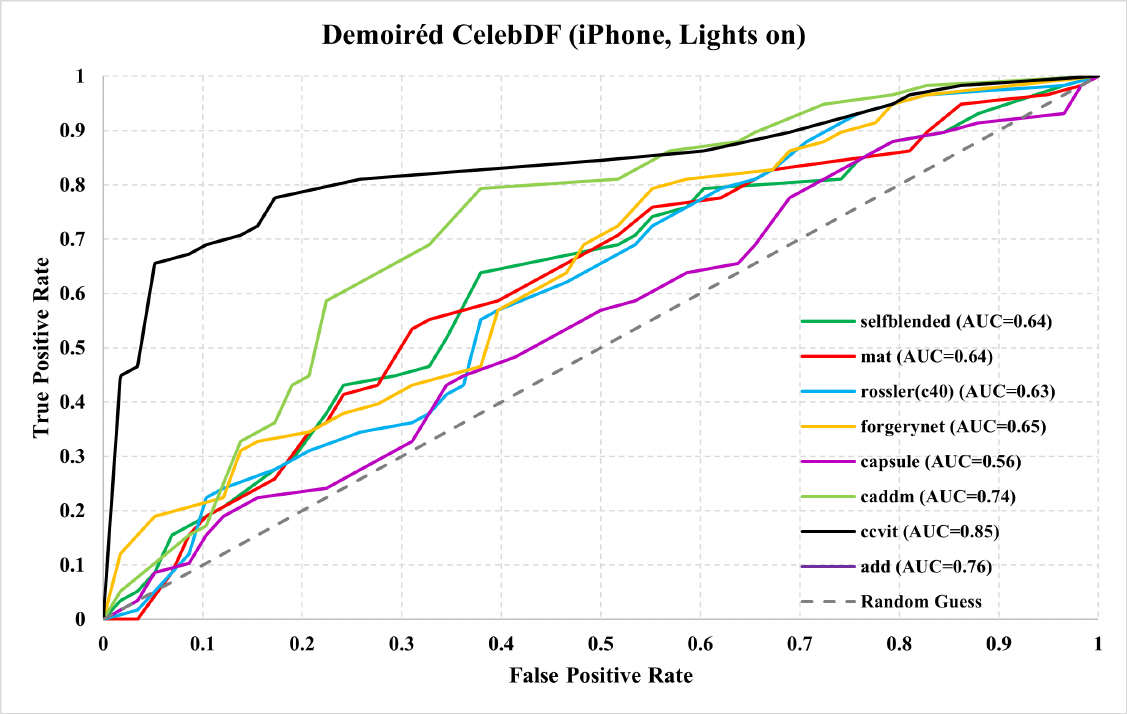}
    \includegraphics[width=0.49\textwidth]{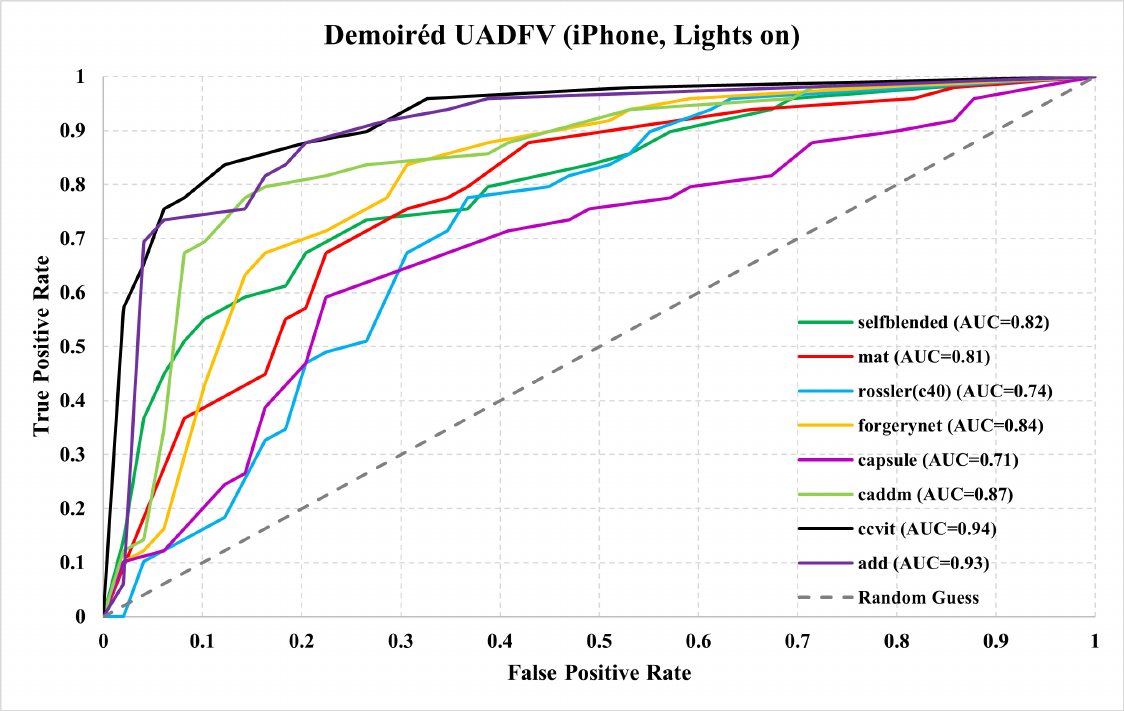}
    \caption{\textbf{\textsc{Performance on Demoiré Datasets (BenQ Monitor, Lights On)}} In the FaceForensics++ dataset, the CADDM method achieved the highest AUC of 0.87\%, followed by the CCViT method at 0.86\%. The MAT method leads with an AUC of 0.80\% for the DFD dataset, closely followed by CCViT with 0.80\%. In the DFDC dataset, CCViT performs best with an AUC of 0.81\%, with CADDM trailing at 0.60\%. For the CelebDF dataset, CCViT also excels with an AUC of 0.85\%, followed by the ADD method at 0.76\%. Lastly, in the UADFV dataset, CCViT demonstrates superior performance with an AUC of 0.94\%, while the ADD method achieves an AUC of 0.93\%. }
\end{figure}
\newpage
\subsubsection{Lights Condition: OFF}
\begin{figure}[h]
    \centering
    \includegraphics[width=0.49\textwidth]{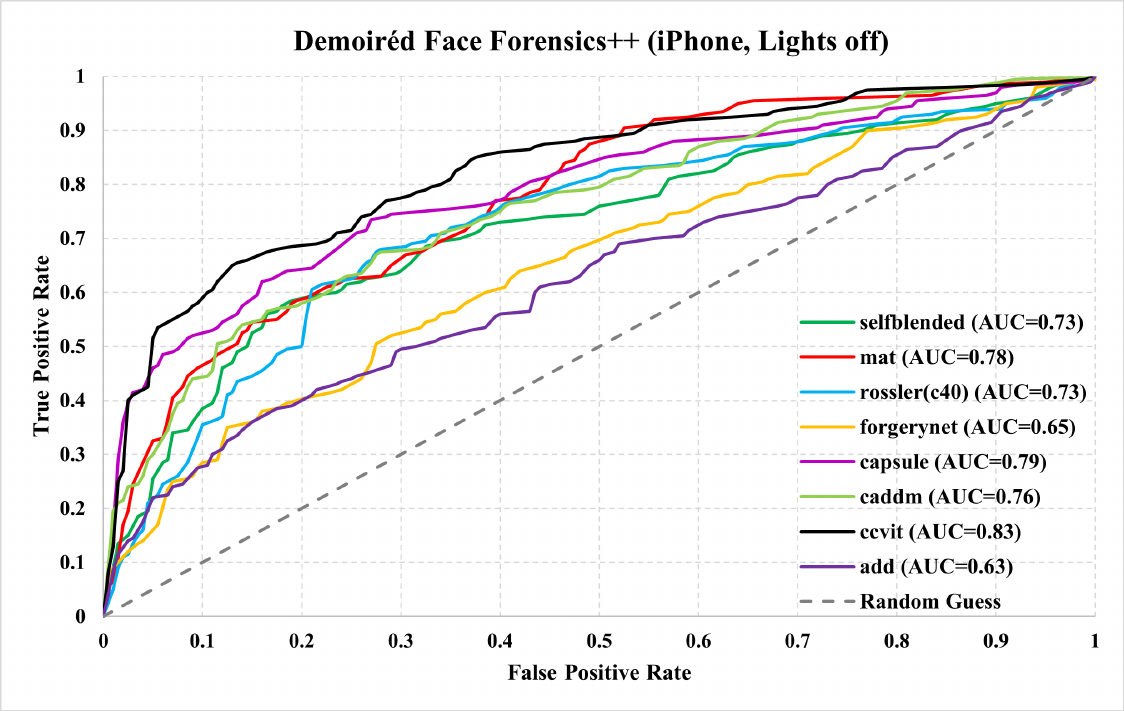}
    \includegraphics[width=0.49\textwidth]{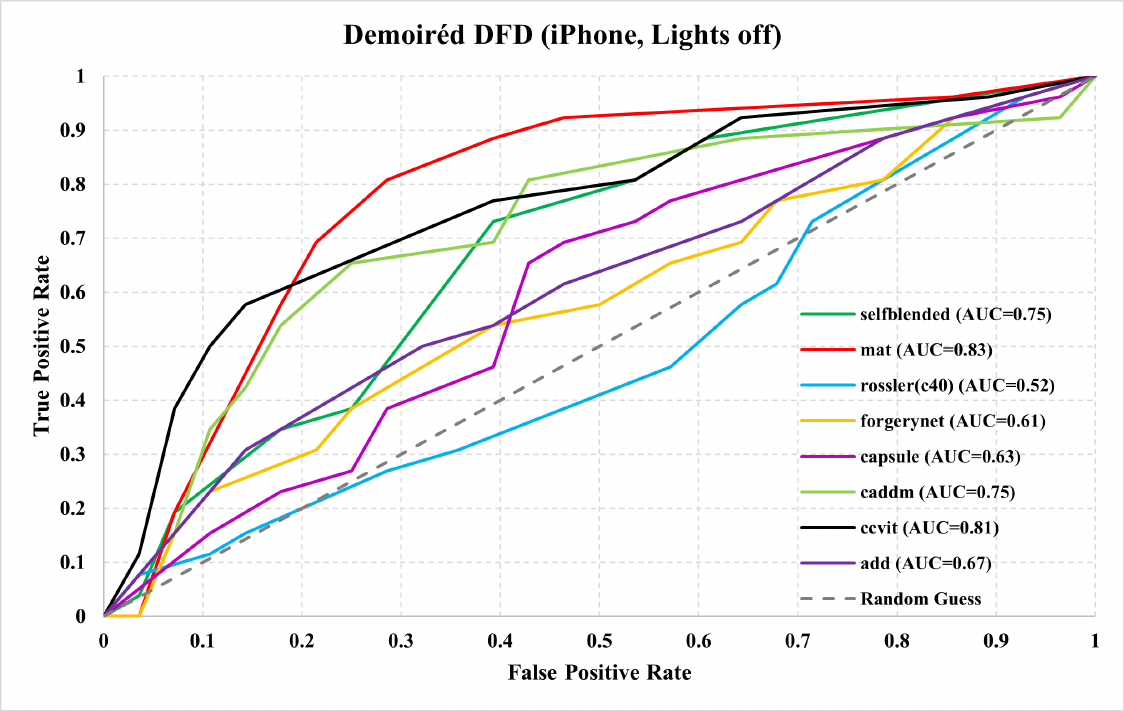}
    \includegraphics[width=0.49\textwidth]{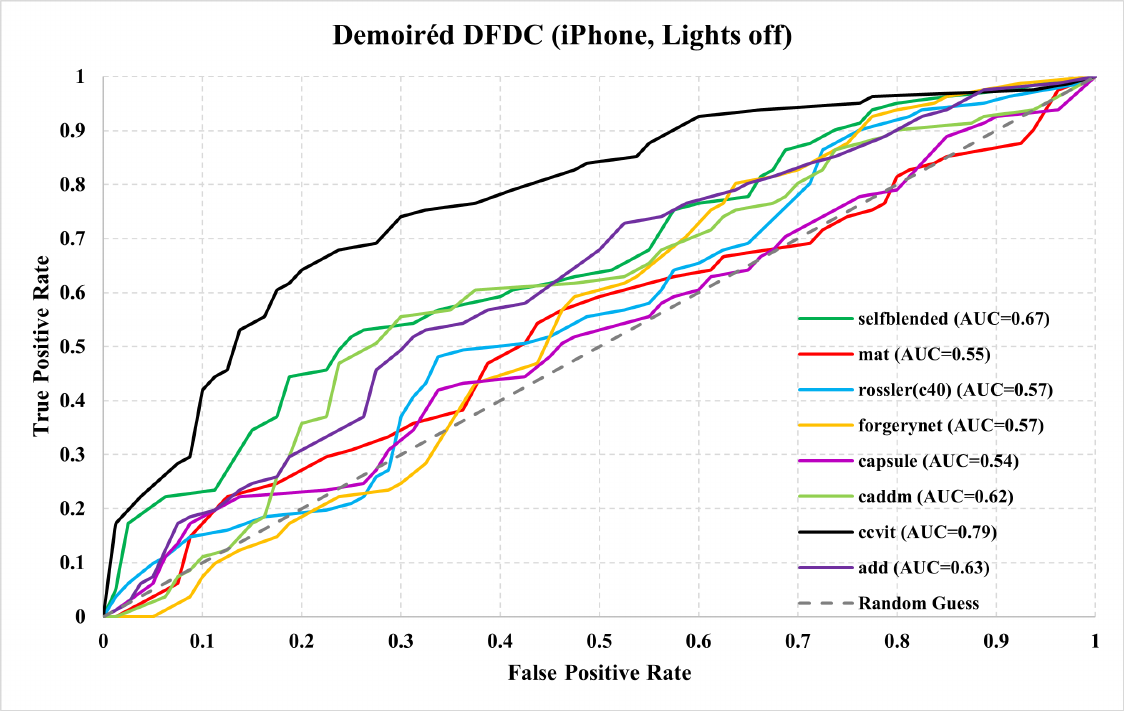}
    \includegraphics[width=0.49\textwidth]{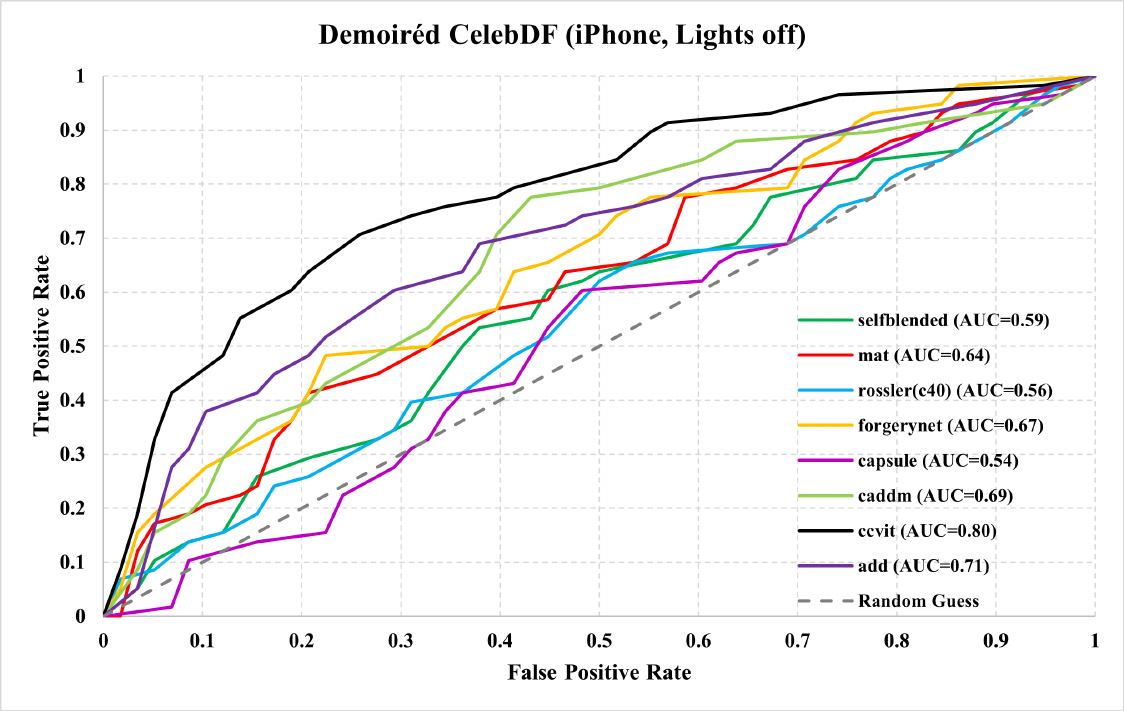}
    \includegraphics[width=0.49\textwidth]{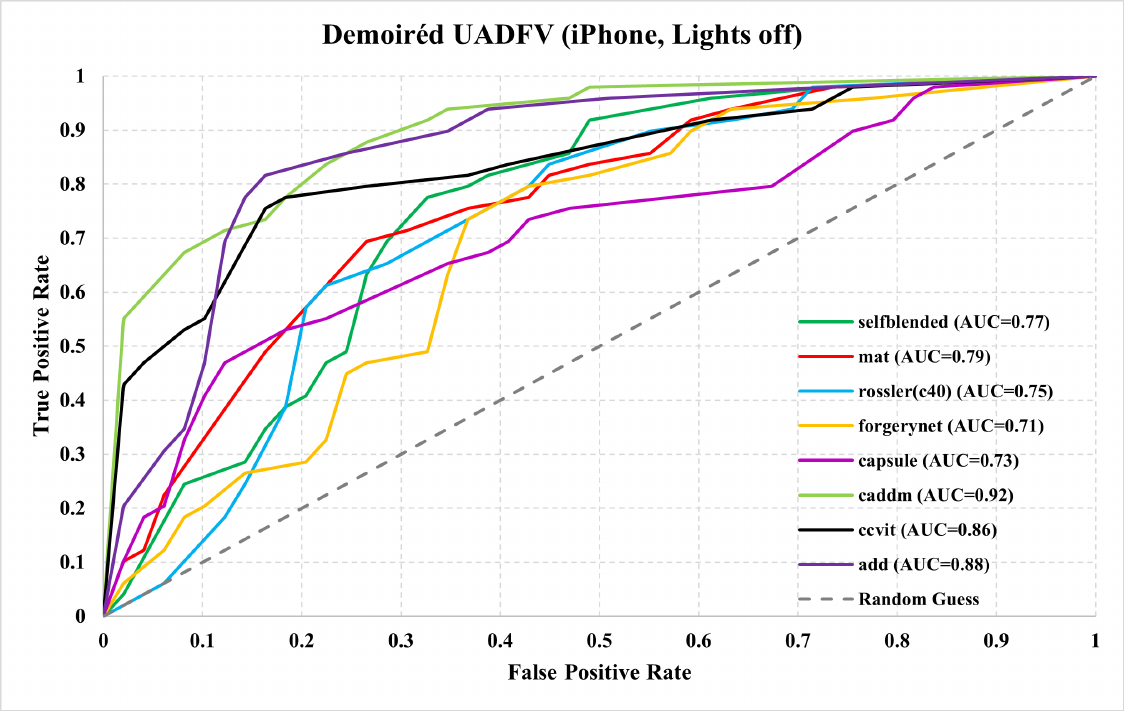}
    \caption{\textbf{\textsc{Performance on Demoiré Datasets (BenQ Monitor, Lights Off) }}The CCViT method consistently attained the highest AUC scores across various datasets, showcasing its superior performance. In detail, CCViT achieved AUC scores of 0.83\% for FaceForensics++, 0.81\% for DFD, 0.79\% for DFDC, 0.80\% for CelebDF, and 0.86\% for UADFV datasets. The MAT and CADDM methods also exhibited strong performance, although their effectiveness varied depending on the dataset.}
\end{figure}

\newpage
\section{Performance after Denoising on DMF Dataset --- ROC Curve}
\subsection{Camera: iPhone 13—Lights Condition: ON}
\begin{figure}[h!]
    \centering
    \includegraphics[width=0.75\textwidth]{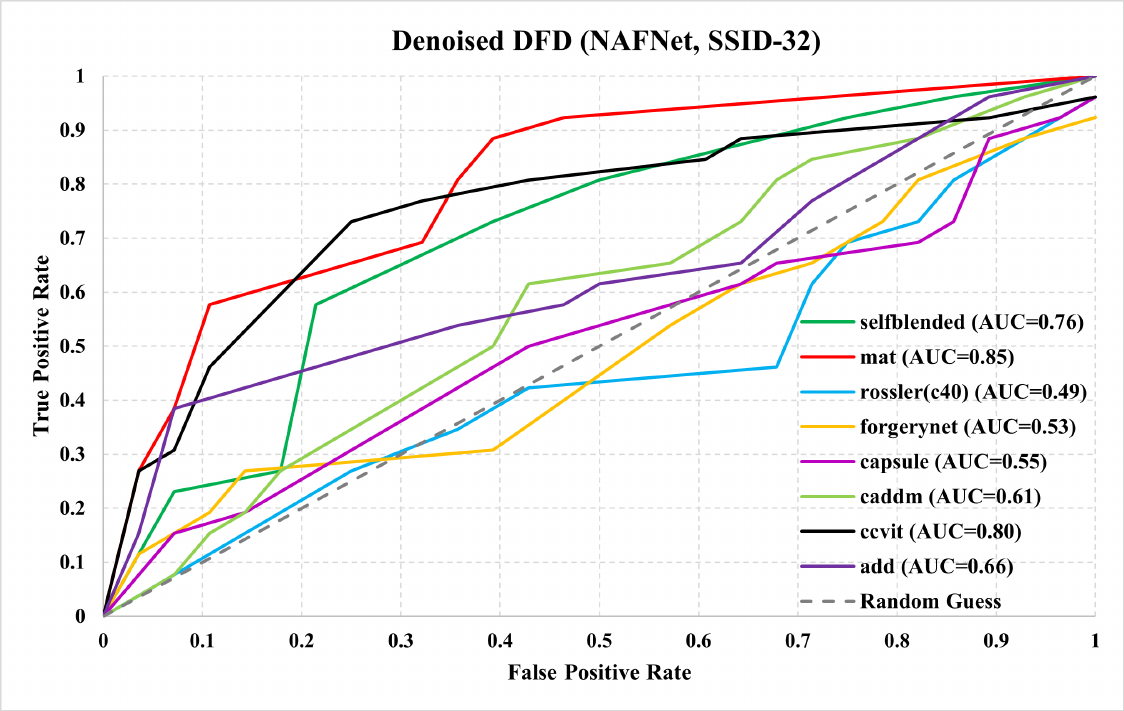}
    \includegraphics[width=0.75\textwidth]{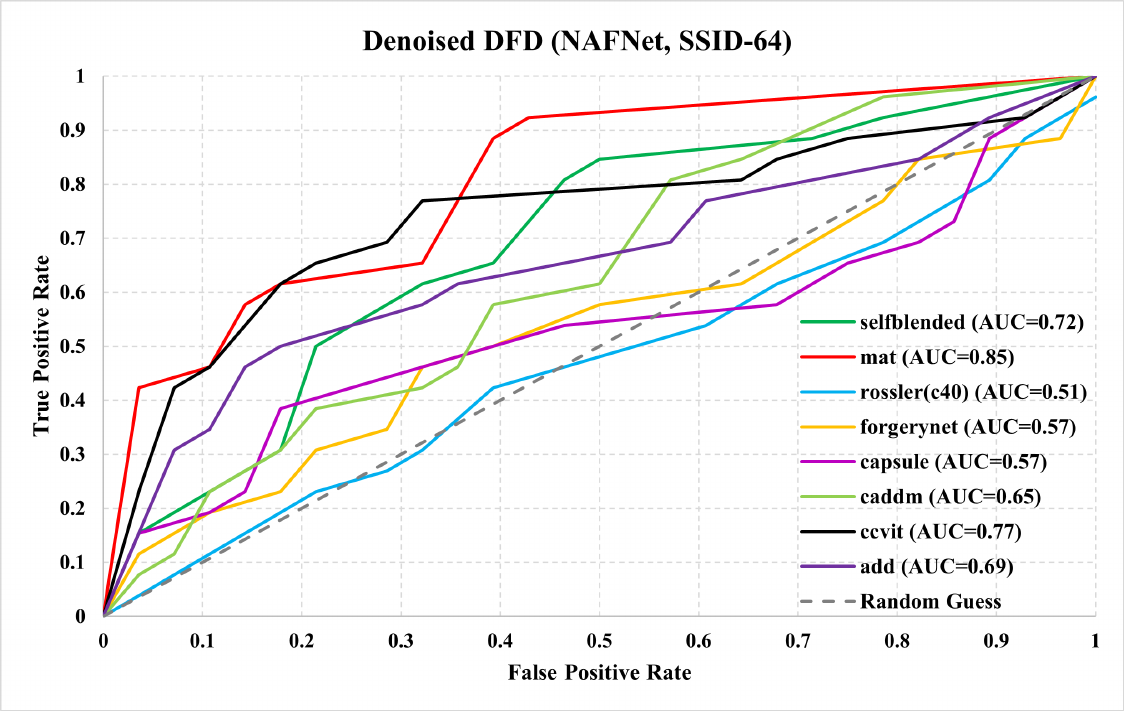}
    \caption{\textbf{\textsc{Performance on Denoise Datasets (BenQ Monitor, Lights ON) }}When tested on different weights of NAFNet (SSID-32/64) to remove the Moiré pattern, the MAT approach consistently achieved a high AUC of 0.85\%. Based on this observation, it is evident that even with applying some denoise to the Moiré Pattern, MAT can keep its AUC in accurately predicting synthetic images on the DFD dataset. CCViT achieved the second-highest Area Under the Curve (AUC) score for both weights at 0.80\% and 0.77\%. Rossler et al. (C40) exhibited worse performance of 0.49\% and 0.51\% when subjected to the weight testing of NAFNet (SSID-32/64).
  }
\end{figure}
\newpage
\section{Performance after Deblurring on DMF Dataset --- ROC Curve}
\subsection{Camera: iPhone 13—Lights Condition: ON}
\begin{figure}[h!]
    \centering
    \includegraphics[width=0.75\textwidth]{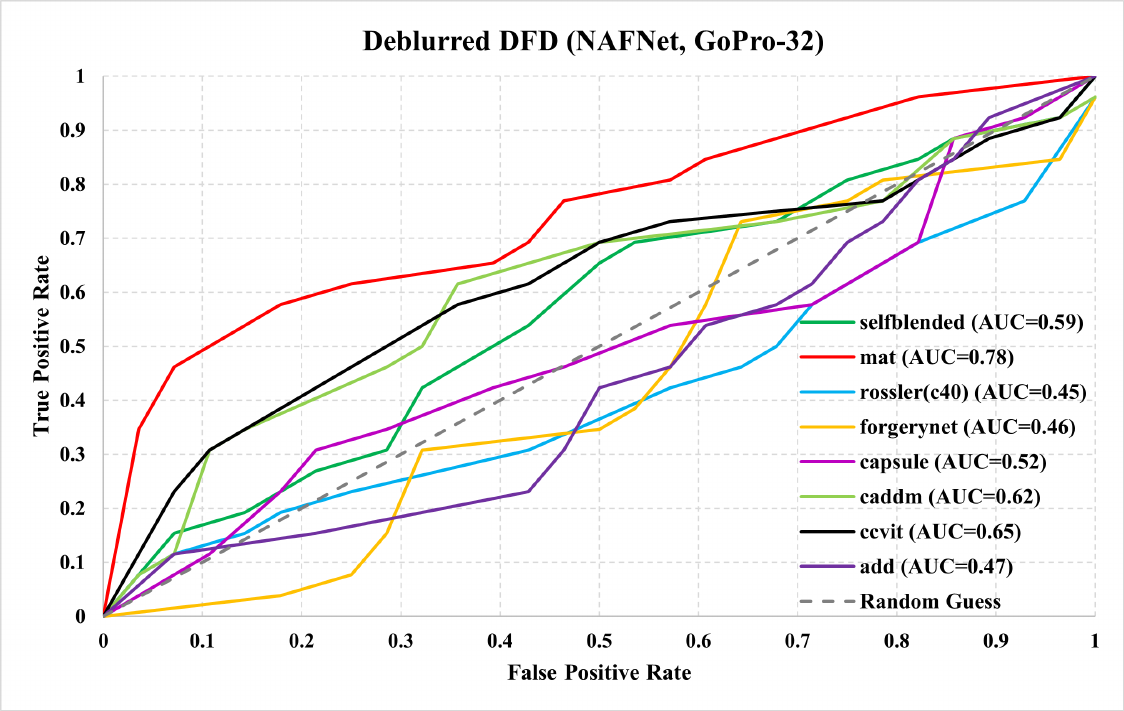}
    \includegraphics[width=0.75\textwidth]{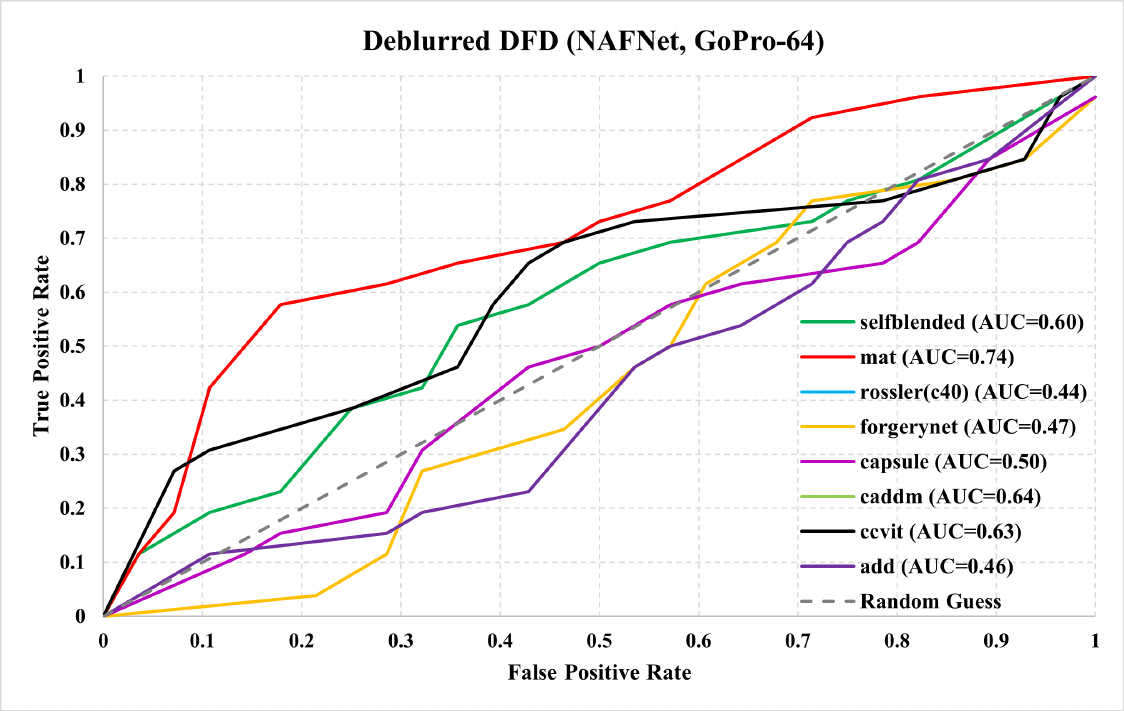}
    \caption{\textbf{\textsc{Performance on Deblur Datasets (BenQ Monitor, Lights ON) }}When deblurring is used to eliminate the Moiré pattern, MAT demonstrates lower performance compared to denoising at 0.78\% and 0.74\%. However, the effectiveness of deblurring is noticeably reduced, suggesting that it also eliminates crucial elements that deepfake detectors rely on for predictions. The AUC of CCViT on NAFNet (GoPro-32) is the second highest at 0.65\%, whereas CADDM achieves the highest AUC of 0.64\% when tested with NAFNet (GoPro-64). ADD exhibited the lowest performance in both cases, with rates of 0.47\% and 0.46\%, respectively. 
 }
    \label{fig:label-h}
\end{figure}

\section{Experimental Environment and Training Setup}

We used TITAN RTX, RTX A5000, and RTX 3090 GPUs to preprocess and test image and video detectors and demoiréing methods. For both finetuning and retraining, we modified only the batch size (16) and number of epochs (10), while keeping the original optimizers, learning rates, and other hyperparameters as specified by the authors. Specifically, we used optimizer Adam with a learning rate of (1e-3) for Rossler, and AdamW for both MAT (1e-4) and CADDM (1e-3), ensuring consistency with the original training setups.

\newpage

\section{Visual Analysis of Moiré Pattern on Frequency Spectrum}
\begin{figure}

    \centering
    \includegraphics[width=1\textwidth]{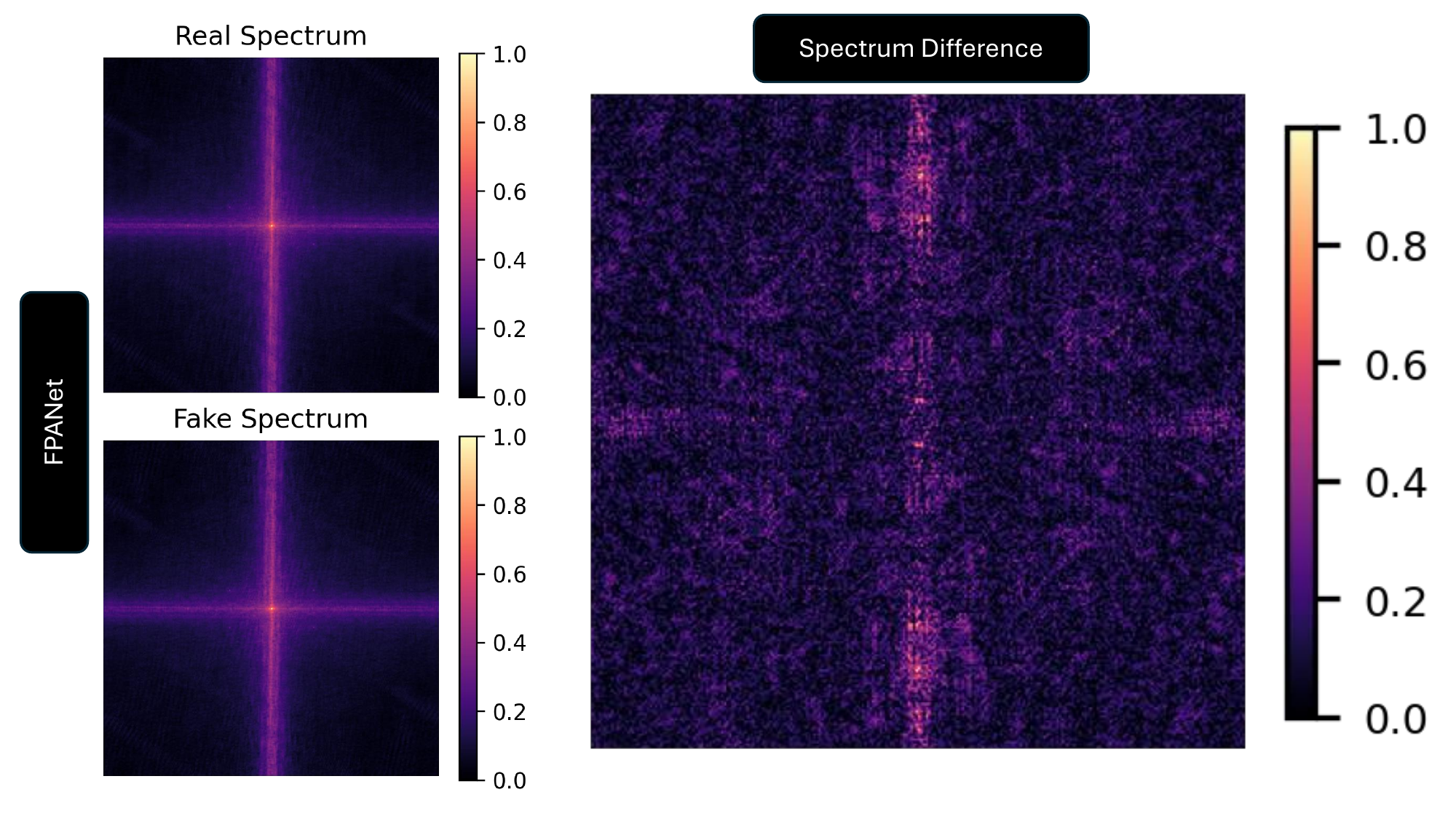}
    \vspace{1em}
    \includegraphics[width=1\textwidth]{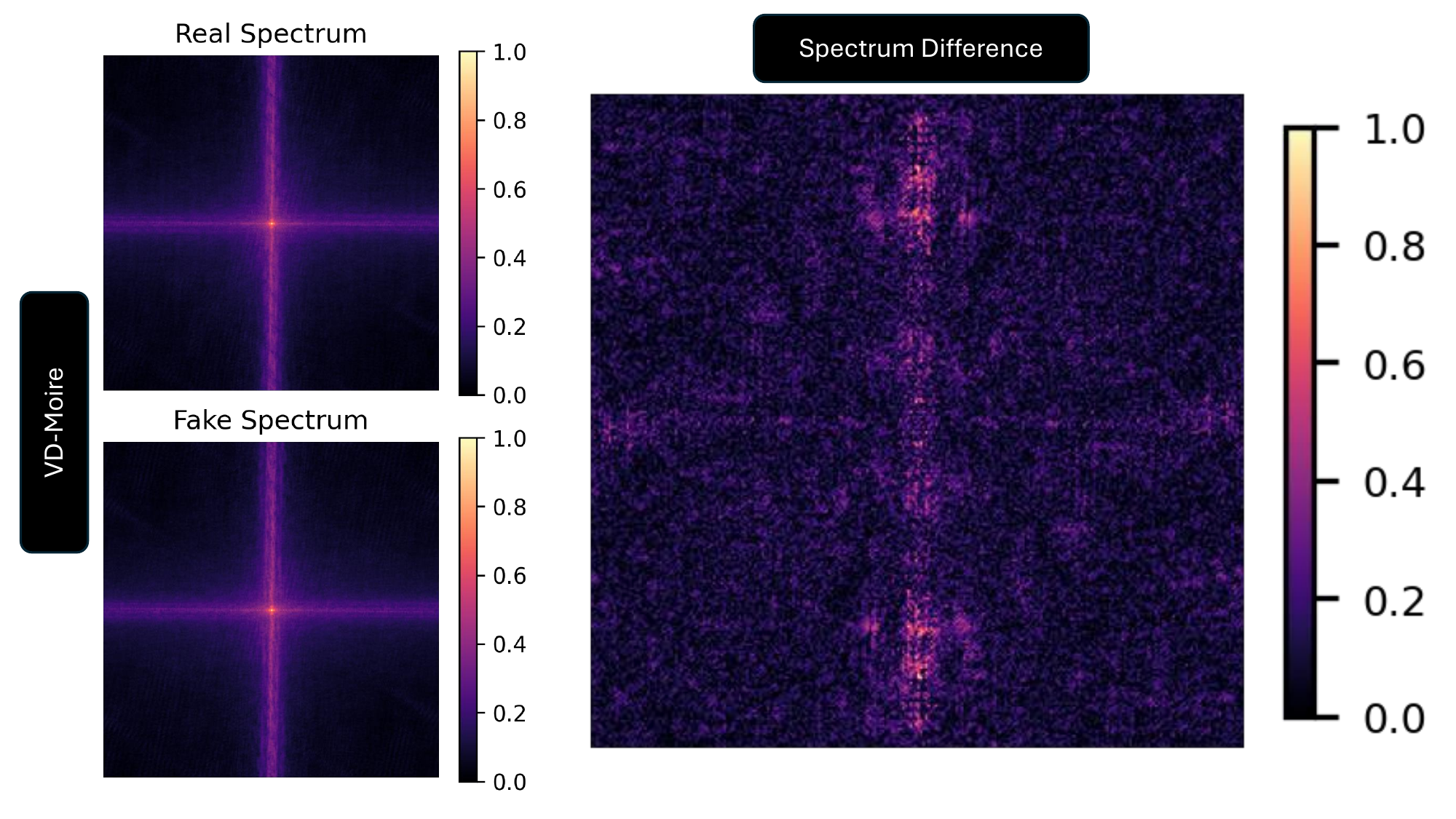}
    \caption{\textbf{\textsc{Frequency spectra of FPAnet and VD-Moire demoiréing}}}
    \label{fig:og_moire_frequency}
\end{figure}

The figures in~\autoref{fig:og_moire_frequency} illustrate how Moiré patterns affect the frequency spectrum. Compared to the original spectrum, Moiré-affected real and fake samples exhibit noticeable distortions, disrupting the clear frequency structure. This degradation helps explain the drop in performance of both image- and video-based deepfake detectors. We also evaluated whether video demoiréing methods could mitigate this impact. As shown in~\autoref{table:videoDemoiring}, methods like LipForensics slightly improved after applying demoiréing. However, other methods, such as AltFreezing and FTCN, exhibited further accuracy decline, even after processing with FPANet, suggesting that these models still struggled to recover accurate predictions despite Moiré removal.




    

\begin{figure}[h]
    \centering
    \includegraphics[width=1\textwidth]{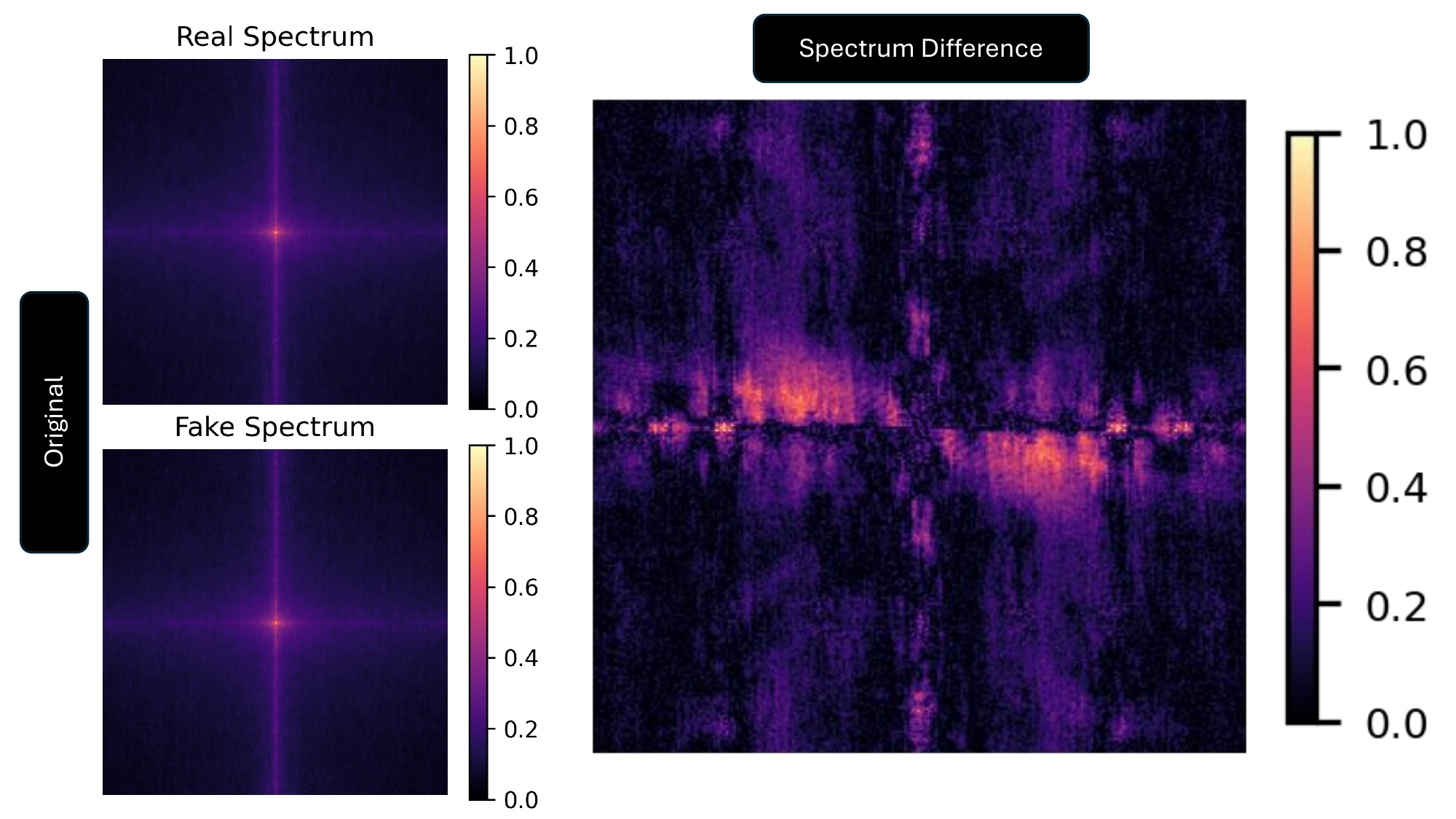}
    \vspace{1em}
    \includegraphics[width=1\textwidth]{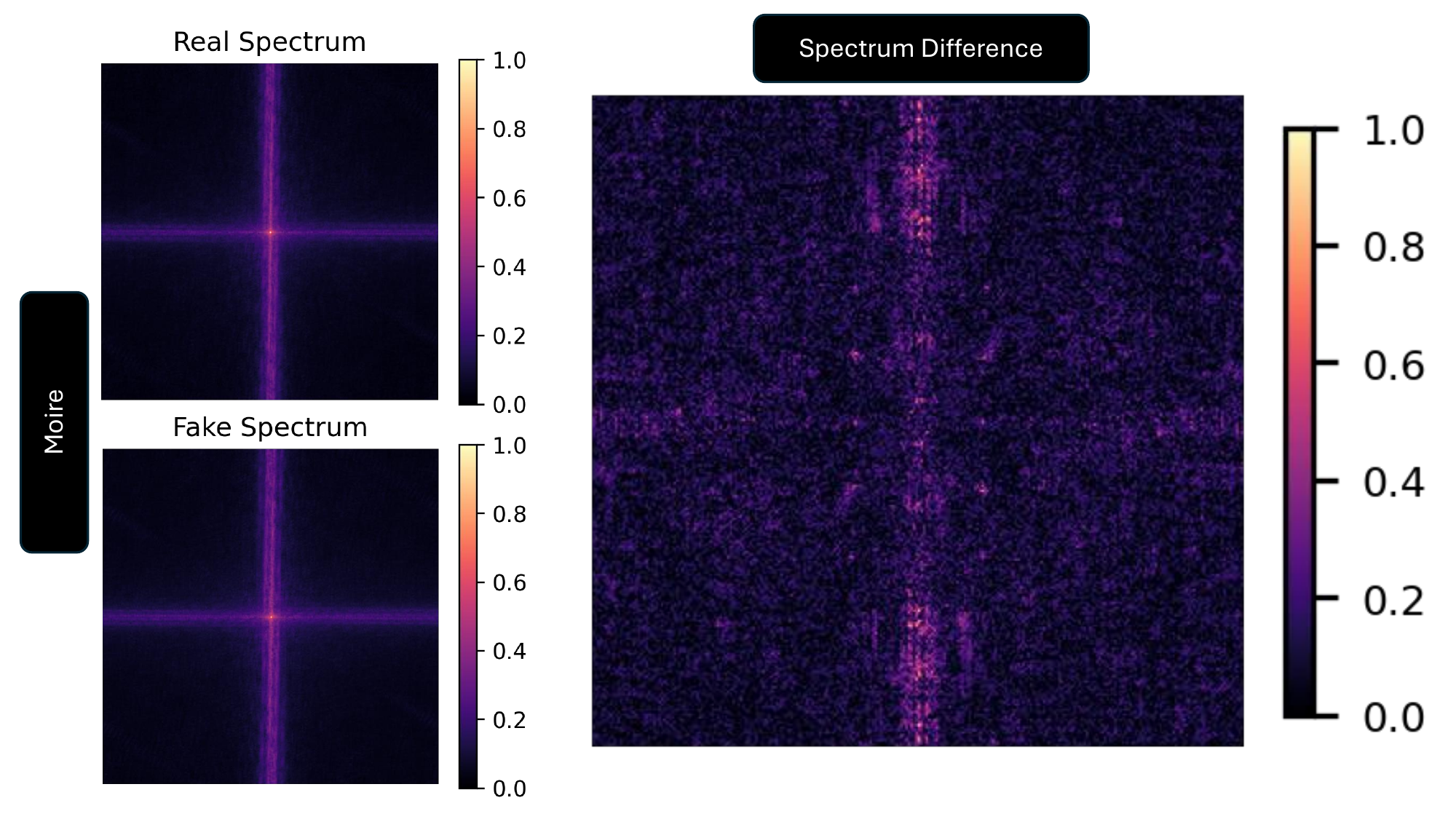}
    {\textbf{\textsc{Frequency spectra of Original and Moire pattern}}}
\end{figure}

\newpage
\section{Visual Analysis of Moiré Pattern in Generative Methods}
\begin{figure}[h!]
    \centering
    \includegraphics[width=0.90\textwidth]{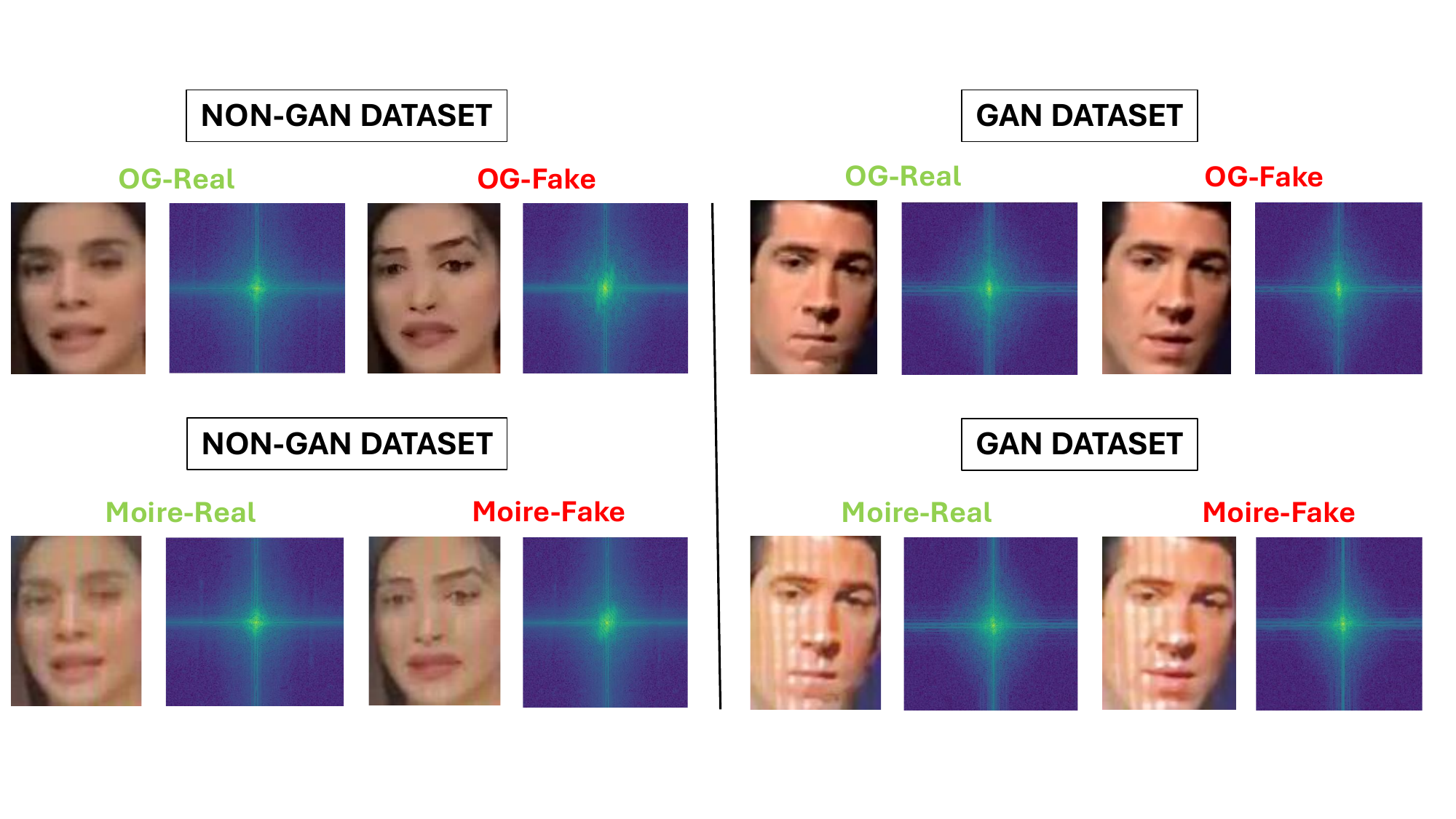}
    \caption{\textbf{\textsc{GAN and Non-GAN images exhibit distinct characteristics under frequency analysis, with Moiré patterns displaying significantly different low-frequency patterns compared to the original images in both datasets.}} 
 }
    \label{fig:gan_nongan}
\end{figure}

\begin{figure}[h!]
    \centering
    \includegraphics[width=0.90\textwidth]{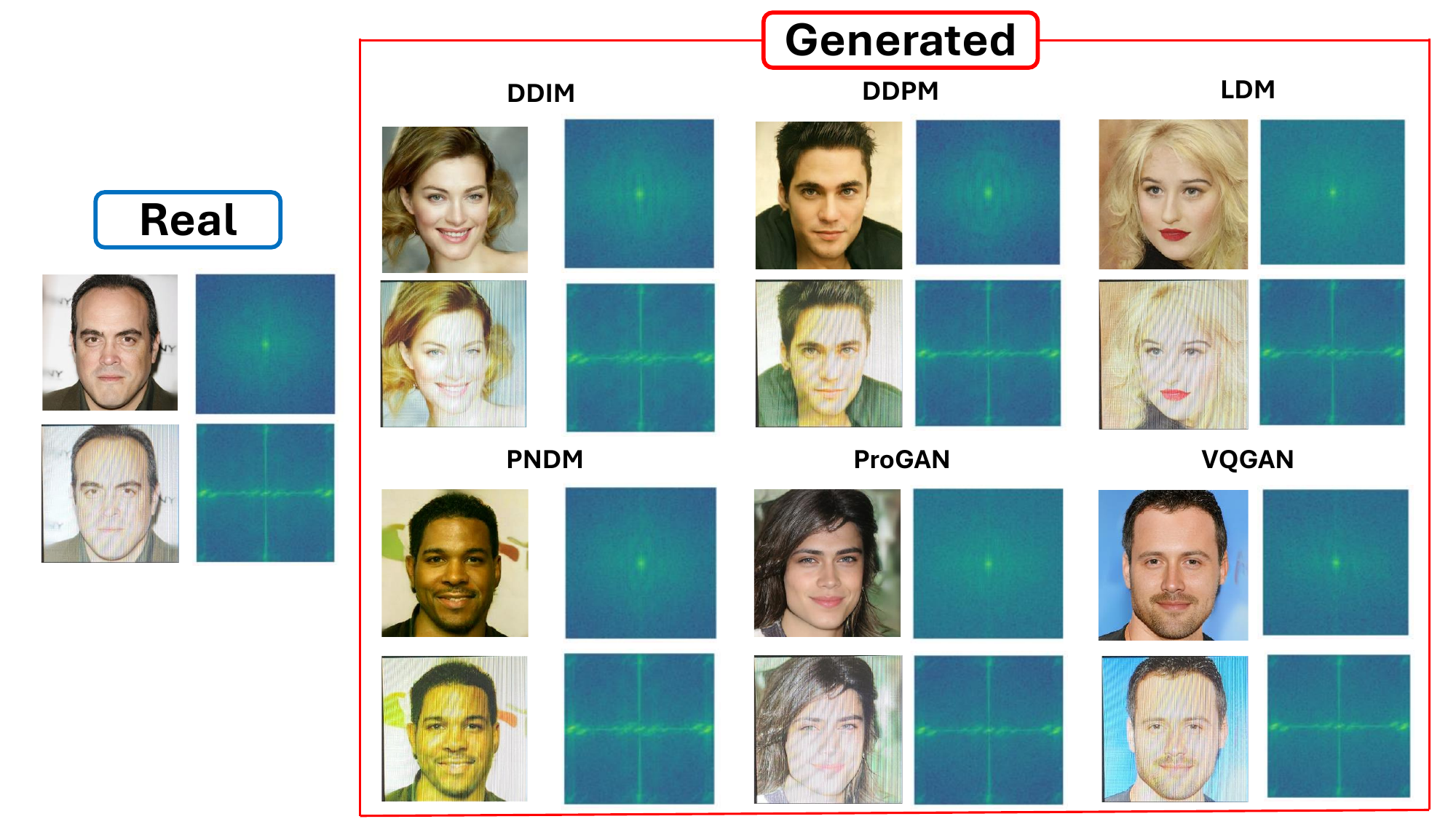}
    \caption{\textbf{\textsc{Images from six generative methods were captured using the Samsung S22 Plus camera, revealing the presence of Moiré patterns. Frequency analysis of these Moiré-impacted images indicated significant differences compared to their original counterparts.}} 
 }
 \label{fig:generative_methods}
\end{figure}

The figures below illustrate the impact of Moiré patterns across generative methods.~\autoref{fig:gan_nongan} compares the frequency characteristics of GAN and Non-GAN images, while~\autoref{fig:generative_methods} shows Moiré patterns in images from six generative methods.
\subsection{Moiré Pattern on GAN vs Non-GAN Generated Datasets}

\subsection{Moiré Pattern on Multiple Generative Methods}

\newpage

\section{Influence of Side-Angle Captures on Visual Artifacts and Model Performance}
\label{sec:new_angles}

\begin{figure}[h!]
    \centering

    \begin{subfigure}[b]{0.32\textwidth}
        \centering
        \includegraphics[width=\textwidth, trim=0cm 4cm 0cm 4cm, clip]{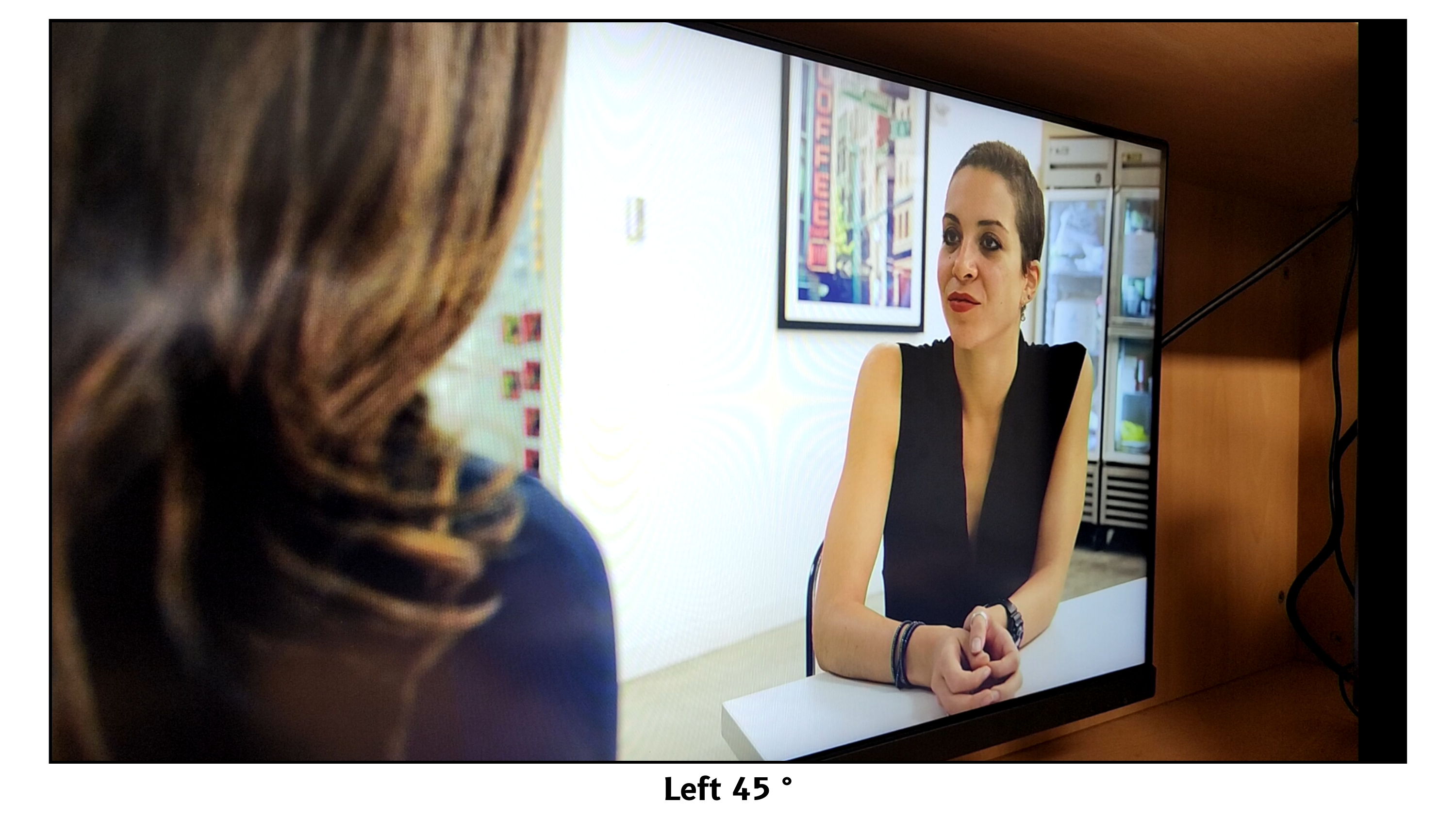}
        \caption{\textsc{Left 45° view}}
    \end{subfigure}
    \hfill
    \begin{subfigure}[b]{0.32\textwidth}
        \centering
        \includegraphics[width=\textwidth, trim=0cm 4cm 0cm 4cm, clip]{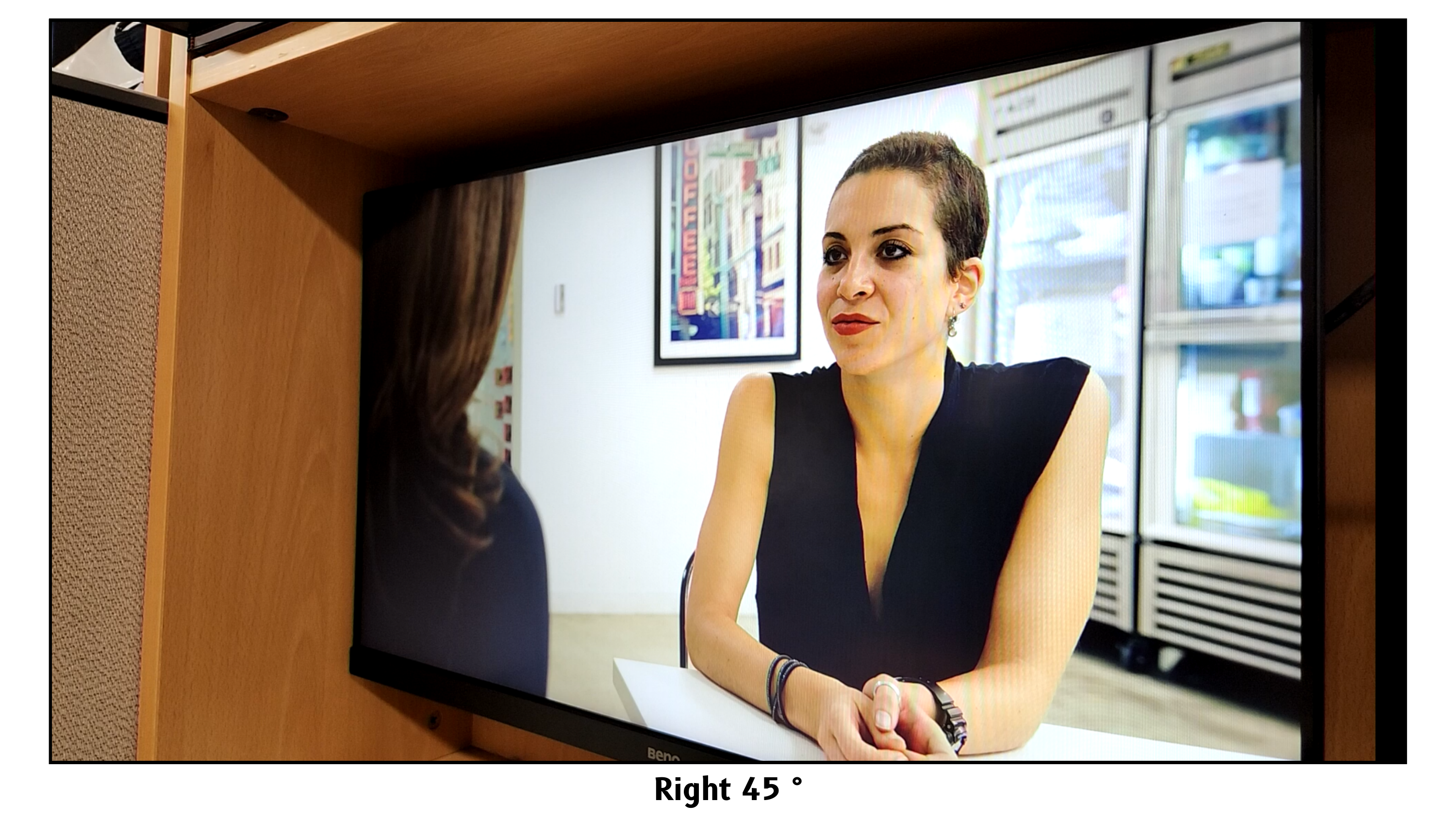}
        \caption{\textsc{Right 45° view}}
    \end{subfigure}
    \hfill
    \begin{subfigure}[b]{0.32\textwidth}
        \centering
        \includegraphics[width=\textwidth, trim=0cm 4cm 0cm 4cm, clip]{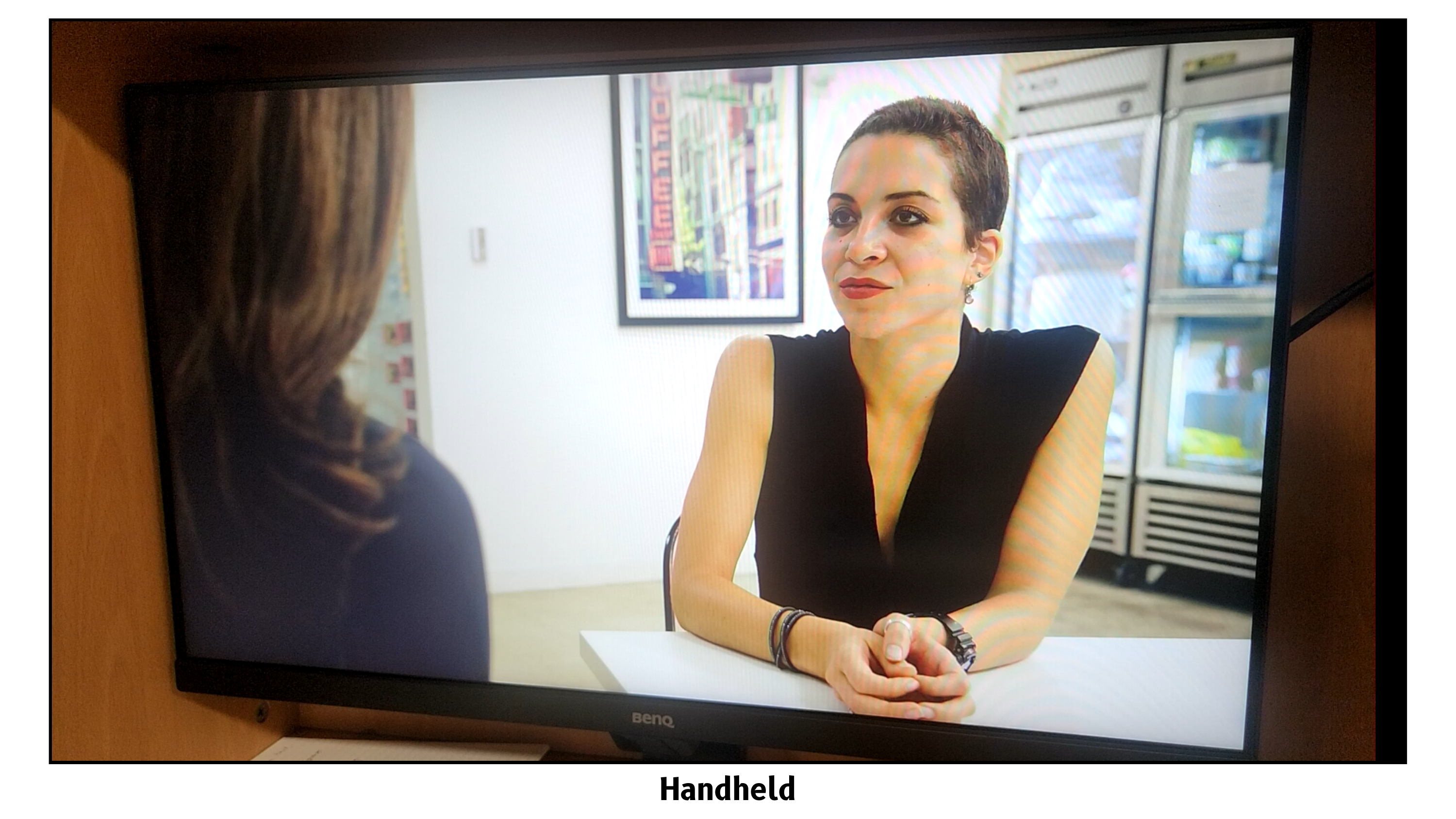}
        \caption{\textsc{Handheld view}}
    \end{subfigure}

    \caption{\textbf{\textsc{Example frames captured from two fixed angles and handheld view. Each showing cropped views of moiré-affected screen content.}}}
    \label{fig:moire_45_views}
\end{figure}

We simulate realistic scenarios in which malicious actors record deepfake content from screens using smartphones. Videos were captured from three different angles to introduce authentic Moiré patterns, as shown in~\autoref{fig:moire_45_views}. Specifically, recordings were taken from a fixed position at a 45° angle from the left (a), a 45° angle from the right (b), and a dynamically moving handheld view (c). This process yielded a total of 12 new videos (6 real, 6 fake) for each of the CelebDF, DFD, DFDC, and UADFV datasets, and 24 new videos (12 real, 12 fake) for the FF++ dataset, reflecting plausible user behaviors and generating diverse real-world distortions.

\subsection{Performance on Moiré Data Captured from ±45° Viewing Directions}


\autoref{table:45degrees} summarizes the performance of both image-based and video-based detectors under authentic moiré conditions induced by varying mobile capture angles. Overall, most detectors exhibited performance degradation under Moiré distortions.
Among video-based models, LipForensics demonstrated the highest resilience, maintaining AUC above 96\% under both 45° angle scenarios, although its performance declined under the more dynamic handheld setting. These observations highlight the complex interplay between real-world artifacts and detector behavior, underscoring the need for comprehensive evaluations beyond ideal clean conditions.

\begin{table}[h!]
\centering
\caption{\textsc{\textbf{Detection AUC of image- and video-based detectors under various moiré pattern conditions. We evaluate four moiré scenarios: LEFT 45°, RIGHT 45°, and handheld captures, in addition to the original (clean) setting. Bold indicates the best performance in each row.}}} 

\label{table:45degrees}
\begin{tabular}{c|c|c|ccc}
\hline
\multicolumn{1}{l|}{\multirow{2}{*}{}} & \multirow{2}{*}{\textsc{\textbf{detectors}}} & \multirow{2}{*}{\textsc{\textbf{original}}} & \multicolumn{3}{c}{\textsc{\textbf{angle}}}                                   \\ \cline{4-6} 
\multicolumn{1}{l|}{}                  &                                     &                                    & \textsc{\textbf{moire left 45}} & \textsc{\textbf{moire right 45}} & \textsc{\textbf{handheld}} \\ \hline
\multirow{3}{*}{\textsc{\textbf{image}}}        & Rossler c23                         & \textbf{90.6}                     & 66.1                  & 71.4                   & 60.0             \\
                                       & MAT                                 & 71.9                              & 73.2                  & \textbf{82.1}          & \textbf{82.9}    \\
                                       & CADDM                               & 78.1                              & \textbf{76.8}         & 78.6                   & 51.4             \\ \hline
\multirow{3}{*}{\textsc{\textbf{video}}}        & Altfreezing                         & \textbf{100.0}                    & 96.4                  & 75.0                   & 59.4             \\
                                       & FTCN                                & 56.3                              & 62.5                  & 68.8                  & 65.6             \\
                                       & LipForensics                        & \textbf{100.0}                    & \textbf{96.9}         & \textbf{96.9}          & \textbf{84.4}    \\ \hline
\end{tabular}
\end{table}

\section{License}

Our dataset is available to download for non-commercial research and education purposes under a Creative Commons Attribution 4.0 International License.
To ensure responsible use of publicly available data, we carefully reviewed the licensing terms of all source datasets used in constructing DeepMoiréFake and contacted the original authors. We have taken appropriate steps to address any potential copyright concerns and designed our distribution process to comply with the varying terms and conditions of the underlying datasets.




%% file: 0_neurips_2025.bbl
\begin{thebibliography}{86}
\providecommand{\natexlab}[1]{#1}
\providecommand{\url}[1]{\texttt{#1}}
\expandafter\ifx\csname urlstyle\endcsname\relax
  \providecommand{\doi}[1]{doi: #1}\else
  \providecommand{\doi}{doi: \begingroup \urlstyle{rm}\Url}\fi

\bibitem[Wierson(2024)]{identityfraud}
Arick Wierson.
\newblock Data breach, identity fraud trends reveal deepfake and generative ai threats.
\newblock \url{https://www.biometricupdate.com/202402/data-breach-identity-fraud-trends-reveal-deepfake-and-generative-ai-threats}, 2024.

\bibitem[Goodfellow et~al.(2014)Goodfellow, Pouget-Abadie, Mirza, Xu, Warde-Farley, Ozair, Courville, and Bengio]{GANs}
Ian Goodfellow, Jean Pouget-Abadie, Mehdi Mirza, Bing Xu, David Warde-Farley, Sherjil Ozair, Aaron Courville, and Yoshua Bengio.
\newblock Generative adversarial nets.
\newblock \emph{Advances in neural information processing systems}, 27, 2014.

\bibitem[Kowalski(2016)]{faceswap}
Marek Kowalski.
\newblock Faceswap - github repository.
\newblock \url{https://github.com/MarekKowalski/FaceSwap}, 2016.

\bibitem[deepfakes(2024)]{deepfakes}
deepfakes.
\newblock faceswap.
\newblock \url{https://github.com/deepfakes/faceswap}, 2024.
\newblock GitHub repository.

\bibitem[Croitoru et~al.(2023)Croitoru, Hondru, Ionescu, and Shah]{DiffusionSurvey}
Florinel-Alin Croitoru, Vlad Hondru, Radu~Tudor Ionescu, and Mubarak Shah.
\newblock Diffusion models in vision: A survey.
\newblock \emph{IEEE Transactions on Pattern Analysis and Machine Intelligence}, 45\penalty0 (9):\penalty0 10850--10869, 2023.

\bibitem[Seow et~al.(2022)Seow, Lim, Phan, and Liu]{seow2022comprehensive}
Jia~Wen Seow, Mei~Kuan Lim, Rapha{\"e}l~CW Phan, and Joseph~K Liu.
\newblock A comprehensive overview of deepfake: Generation, detection, datasets, and opportunities.
\newblock \emph{Neurocomputing}, 513:\penalty0 351--371, 2022.

\bibitem[Hyeongjun~Choi(2025)]{41}
Simon S~Woo Hyeongjun~Choi, Inho~Jung.
\newblock Combating dataset misalignment for robust ai-generated image detection in the real world.
\newblock In \emph{Proceedings of the 4th Workshop on Security Implications of Deepfakes and Cheapfakes}, pages 15--20, 2025.

\bibitem[Choi and Woo(2025)]{36}
Hyeongjun Choi and Simon~S Woo.
\newblock Gan or dm? in-depth analysis and evaluation of ai-generated face data for generalizable deepfake detection.
\newblock In \emph{Proceedings of the 40th ACM/SIGAPP Symposium on Applied Computing}, pages 759--766, 2025.

\bibitem[Tariq et~al.(2023)Tariq, Abuadbba, and Moore]{shahroz-acm}
Shahroz Tariq, Alsharif Abuadbba, and Kristen Moore.
\newblock Deepfake in the metaverse: Security implications for virtual gaming, meetings, and offices.
\newblock In \emph{Proceedings of the 2nd Workshop on Security Implications of Deepfakes and Cheapfakes}, WDC '23, page 16–19, New York, NY, USA, 2023. Association for Computing Machinery.
\newblock ISBN 9798400702037.
\newblock \doi{10.1145/3595353.3595880}.
\newblock URL \url{https://doi.org/10.1145/3595353.3595880}.

\bibitem[Wang et~al.(2023{\natexlab{a}})Wang, Guo, Hu, Chang, and Lyu]{wang2023gan}
Xin Wang, Hui Guo, Shu Hu, Ming-Ching Chang, and Siwei Lyu.
\newblock Gan-generated faces detection: A survey and new perspectives.
\newblock \emph{ECAI 2023}, pages 2533--2542, 2023{\natexlab{a}}.

\bibitem[Kumar et~al.(2023)Kumar, Sharma, et~al.]{kumar2023gan}
Manoj Kumar, Hitesh~Kumar Sharma, et~al.
\newblock A gan-based model of deepfake detection in social media.
\newblock \emph{Procedia Computer Science}, 218:\penalty0 2153--2162, 2023.

\bibitem[Javed et~al.(2022)Javed, Malik, et~al.]{javed2022faceswap}
Ali Javed, Khalid~Mahmood Malik, et~al.
\newblock Faceswap deepfakes detection using novel multi-directional hexadecimal feature descriptor.
\newblock In \emph{2022 19th International Bhurban Conference on Applied Sciences and Technology (IBCAST)}, pages 273--278. IEEE, 2022.

\bibitem[Jain et~al.(2022)Jain, Memon, and Togelius]{jain2022dataless}
Anubhav Jain, Nasir Memon, and Julian Togelius.
\newblock A dataless faceswap detection approach using synthetic images.
\newblock In \emph{2022 IEEE International Joint Conference on Biometrics (IJCB)}, pages 1--7. IEEE, 2022.

\bibitem[Jeon et~al.(2020)Jeon, Bang, Kim, and Woo]{t-gd}
Hyeonseong Jeon, Young~Oh Bang, Junyaup Kim, and Simon Woo.
\newblock T-gd: Transferable gan-generated images detection framework.
\newblock In \emph{International Conference on Machine Learning}, pages 4746--4761. PMLR, 2020.

\bibitem[Tariq et~al.(2021{\natexlab{a}})Tariq, Lee, and Woo]{tariq2021one}
Shahroz Tariq, Sangyup Lee, and Simon Woo.
\newblock One detector to rule them all: Towards a general deepfake attack detection framework.
\newblock In \emph{Proceedings of the web conference 2021}, pages 3625--3637, 2021{\natexlab{a}}.

\bibitem[Muneer and Woo(2025)]{35}
Muhammad~Shahid Muneer and Simon~S Woo.
\newblock Towards safe synthetic image generation on the web: A multimodal robust nsfw defense and million scale dataset.
\newblock In \emph{Companion Proceedings of the ACM on Web Conference 2025}, pages 1209--1213, 2025.

\bibitem[Alam et~al.(2024)Alam, Muneer, and Woo]{28}
Inzamamul Alam, Muhammad~Shahid Muneer, and Simon~S Woo.
\newblock Ugad: Universal generative ai detector utilizing frequency fingerprints.
\newblock In \emph{Proceedings of the 33rd ACM International Conference on Information and Knowledge Management}, pages 4332--4340, 2024.

\bibitem[Park et~al.(2025)Park, Moon, Jeon, Jung, and Woo]{37}
Chan Park, Bohyun Moon, Minsun Jeon, Jee-weon Jung, and Simon~S Woo.
\newblock X3a: Efficient multimodal deepfake detection with score-level fusion.
\newblock In \emph{Proceedings of the 40th ACM/SIGAPP Symposium on Applied Computing}, pages 767--774, 2025.

\bibitem[Woo et~al.(2022)]{binh}
Simon Woo et~al.
\newblock Add: Frequency attention and multi-view based knowledge distillation to detect low-quality compressed deepfake images.
\newblock In \emph{Proceedings of the AAAI conference on artificial intelligence}, volume~36, pages 122--130, 2022.

\bibitem[Tariq et~al.(2021{\natexlab{b}})Tariq, Lee, and Woo]{Shahrozpaper}
Shahroz Tariq, Sangyup Lee, and Simon Woo.
\newblock {One Detector to Rule Them All: Towards a General Deepfake Attack Detection Framework}.
\newblock In \emph{Proceedings of the Web Conference 2021}, WWW '21, page 3625–3637, New York, NY, USA, 2021{\natexlab{b}}. Association for Computing Machinery.
\newblock ISBN 9781450383127.
\newblock \doi{10.1145/3442381.3449809}.
\newblock URL \url{https://doi.org/10.1145/3442381.3449809}.

\bibitem[Khalid et~al.(2021{\natexlab{a}})Khalid, Kim, Tariq, and Woo]{HasamADGD}
Hasam Khalid, Minha Kim, Shahroz Tariq, and Simon~S. Woo.
\newblock {Evaluation of an Audio-Video Multimodal Deepfake Dataset Using Unimodal and Multimodal Detectors}.
\newblock In \emph{Proceedings of the 1st Workshop on Synthetic Multimedia - Audiovisual Deepfake Generation and Detection}, ADGD '21, page 7–15, New York, NY, USA, 2021{\natexlab{a}}. Association for Computing Machinery.
\newblock ISBN 9781450386821.
\newblock \doi{10.1145/3476099.3484315}.
\newblock URL \url{https://doi.org/10.1145/3476099.3484315}.

\bibitem[Jeon and Woo(2025)]{jeon2025seeing}
Minsun Jeon and Simon~S. Woo.
\newblock Seeing through the blur: Unlocking defocus maps for deepfake detection, 2025.

\bibitem[He et~al.(2019)He, Wang, Shi, and Duan]{He_2019_ICCV}
Bin He, Ce~Wang, Boxin Shi, and Ling-Yu Duan.
\newblock Mop moire patterns using mopnet.
\newblock In \emph{Proceedings of the IEEE/CVF International Conference on Computer Vision (ICCV)}, October 2019.

\bibitem[Yang et~al.(2023)Yang, Yang, Ke, Chen, Grzegorzek, and See]{yang2023doing}
Cong Yang, Zhenyu Yang, Yan Ke, Tao Chen, Marcin Grzegorzek, and John See.
\newblock Doing more with moir{\'e} pattern detection in digital photos.
\newblock \emph{IEEE Transactions on Image Processing}, 32:\penalty0 694--708, 2023.

\bibitem[Reid(2023)]{putintv1}
Joy Reid.
\newblock Deepfake of purported putin declaring martial law fits disturbing pattern.
\newblock \url{https://www.msnbc.com/the-reidout/reidout-blog/putin-deepfake-russia-rcna88014}, 2023.

\bibitem[igorsushko(2024)]{putintv2}
igorsushko.
\newblock Putin's new year's address was computer-generated.
\newblock \url{https://x.com/igorsushko/status/1741777672647418168}, 2024.

\bibitem[Shayan86(2024)]{fakenews}
Shayan86.
\newblock President macron has cancelled a scheduled visit to ukraine over fears of an assassination attempt.
\newblock \url{https://x.com/Shayan86/status/1758235524957893064}, 2024.

\bibitem[Rossler et~al.(2019)Rossler, Cozzolino, Verdoliva, Riess, Thies, and Nie{\ss}ner]{faceforensics++}
Andreas Rossler, Davide Cozzolino, Luisa Verdoliva, Christian Riess, Justus Thies, and Matthias Nie{\ss}ner.
\newblock Faceforensics++: Learning to detect manipulated facial images.
\newblock In \emph{Proceedings of the IEEE/CVF international conference on computer vision}, pages 1--11, 2019.

\bibitem[Dufour et~al.(2019)Dufour, Gully, Karlsson, Vorbyov, Leung, Childs, and Bregler]{DFD}
Nicholas Dufour, Andrew Gully, Per Karlsson, Alexey~Victor Vorbyov, Thomas Leung, Jeremiah Childs, and Christoph Bregler.
\newblock Deepfakes detection dataset by google \& jigsaw.
\newblock \url{https://research.google/blog/contributing-data-to-deepfake-detection-research/}, 2019.
\newblock arXiv preprint arXiv:1901.08971; dataset release by Google \& Jigsaw.

\bibitem[Dolhansky et~al.(2019)Dolhansky, Howes, Pflaum, Baram, and Ferrer]{DFDC}
Brian Dolhansky, Russ Howes, Ben Pflaum, Nicole Baram, and Cristian~Canton Ferrer.
\newblock The deepfake detection challenge (dfdc) preview dataset.
\newblock \emph{arXiv preprint arXiv:1910.08854}, 2019.

\bibitem[Li et~al.(2020)Li, Yang, Sun, Qi, and Lyu]{CelebDF}
Yuezun Li, Xin Yang, Pu~Sun, Honggang Qi, and Siwei Lyu.
\newblock Celeb-df: A large-scale challenging dataset for deepfake forensics.
\newblock In \emph{Proceedings of the IEEE/CVF conference on computer vision and pattern recognition}, pages 3207--3216, 2020.

\bibitem[Yang et~al.(2019)Yang, Li, and Lyu]{UADFV}
Xin Yang, Yuezun Li, and Siwei Lyu.
\newblock Exposing deep fakes using inconsistent head poses.
\newblock In \emph{ICASSP 2019-2019 IEEE International Conference on Acoustics, Speech and Signal Processing (ICASSP)}, pages 8261--8265. IEEE, 2019.

\bibitem[Tariq et~al.(2024{\natexlab{a}})Tariq, Tariq, and Woo]{10647902}
Razaib Tariq, Shahroz Tariq, and Simon~S. Woo.
\newblock Exploring the impact of moire pattern on deepfake detectors.
\newblock In \emph{2024 IEEE International Conference on Image Processing (ICIP)}, pages 3813--3819, 2024{\natexlab{a}}.
\newblock \doi{10.1109/ICIP51287.2024.10647902}.

\bibitem[Tariq et~al.(2024{\natexlab{b}})Tariq, Heo, Woo, and Tariq]{tariq2024beyond}
Razaib Tariq, Minji Heo, Simon~S Woo, and Shahroz Tariq.
\newblock Beyond the screen: Evaluating deepfake detectors under moire pattern effects.
\newblock In \emph{Proceedings of the IEEE/CVF Conference on Computer Vision and Pattern Recognition}, pages 4429--4439, 2024{\natexlab{b}}.

\bibitem[Kingma and Welling(2013)]{VAE}
Diederik~P Kingma and Max Welling.
\newblock Auto-encoding variational bayes.
\newblock \emph{arXiv preprint arXiv:1312.6114}, 2013.

\bibitem[Ho et~al.(2020)Ho, Jain, and Abbeel]{diffusionpaper}
Jonathan Ho, Ajay Jain, and Pieter Abbeel.
\newblock Denoising diffusion probabilistic models.
\newblock \emph{Advances in neural information processing systems}, 33:\penalty0 6840--6851, 2020.

\bibitem[Gamb{\'\i}n et~al.(2024)Gamb{\'\i}n, Yazidi, Vasilakos, Haugerud, and Djenouri]{DeepfakeSurvey1}
{\'A}ngel~Fern{\'a}ndez Gamb{\'\i}n, Anis Yazidi, Athanasios Vasilakos, H{\aa}rek Haugerud, and Youcef Djenouri.
\newblock Deepfakes: current and future trends.
\newblock \emph{Artificial Intelligence Review}, 57\penalty0 (3):\penalty0 64, 2024.

\bibitem[Le et~al.(2025)Le, Kim, Woo, Moore, Abuadbba, and Tariq]{sok}
Binh~M Le, Jiwon Kim, Simon~S Woo, Kristen Moore, Alsharif Abuadbba, and Shahroz Tariq.
\newblock Sok: Systematization and benchmarking of deepfake detectors in a unified framework.
\newblock In \emph{2025 IEEE 10th European Symposium on Security and Privacy (EuroS\&P)}, pages 883--902. IEEE, 2025.

\bibitem[Thies et~al.(2019)Thies, Zollh{\"o}fer, and Nie{\ss}ner]{NeuralTexture}
Justus Thies, Michael Zollh{\"o}fer, and Matthias Nie{\ss}ner.
\newblock Deferred neural rendering: Image synthesis using neural textures.
\newblock \emph{Acm Transactions on Graphics (TOG)}, 38\penalty0 (4):\penalty0 1--12, 2019.

\bibitem[Thies et~al.(2016)Thies, Zollhofer, Stamminger, Theobalt, and Nie{\ss}ner]{Face2Face}
Justus Thies, Michael Zollhofer, Marc Stamminger, Christian Theobalt, and Matthias Nie{\ss}ner.
\newblock Face2face: Real-time face capture and reenactment of rgb videos.
\newblock In \emph{Proceedings of the IEEE conference on computer vision and pattern recognition}, pages 2387--2395, 2016.

\bibitem[Beckmann et~al.(2023)Beckmann, Hilsmann, and Eisert]{FaceRenactment}
Arian Beckmann, Anna Hilsmann, and Peter Eisert.
\newblock Fooling state-of-the-art deepfake detection with high-quality deepfakes.
\newblock In \emph{Proceedings of the 2023 ACM Workshop on Information Hiding and Multimedia Security}, pages 175--180, 2023.

\bibitem[Wang et~al.(2023{\natexlab{b}})Wang, Huang, Cheng, Ma, and Wang]{FaceAttributeEditing}
Tianyi Wang, Mengxiao Huang, Harry Cheng, Bin Ma, and Yinglong Wang.
\newblock Robust identity perceptual watermark against deepfake face swapping.
\newblock \emph{arXiv preprint arXiv:2311.01357}, 2023{\natexlab{b}}.

\bibitem[Heo and Woo(2025)]{heo2025fakechain}
Minji Heo and Simon~S. Woo.
\newblock Fakechain: Exposing shallow cues in multi-step deepfake detection, 2025.

\bibitem[Kim et~al.(2022)Kim, Kim, Cho, Seo, Nam, Lee, Kim, and Lee]{FaceSynthesis}
Kihong Kim, Yunho Kim, Seokju Cho, Junyoung Seo, Jisu Nam, Kychul Lee, Seungryong Kim, and KwangHee Lee.
\newblock Diffface: Diffusion-based face swapping with facial guidance.
\newblock \emph{arXiv preprint arXiv:2212.13344}, 2022.

\bibitem[Karras et~al.(2017)Karras, Aila, Laine, and Lehtinen]{ProGAN}
Tero Karras, Timo Aila, Samuli Laine, and Jaakko Lehtinen.
\newblock Progressive growing of gans for improved quality, stability, and variation.
\newblock \emph{arXiv preprint arXiv:1710.10196}, 2017.

\bibitem[Korshunov and Marcel(2018)]{df-timit}
Pavel Korshunov and S{\'{e}}bastien Marcel.
\newblock Deepfakes: a new threat to face recognition? assessment and detection.
\newblock \emph{CoRR}, abs/1812.08685, 2018.
\newblock URL \url{http://arxiv.org/abs/1812.08685}.

\bibitem[Jiang et~al.(2020)Jiang, Li, Wu, Qian, and Loy]{jiang2020deeperforensics}
Liming Jiang, Ren Li, Wayne Wu, Chen Qian, and Chen~Change Loy.
\newblock Deeperforensics-1.0: A large-scale dataset for real-world face forgery detection.
\newblock In \emph{Proceedings of the IEEE/CVF Conference on Computer Vision and Pattern Recognition}, pages 2889--2898, 2020.

\bibitem[Dolhansky et~al.(2020)Dolhansky, Bitton, Pflaum, Lu, Howes, Wang, and Ferrer]{dolhansky2020deepfake}
Brian Dolhansky, Joanna Bitton, Ben Pflaum, Jikuo Lu, Russ Howes, Menglin Wang, and Cristian~Canton Ferrer.
\newblock The deepfake detection challenge dataset.
\newblock \emph{arXiv preprint arXiv:2006.07397}, 2020.

\bibitem[Kwon et~al.(2021)Kwon, You, Nam, Park, and Chae]{kwon2021kodf}
Patrick Kwon, Jaeseong You, Gyuhyeon Nam, Sungwoo Park, and Gyeongsu Chae.
\newblock Kodf: A large-scale korean deepfake detection dataset.
\newblock \emph{arXiv preprint arXiv:2103.10094}, 2021.

\bibitem[Khalid et~al.(2021{\natexlab{b}})Khalid, Tariq, Kim, and Woo]{FakeAVCeleb}
Hasam Khalid, Shahroz Tariq, Minha Kim, and Simon~S Woo.
\newblock Fakeavceleb: A novel audio-video multimodal deepfake dataset.
\newblock \emph{arXiv preprint arXiv:2108.05080}, 2021{\natexlab{b}}.

\bibitem[Tariq et~al.(2018)Tariq, Lee, Kim, Shin, and Woo]{Shahroz1}
Shahroz Tariq, Sangyup Lee, Hoyoung Kim, Youjin Shin, and Simon~S. Woo.
\newblock {Detecting Both Machine and Human Created Fake Face Images In the Wild}.
\newblock In \emph{Proceedings of the 2nd International Workshop on Multimedia Privacy and Security}, MPS '18, page 81–87, New York, NY, USA, 2018. Association for Computing Machinery.
\newblock ISBN 9781450359887.
\newblock \doi{10.1145/3267357.3267367}.
\newblock URL \url{https://doi.org/10.1145/3267357.3267367}.

\bibitem[Tariq et~al.(2019)Tariq, Lee, Kim, Shin, and Woo]{Shahroz2}
Shahroz Tariq, Sangyup Lee, Hoyoung Kim, Youjin Shin, and Simon~S. Woo.
\newblock {GAN is a Friend or Foe? A Framework to Detect Various Fake Face Images}.
\newblock In \emph{Proceedings of the 34th ACM/SIGAPP Symposium on Applied Computing}, SAC '19, page 1296–1303, New York, NY, USA, 2019. Association for Computing Machinery.
\newblock ISBN 9781450359337.
\newblock \doi{10.1145/3297280.3297410}.
\newblock URL \url{https://doi.org/10.1145/3297280.3297410}.

\bibitem[Kim et~al.(2021{\natexlab{a}})Kim, Tariq, and Woo]{Minha_FReTAL}
Minha Kim, Shahroz Tariq, and Simon~S. Woo.
\newblock {FReTAL: Generalizing Deepfake Detection Using Knowledge Distillation and Representation Learning}.
\newblock In \emph{Proceedings of the IEEE/CVF Conference on Computer Vision and Pattern Recognition (CVPR) Workshops}, pages 1001--1012, June 2021{\natexlab{a}}.

\bibitem[Kim et~al.(2021{\natexlab{b}})Kim, Tariq, and Woo]{Minha_CoReD}
Minha Kim, Shahroz Tariq, and Simon~S. Woo.
\newblock {CoReD: Generalizing Fake Media Detection with Continual Representation Using Distillation}.
\newblock In \emph{Proceedings of the 29th ACM International Conference on Multimedia}, MM '21, page 337–346, New York, NY, USA, 2021{\natexlab{b}}. Association for Computing Machinery.
\newblock ISBN 9781450386517.
\newblock \doi{10.1145/3474085.3475535}.
\newblock URL \url{https://doi.org/10.1145/3474085.3475535}.

\bibitem[Lee et~al.(2021{\natexlab{a}})Lee, Tariq, Shin, and Woo]{SAMGAN}
Sangyup Lee, Shahroz Tariq, Youjin Shin, and Simon~S. Woo.
\newblock {Detecting handcrafted facial image manipulations and GAN-generated facial images using Shallow-FakeFaceNet}.
\newblock \emph{Applied Soft Computing}, 105:\penalty0 107256, 2021{\natexlab{a}}.
\newblock ISSN 1568-4946.
\newblock \doi{https://doi.org/10.1016/j.asoc.2021.107256}.
\newblock URL \url{https://www.sciencedirect.com/science/article/pii/S1568494621001794}.

\bibitem[Lee et~al.(2021{\natexlab{b}})Lee, Tariq, Kim, and Woo]{SAMTAR}
Sangyup Lee, Shahroz Tariq, Junyaup Kim, and Simon~S. Woo.
\newblock {TAR: Generalized Forensic Framework to Detect Deepfakes Using Weakly Supervised Learning}.
\newblock In Audun J{\o}sang, Lynn Futcher, and Janne Hagen, editors, \emph{ICT Systems Security and Privacy Protection}, pages 351--366, Cham, 2021{\natexlab{b}}. Springer International Publishing.
\newblock ISBN 978-3-030-78120-0.

\bibitem[Tariq et~al.(2020)Tariq, Lee, and Woo]{CLRNet}
Shahroz Tariq, Sangyup Lee, and Simon~S Woo.
\newblock {A Convolutional LSTM based Residual Network for Deepfake Video Detection}.
\newblock \emph{arXiv preprint arXiv:2009.07480}, 2020.

\bibitem[Coccomini et~al.(2022)Coccomini, Messina, Gennaro, and Falchi]{coccomini2022combining}
Davide~Alessandro Coccomini, Nicola Messina, Claudio Gennaro, and Fabrizio Falchi.
\newblock Combining efficientnet and vision transformers for video deepfake detection.
\newblock In \emph{International conference on image analysis and processing}, pages 219--229. Springer, 2022.

\bibitem[Khormali and Yuan(2022)]{app12062953}
Aminollah Khormali and Jiann-Shiun Yuan.
\newblock Dfdt: An end-to-end deepfake detection framework using vision transformer.
\newblock \emph{Applied Sciences}, 12\penalty0 (6), 2022.
\newblock ISSN 2076-3417.
\newblock \doi{10.3390/app12062953}.
\newblock URL \url{https://www.mdpi.com/2076-3417/12/6/2953}.

\bibitem[Wang et~al.(2021)Wang, Wu, Chen, and Jiang]{M2TR}
Junke Wang, Zuxuan Wu, Jingjing Chen, and Yu{-}Gang Jiang.
\newblock {M2TR:} multi-modal multi-scale transformers for deepfake detection.
\newblock \emph{CoRR}, abs/2104.09770, 2021.
\newblock URL \url{https://arxiv.org/abs/2104.09770}.

\bibitem[Niu et~al.(2021)Niu, Guo, and Wang]{MoiréAttack}
Dantong Niu, Ruohao Guo, and Yisen Wang.
\newblock Moir{\'{e}} attack {(MA):} {A} new potential risk of screen photos.
\newblock \emph{CoRR}, abs/2110.10444, 2021.
\newblock URL \url{https://arxiv.org/abs/2110.10444}.

\bibitem[Choi(2024)]{MoireAttack2}
Sungjun Choi.
\newblock Simulating moire effects seen in photos of digital device screens.
\newblock \url{https://github.com/mr3coi/screen_photo_simulator}, 2024.

\bibitem[Shiohara and Yamasaki(2022)]{selfblended}
Kaede Shiohara and Toshihiko Yamasaki.
\newblock Detecting deepfakes with self-blended images.
\newblock In \emph{Proceedings of the IEEE/CVF Conference on Computer Vision and Pattern Recognition}, pages 18720--18729, 2022.

\bibitem[He et~al.(2021)He, Gan, Chen, Zhou, Yin, Song, Sheng, Shao, and Liu]{he2021forgerynet}
Yinan He, Bei Gan, Siyu Chen, Yichun Zhou, Guojun Yin, Luchuan Song, Lu~Sheng, Jing Shao, and Ziwei Liu.
\newblock Forgerynet: A versatile benchmark for comprehensive forgery analysis.
\newblock In \emph{Proceedings of the IEEE/CVF conference on computer vision and pattern recognition}, pages 4360--4369, 2021.

\bibitem[Nguyen et~al.(2019)Nguyen, Yamagishi, and Echizen]{nguyen2019capsule}
Huy~H Nguyen, Junichi Yamagishi, and Isao Echizen.
\newblock Capsule-forensics: Using capsule networks to detect forged images and videos.
\newblock In \emph{ICASSP 2019-2019 IEEE international conference on acoustics, speech and signal processing (ICASSP)}, pages 2307--2311. IEEE, 2019.

\bibitem[Zhao et~al.(2021)Zhao, Zhou, Chen, Wei, Zhang, and Yu]{DBLP:journals/corr/abs-2103-02406}
Hanqing Zhao, Wenbo Zhou, Dongdong Chen, Tianyi Wei, Weiming Zhang, and Nenghai Yu.
\newblock Multi-attentional deepfake detection.
\newblock \emph{CoRR}, abs/2103.02406, 2021.
\newblock URL \url{https://arxiv.org/abs/2103.02406}.

\bibitem[Dong et~al.(2023)Dong, Wang, Ji, Liang, Fan, and Ge]{dong2023implicit}
Shichao Dong, Jin Wang, Renhe Ji, Jiajun Liang, Haoqiang Fan, and Zheng Ge.
\newblock Implicit identity leakage: The stumbling block to improving deepfake detection generalization.
\newblock In \emph{Proceedings of the IEEE/CVF Conference on Computer Vision and Pattern Recognition}, pages 3994--4004, 2023.

\bibitem[Wang et~al.(2023{\natexlab{c}})Wang, Bao, Zhou, Wang, and Li]{wang2023altfreezing}
Zhendong Wang, Jianmin Bao, Wengang Zhou, Weilun Wang, and Houqiang Li.
\newblock Altfreezing for more general video face forgery detection.
\newblock In \emph{Proceedings of the IEEE/CVF Conference on Computer Vision and Pattern Recognition}, pages 4129--4138, 2023{\natexlab{c}}.

\bibitem[Zheng et~al.(2021)Zheng, Bao, Chen, Zeng, and Wen]{zheng2021exploring}
Yinglin Zheng, Jianmin Bao, Dong Chen, Ming Zeng, and Fang Wen.
\newblock Exploring temporal coherence for more general video face forgery detection.
\newblock In \emph{Proceedings of the IEEE/CVF international conference on computer vision}, pages 15044--15054, 2021.

\bibitem[Sun et~al.(2021)Sun, Han, Hua, Ruan, and Jia]{sun2021improving}
Zekun Sun, Yujie Han, Zeyu Hua, Na~Ruan, and Weijia Jia.
\newblock Improving the efficiency and robustness of deepfakes detection through precise geometric features.
\newblock In \emph{Proceedings of the IEEE/CVF Conference on Computer Vision and Pattern Recognition}, pages 3609--3618, 2021.

\bibitem[Haliassos et~al.(2021)Haliassos, Vougioukas, Petridis, and Pantic]{haliassos2021lips}
Alexandros Haliassos, Konstantinos Vougioukas, Stavros Petridis, and Maja Pantic.
\newblock Lips don't lie: A generalisable and robust approach to face forgery detection.
\newblock In \emph{Proceedings of the IEEE/CVF conference on computer vision and pattern recognition}, pages 5039--5049, 2021.

\bibitem[Schallek et~al.(2009)Schallek, Li, Kardon, Kwon, Abràmoff, Soliz, and Ts'o]{Schallek2009Stimulus-evoked}
Jesse Schallek, Hongbin Li, R.~Kardon, Young~H. Kwon, M.~Abràmoff, P.~Soliz, and D.~Ts'o.
\newblock Stimulus-evoked intrinsic optical signals in the retina: spatial and temporal characteristics.
\newblock \emph{Investigative ophthalmology \& visual science}, 50 10:\penalty0 4865--72, 2009.
\newblock \doi{10.1167/iovs.08-3290}.

\bibitem[Saveljev and Kim(2012)]{Saveljev:12}
Vladimir Saveljev and Sung-Kyu Kim.
\newblock Simulation and measurement of moir\'{e} patterns at finite distance.
\newblock \emph{Opt. Express}, 20\penalty0 (3):\penalty0 2163--2177, Jan 2012.
\newblock \doi{10.1364/OE.20.002163}.
\newblock URL \url{https://opg.optica.org/oe/abstract.cfm?URI=oe-20-3-2163}.

\bibitem[Yan et~al.(2024)Yan, Zhang, Yuan, Lyu, and Wu]{DeepfakeBench}
Zhiyuan Yan, Yong Zhang, Xinhang Yuan, Siwei Lyu, and Baoyuan Wu.
\newblock Deepfakebench: a comprehensive benchmark of deepfake detection.
\newblock In \emph{Proceedings of the 37th International Conference on Neural Information Processing Systems}, NIPS '23, Red Hook, NY, USA, 2024. Curran Associates Inc.

\bibitem[Compression(2024)]{video_compression_social_media}
Video Compression.
\newblock Video compression algorithm by social media application.
\newblock \url{https://getstream.io/glossary/video-compression/}, 2024.

\bibitem[ian bremmer(2023)]{moireonline1}
ian bremmer.
\newblock Visual moir{\'e} artifacts in video of putin calling for martial law.
\newblock \url{https://x.com/ianbremmer/status/1665841241349668864}, 2023.

\bibitem[Flibustier(2024)]{moireonline2}
Vincent Flibustier.
\newblock interesting technique to give credibility to a fake: filming.
\newblock \url{https://x.com/vinceflibustier/status/1758521285628383505}, 2024.

\bibitem[Sun et~al.(2018)Sun, Yu, and Wang]{DMCNN}
Yujing Sun, Yizhou Yu, and Wenping Wang.
\newblock Moir{\'e} photo restoration using multiresolution convolutional neural networks.
\newblock \emph{IEEE Transactions on Image Processing}, 27\penalty0 (8):\penalty0 4160--4172, 2018.

\bibitem[Zheng et~al.(2020)Zheng, Yuan, Slabaugh, and Leonardis]{MBCNN}
Bolun Zheng, Shanxin Yuan, Gregory Slabaugh, and Ales Leonardis.
\newblock Image demoireing with learnable bandpass filters.
\newblock In \emph{Proceedings of the IEEE/CVF conference on computer vision and pattern recognition}, pages 3636--3645, 2020.

\bibitem[Yu et~al.(2022)Yu, Dai, Li, Ma, Shen, Li, and Qi]{ESDNet}
Xin Yu, Peng Dai, Wenbo Li, Lan Ma, Jiajun Shen, Jia Li, and Xiaojuan Qi.
\newblock Towards efficient and scale-robust ultra-high-definition image demoir{\'e}ing.
\newblock In \emph{European Conference on Computer Vision}, pages 646--662. Springer, 2022.

\bibitem[Zhang et~al.(2023)Zhang, Lin, Li, Liu, Wang, Chao, Ren, Wen, Chen, and Ji]{DDA}
Yuxin Zhang, Mingbao Lin, Xunchao Li, Han Liu, Guozhi Wang, Fei Chao, Shuai Ren, Yafei Wen, Xiaoxin Chen, and Rongrong Ji.
\newblock Real-time image demoireing on mobile devices.
\newblock \emph{arXiv preprint arXiv:2302.02184}, 2023.

\bibitem[OpenCV(2009{\natexlab{a}})]{smooth_1}
OpenCV.
\newblock Gaussianblur.
\newblock \url{https://docs.opencv.org/4.x/d4/d13/tutorial_py_filtering.html}, 2009{\natexlab{a}}.

\bibitem[OpenCV(2009{\natexlab{b}})]{sharp}
OpenCV.
\newblock Sharpening.
\newblock \url{https://docs.opencv.org/4.x/d2/dbd/tutorial_distance_transform.html}, 2009{\natexlab{b}}.

\bibitem[Dai et~al.(2022)Dai, Yu, Ma, Zhang, Li, Li, Shen, and Qi]{dai2022video}
Peng Dai, Xin Yu, Lan Ma, Baoheng Zhang, Jia Li, Wenbo Li, Jiajun Shen, and Xiaojuan Qi.
\newblock Video demoireing with relation-based temporal consistency.
\newblock In \emph{Proceedings of the IEEE/CVF Conference on Computer Vision and Pattern Recognition}, 2022.

\bibitem[Oh et~al.(2025)Oh, Kim, Gu, Yoon, Kim, and Kim]{oh2025fpanet}
Gyeongrok Oh, Sungjune Kim, Heon Gu, Sang~Ho Yoon, Jinkyu Kim, and Sangpil Kim.
\newblock Fpanet: Frequency-based video demoireing using frame-level post alignment.
\newblock \emph{Neural Networks}, 184:\penalty0 107021, 2025.

\bibitem[Chen et~al.(2022)Chen, Chu, Zhang, and Sun]{chen2022simple}
Liangyu Chen, Xiaojie Chu, Xiangyu Zhang, and Jian Sun.
\newblock Simple baselines for image restoration.
\newblock \emph{arXiv preprint arXiv:2204.04676}, 2022.

\end{thebibliography}
